# Probabilistic Extension to the Concurrent Constraint Factor Oracle Model for Music Improvisation

Mauricio Toro

Universidad EAFIT

**Abstract**

We can program a Real-Time (RT) music improvisation system in C++ without a formal semantic or we can model it with process calculi such as the Non-deterministic Timed Concurrent Constraint (ntcc) calculus. "A Concurrent Constraints Factor Oracle (FO) model for Music Improvisation" (Ccfomi) is an improvisation model specified on ntcc. Since Ccfomi improvises non-deterministically, there is no control on choices and therefore little control over the sequence variation during the improvisation. To avoid this, we extended Ccfomi using the Probabilistic Non-deterministic Timed Concurrent Constraint calculus. Our extension to Ccfomi does not change the time and space complexity of building the FO, thus making our extension compatible with RT. However, there was not a ntcc interpreter capable of RT to execute Ccfomi. We developed Ntccrt –a RT capable interpreter for ntcc– and we executed Ccfomi on Ntccrt. In the future, we plan to extend Ntccrt to execute our extension to Ccfomi.

# 1 Introduction

There are two different approaches to develop multimedia interaction systems (e.g., machine improvisation).

One may think that in order to implement real-time capable systems, those systems should be written directly in C++ for efficiency. In contrast, one may argue that multimedia interaction systems –inherently concurrent– should not be written directly in C or C++ because there is not a formalism to reason about concurrency in C++. We argue that those systems should be modeled using a process calculus with formal semantics and verification procedures, and execute those models on a real-time capable interpreter. That will be our definition for real-time in the rest of this document.

Garavel explains in [13] that models based on process calculi are not widespread because there are many calculi and many variants for each calculus, being difficult to choose the most appropriate. In addition, it is difficult to express an explicit notion of time and real-time requirements in process calculi. Finally, he argues that existing tools for process calculi are not user-friendly.

## 1.1 Motivation

Defending the calculi approach, Rueda et al. [38],[40] explain that using the semantics and logic underlying the Non-deterministic Timed Concurrent Constraint (`ntcc`) [24] calculus, it is possible to prove properties of the `ntcc` models before executing them and execute the models on a `ntcc` interpreter. We define soft real-time multimedia interaction means that the system reacts fast enough to interact with human players without letting them notice delays.

One may disagree with Rueda et al., arguing that although there are several interpreters for `ntcc` such as Lman [22] and Rueda's Interpreter [40], there is not a generic interpreter to run `ntcc` models in real-time.

We agree with Rueda et al. about the way to develop those systems, but we also argue that currently there are no `ntcc` interpreters capable of real-time. We argue, in agreement with Rueda et al.'s argument, that models based on `ntcc` such as *"A Concurrent Constraints Factor Oracle model for Music Improvisation"* (*Ccfomi*) [40] are a good alternative to model multimedia interaction because synchronization is presented declaratively by means of variable sharing among concurrent agents reasoning about information contained in a global *store*. However, due to non-deterministic choices, improvisation in Ccfomi can be repetitive (i.e., it produces loops without control). In addition, since Ccfomi does not change the intensity of the learned notes, Ccfomi may produce a sharp difference in the relative loudness between what a musician plays and what the improviser plays.

Process calculi has been applied to the modeling of interactive music systems [3, 58, 54, 26, 52, 49, 51, 53, 4, 57, 50, 55, 56, 48] and ecological systems [30, 60, 31, 59].

Our main objective is extending *Ccfomi* to model probabilistic choice of musical sequences. We also want to show that a `ntcc` model can interact with a human player in soft real-time using a `ntcc` interpreter. For that reason, we developed Ntccrt, a generic real-time interpreter for `ntcc`.

The rest of this introduction is organized as follows. First section, gives a definition of music improvisation. Second section, explains machine improvisation. Third section, gives a brief introduction to `ntcc` and presents systems modeled with `ntcc`. After explaining the intuitions about music improvisation, machine improvisation, and `ntcc` we explain our solution to extend *Ccfomi* in Section fourth section. Fifth section explains the contributions of this thesis work. Finally, sixth section explains the organization of the following chapters.

## 1.2 Music improvisation

"Musical improvisation is the spontaneous creative process of making music while it is being performed. To use a linguistic analogy, improvisation is like speaking or having a conversation as opposed to reciting a written text. Among jazz musicians there is an adage, *improvisation is composition speeded up*, and vice versa, *composition is improvisation slowed down*."[23]

Improvisation exists in almost all music generel. However, improvisation is most frequently associated with melodic improvisation as it is found in jazz. However, spontaneous real-time variation in performance of tempo and dynamics within a classical performance may also be considered as improvisation [23]

The reader may see an example of music improvisation in [35], where musician Alberto Riascos improvised in the Colombian music genre *Guabina*[1] and explained us how he did it.

## 1.3 Machine improvisation

Machine improvisation is the simulation of music improvisation by the computer. This process builds a representation of music, either by explicit coding of rules or applying machine learning methods. For real-time machine improvisation it is necessary to perform two phases concurrently: *Stylistic learning* and *Stylistic*

1 Guabina is a Colombian traditional music very common in the regions of Antioquia, Santander, Boyacá, Cundinamarca, Tolima, and Huila.

*simulation*. In addition, to perform both phases concurrently, the system must be able to interact in real-time with human players [40].

Rueda et al. define *Stylistic learning* as the process of applying such methods to musical sequences in order to capture important musical features and organize these features into a model, and the *Stylistic simulation* as the process producing musical sequences stylistically consistent with the learned style [40]. An example of a system running concurrently both phases is *Ccfomi*, a system using the Factor Oracle (*FO*) to store the information of the learned sequences and the `ntcc` calculus to synchronize both phases of the improvisation.

## 1.4 Introduction to `ntcc`

The `ntcc` calculus is a mathematic formalism used to represent reactive systems with synchronous, asynchronous and/or non-deterministic behavior. This formalism and its extensions have been used to model systems such as: musical improvisation systems [40], [29], [45], an audio processing framework [42], and interactive scores [2], [45].

`Ntcc` is not only useful for multimedia semantic interaction, it has also been used in other fields such as modeling molecular biology [41], analyzing biological systems [16], and security protocols [20] because these fields also include the study of complex interactions where we want to observe certain properties showing up and to model the answer of the system to them. Modeling of molecular biology, security protocols, and multimedia semantic interaction using process calculi are the base of the project Robust Theories for Emerging Applications in Concurrency Theory (REACT[2]).

The novelty of this approach is the specification of the synchronization in a declarative way, opposed to programming languages such as C++, where the programmer has to specify multiple steps to guarantee a correct synchronization and safe access to shared resources. Further explanation about `ntcc` and how to model musical processes in `ntcc` is presented in Chapter 1.

## 1.5 Our solution

To avoid a repetitive improvisation, we extend *Ccfomi* with the Probabilistic Non-deterministic Timed Concurrent Constraint (`pntcc`) calculus [29] to decrease the probability of choosing a sequence previously improvised. This idea is based on the Probabilistic *Ccfomi* model [29] developed by Pérez and Rueda. That model, chooses the improvised sequences probabilistically, based on a probability distribution. Unfortunately, Probabilistic *Ccfomi* does not give information about how that probability distribution can be built nor how it can change through time according to the user and the computer interaction. Our model is the first `pntcc` model, as far as we know, where probability distributions change from a time-unit to another.

For instance, consider that our system can play in a certain moment the pitches (i.e., the frecuency of the notes) *a*,*b* and *c* with an equal probability. Then it outputs the sequence "*aaba*". After that, it is going to choose another pitch. When choosing this pitch, *c* has a greater probability to be chosen than *b*, and *b* has a greater probability to be chosen than *a* because *a* was played three times and *b* once in the last sequence. Using this probabilistic extension, we avoid multiple cycles in the improvisation which can happen without control in *Ccfomi*.

On the other hand, to be coherent with the relative loudness on which the user is currently playing, we change the intensity of the improvised notes. This idea is based on interviews with musicians Riascos [35] and Juan Manuel Collazos [7], where they

2 This thesis is partially funded by the REACT project, sponsored by Colciencias.

argue that this is a technique they use when improvising and improves the "quality" of the improvisation, when two or more persons improvise at the same time.

For instance, if the computer plays five notes with intensities (measured from 0 to 127) 54, 65, 30, 58, 91 and the user plays, at the same time, four notes with intensities 10, 21, 32, 5; they are incoherence results because the user is playing low and the computer is playing loud. For that reason, our system multiplies its intensities by a factor of 0.29 (the relation of the average of both sequences) changing the intensities of the computer output to 16, 19, 9, 17, 26.

## 1.6 Contributions

**Gecol 2**. Our first approach to provide an interface for Gecode to Common Lisp was to extend Gecol to work with Gecode 2. Using Gecol 2 we wrote several prototypes for the `ntcc` interpreter. Examples, sources, and binaries can be found at http://common-lisp.net/project/gecol/.

**Ntccrt**. A real-time capable interpreter for `ntcc`. Using Ntccrt, we executed *Ccfomi*. Examples, sources and binaries can be found at http://ntccrt.sourceforge.net. An article about Ntccrt is to be published this year. **Gelisp**. A new graphical constraint solving library for OpenMusic. We plan to use it in the future for a closer integration between *Ntccrt* and OpenMusic. Examples, sources, and binaries can be found at http://gelisp.sourceforge.net. An article about Gelisp is to be publish this year. The original version of Gelisp was developed by Rueda for Common Lisp [39]. **Technical report**. A report including all the implementation details of Ntccrt, the graphical interface for Gelisp, Gecol 2, applications of Ntccrt, and our previous attempts to develop a real-time ntcc interpreter [48].

## 1.7 Organization

The structure of this thesis is the following. In Chapter 1, we explain the background concepts. Chapter 2 focuses on the modeling of *Ccfomi* to allow probabilistic choice of musical sequences. Chapter 3 explains the modifications to *Ccfomi* to allow variation of the intensity of learned notes during the style simulation phase. Chapter 4 describes our model in `pntcc`. Chapter 5 explains the design and implementation of Ntccrt, our real-time interpreter for `ntcc`. Chapter 6 shows some results and tests made with the interpreter. Finally, in Chapter 7, we present a summary of this thesis, concluding remarks, and propose some future work.

# 2 Background

## 2.1 Concurrent Constraint Programming (CCP)

*Concurrent Constraint Programming (CCP [44])* is a model for concurrent systems. In CCP, a concurrent system is modeled in terms of constraints over the system variables and in terms of agents interacting with partial information obtained from those variables. A constraint is a formula representing partial information about the values of some of the system variables. Programming languages based on the CCP model, provide a *propagator* for each user-defined constraint.

*Propagators* can be seen as operators reducing the set of possible values for some variables. For instance, in a system with variables and taking Musical Instrument Digital Interface (MIDI) values, the constraint specifies possible values for and (where is at least one tone higher than ). In MIDI notation, each MIDI pitch unit represents a semi-tone.

The CCP model includes a set of constraints and a relation to know when a constraint can be deduced from others (named entailment relation $\models$). This relation gives a way of deducing a constraint from the information supplied by other constraints.

"The idea of the CCP model is to accumulate information in a *store*. The information on the *store* can increase but it cannot decrease. Concurrent processes interact with the *store* by either adding more information or by asking if some constraint can be deduced from the current *store*. If the constraint cannot be deduced, this process blocks until there is enough information to deduce the constraint" [40].

Consider for instance four agents interacting concurrently (fig. 1). The processes **tell** and **tell** add new information to the *store*. The processes **ask do** *P* and **ask do** *Q* launch process *P* and *Q* (*P* and *Q* can be any process) respectively, when their condition can be entailed from the *store*. After the execution of the **tell** processes, process **ask do** *P* launches process *P*, but the process **ask do** *Q* will be suspended until its condition can be entailed from the *store*.

0.6

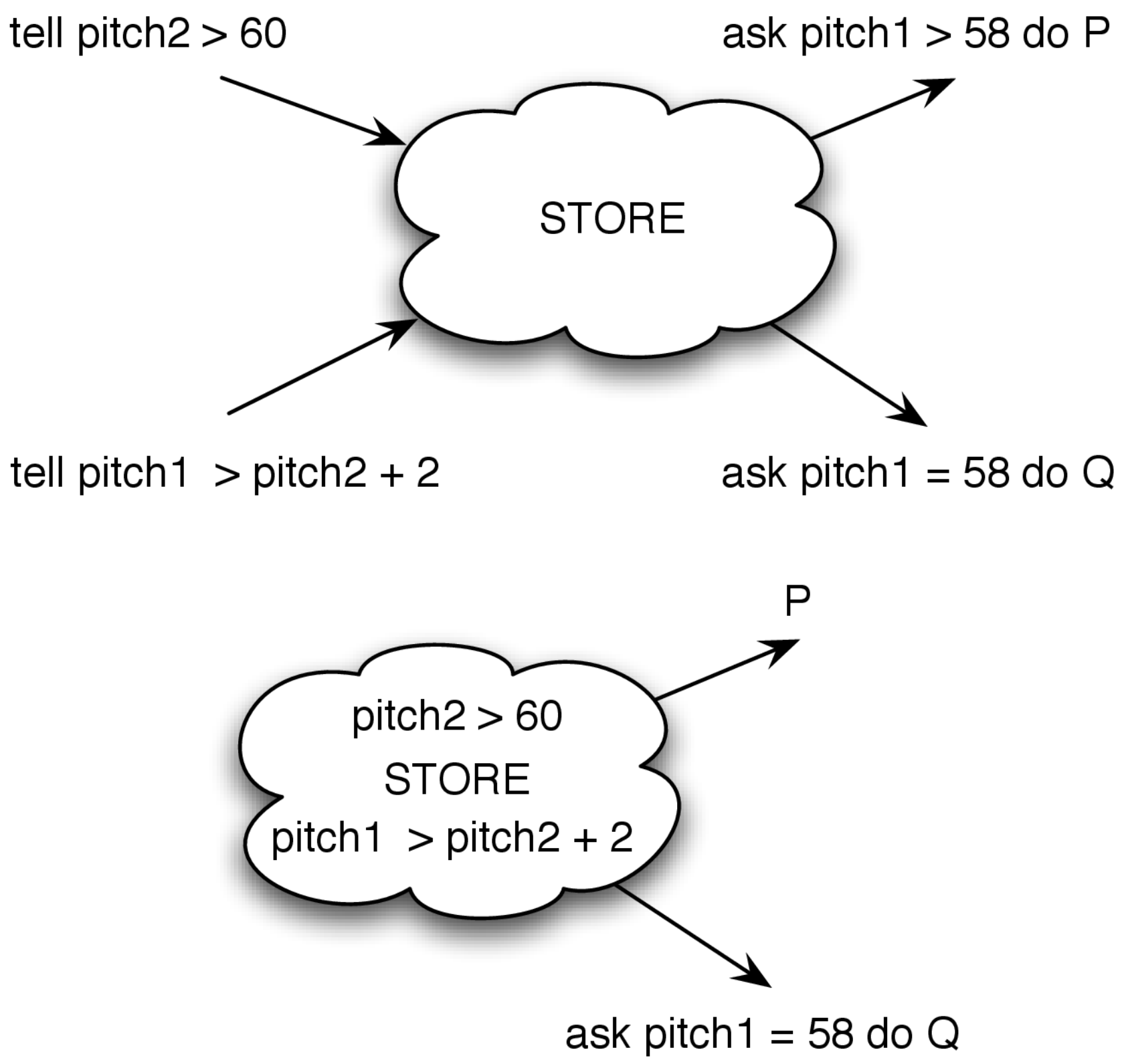

0.6

Figure 1: Process interaction in CCP

Formally, the CCP model is based on the idea of a constraint system. "A constraint system is a structure $<D,\vdash,Var>$ where D is a (countable) set of primitive constraints (or tokens), $\vdash \in D \times D$ is an inference relation (logical entailment) that relates tokens to tokens and *Var* is an infinite set of variables" [44]. A (non primitive) constraint can be composed out of primitive constraints.

The formal definition of CCP does not specify which types of constraints can be used. Thus, a constraint system can be adapted to a particular need depending on the set *D*. For instance, *finite domain (FD)* constraint system provides primitive constraints (also called *basic constraints*) such as $x \in R$, where *R* is a set of ranges of integers. On the other hand, *finite set (FS)* constraint system provides primitive constraints such as $y \in S$, where *S* is a set of FD variables and *y* is an FD variable. Constraints systems may also include expressions over trees, graphs, and sets.

Valencia and Rueda argue in [43] that the CCP model posses difficulties for modeling reactive systems where information on a given variable changes depending on the interactions of a system with its environment. The problem arises because information can only be added to the *store*, not deleted nor changed.

## 2.2 Non-deterministic Timed Concurrent Constraint (`ntcc`)

`Ntcc` introduces the notion of discrete time as a sequence of *time-units*. Each *time-unit* starts with a store (possibly empty) supplied by the environment, then `ntcc` executes all processes scheduled for that *time-unit*. In contrast to CCP, in `ntcc` variables, changing values along time can be modeled. In `ntcc` we can have a variable *x* taking different values along *time-units*. To model that in CCP, we would have to create a new variable each time we change the value of *x*.

Following, we give some examples of how the computational agents of `ntcc` can be used. The operational semantic of all `ntcc` agents can be found in Appendix 9.5 and a summary can be found in table 1. Using the **tell** agent with a FD constraint system, it is possible to add constraints such as (meaning the must be equal to 60) or (meaning that is an integer between 60 and 100).

The **when** agent can be used to describe how the system reacts to different events, for instance **when do tell**(*CMayor=true*) is a process reacting as soon as the pitch sequence C, E, G (represented as 48, 52, 55 in MIDI notation) has been played, adding the constraint *CMayor=true* to the *store* in the current *time-unit*.

| Agent | Meaning |
|---|---|
| **tell** (*c*) | Adds the constraint c to the current *store* |
| **when** (*c*) **do** *A* | If *c* holds now run *A* |
| **local** (*x*) **in** *P* | Runs *P* with local variable *x* |
| *A* \| *B* | Parallel composition |
| **next** *A* | Runs *A* at the next *time-unit* |
| **unless** (*c*) **next** *A* | Unless *c* can be inferred now, run *A* |
| **when do** | Non deterministically chooses s.t. holds |
| **P* | Delays P indefinitely (not forever) |
| !*P* | Executes P each *time-unit* (from now) |

Table 1: `Ntcc` Agents

**Parallel composition** allows us to represent concurrent processes, for instance **tell** | **when do tell** (*Instrument*=1) is a process telling the *store* that is 62 and concurrently reacts when is in the octave -1, assigning *instrument* to 1 (fig. 2). The number one represents the acoustic piano in MIDI notation.

The **next** agent is useful when we want to model variables changing through time, for instance **when  do next tell** , means that if  is equal to 60 in the current *time-unit*, it will be different from 60 in the next *time-unit*.

0.6

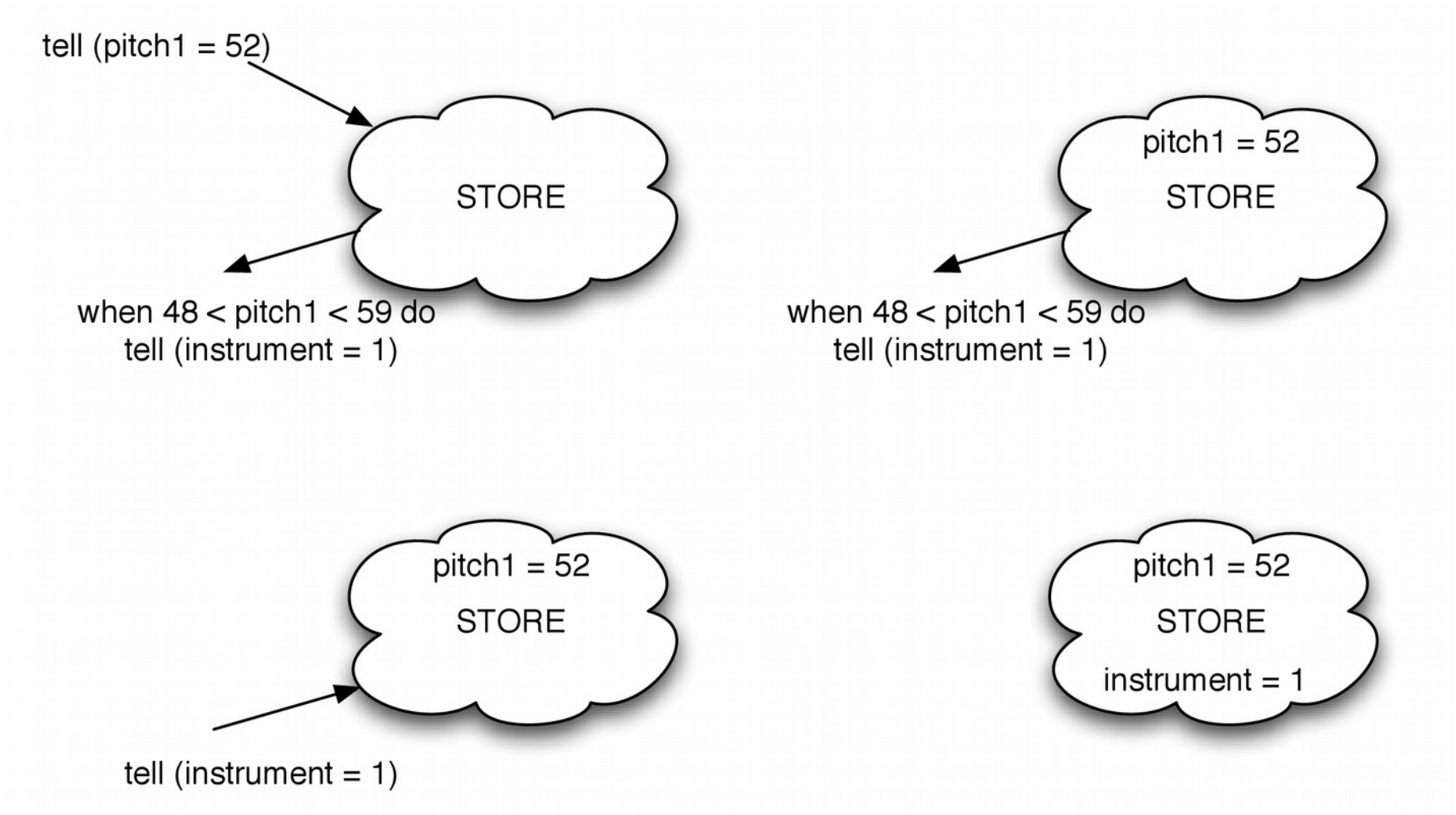

Figure 2: **Tell**, **when**, and **parallel** agents in *Ntcc*

The **unless** agent is useful to model systems reacting when a condition is not satisfied or it cannot be deduced from the *store*. For instance, **unless  next tell** (*lastpitch*<>60), reacts when  is false or when  cannot be deduced from the *store* (i.e., was not played in the current *time-unit*), telling the *store* in the next *time-unit* that *lastpitch* is not 60 ( fig. 3).

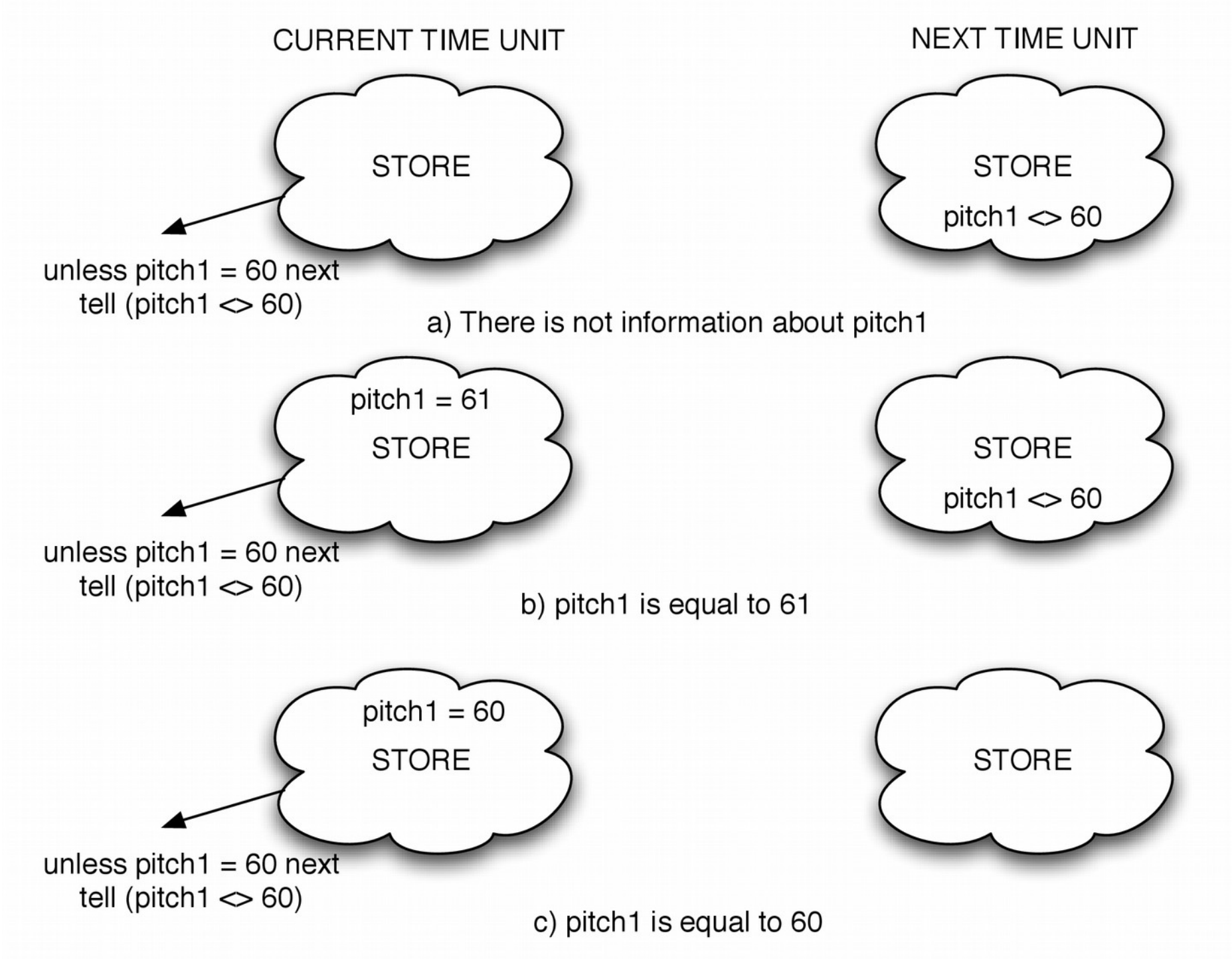

Figure 3: Unless agent in `ntcc`

The ***** agent may be used in music to delay the end of a music process indefinitely, but not forever (i.e., we know that the process will be executed, but we do not know when). For instance, ***tell** (*End*=*true*). The ! agent executes a certain process in every *time-unit* after its execution. For instance, !**tell** (*PlaySong*=*true*). The agent is used to model non-deterministic choices. For instance, ! **when** *true* **do tell** (*pitch*=*i*) models a system where each *time-unit*, a note is chosen from the C major chord (represented by the MIDI numbers 48,52 and 55) to be played (fig. 4[3]).

The agents presented in table 2 are derived from the basic operators. The agent *A* + *B* non-deterministically chooses to execute either *A* or *B*. The **persistent assignation** process changes the value of *x* to the current value of *t* in the following *time-units*. In a similar way, the agents in table 3 are used to model cells. Cells are variables which value can be re-assigned in terms of its previous value. *x*: (*z*) creates a new cell *x* with initial value *z*, changes the value of a cell (this is different from which changes the value of x only once), and exchanges the value of cell *x* and *z*.

3 ! **when** *true* **do tell** (*pitch*=*i*) can be expressed as !(**tell** (*pitch*=48) + **tell** (*pitch*=52) + **tell** (*pitch*=55))

0.6

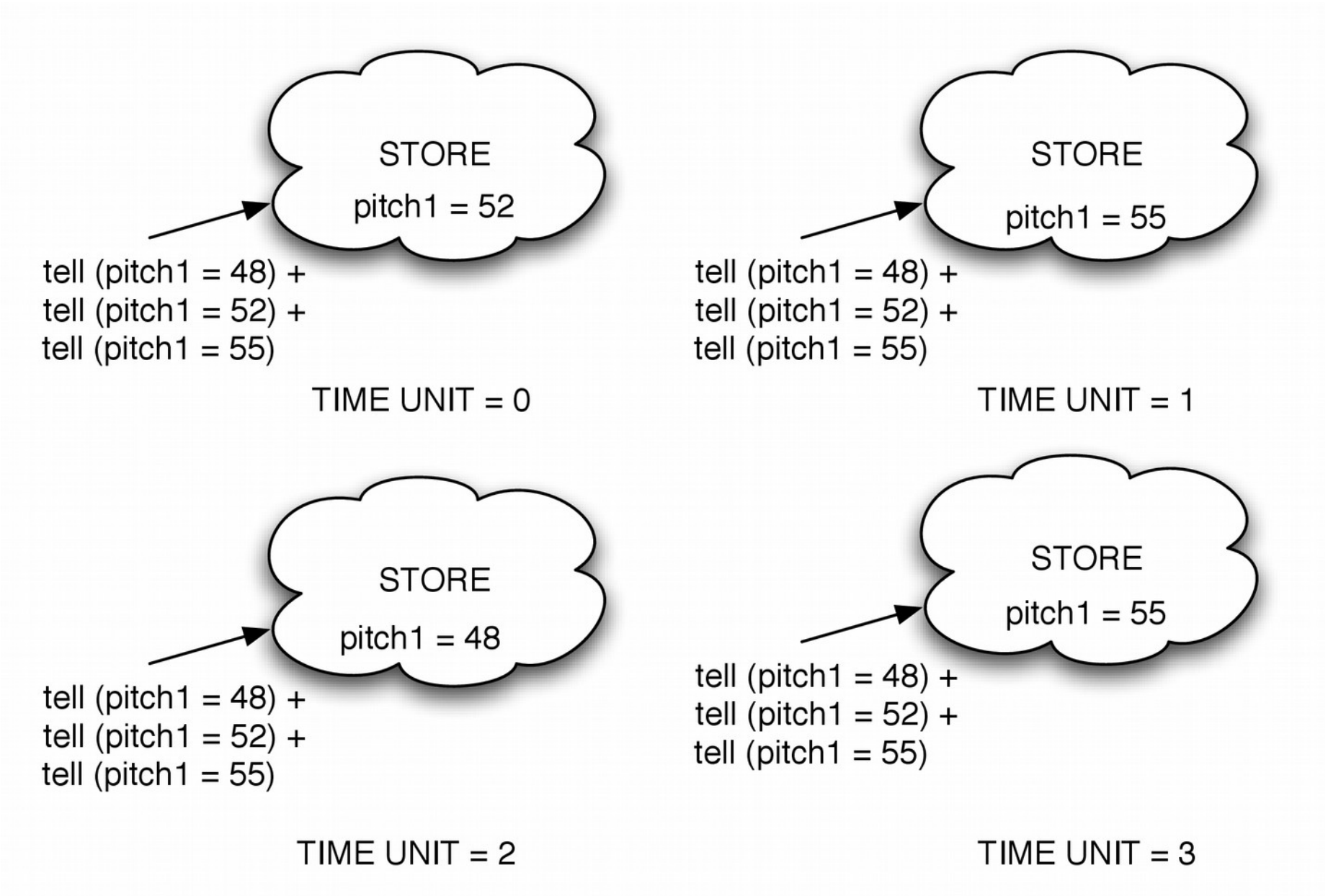

Figure 4: Execution of a non-deterministic process in `ntcc`

| Agent | Meaning |
|---|---|
| A + B | **when** *true* **do** (**when** *i*=1 **do** *A* \| **when** *i*=2 **do** *B* )<br>**local** *v* **in when** *t*=*v* **do next !tell** (*x*=*v*) |

Table 2: Derived `ntcc` agents

| Agent | Meaning |
|---|---|
| *x*: (*z*) | **tell**(*x*=*z*) \| **unless** change(*x*) **next** *x*: (z)<br>**local** *v* **when** *x*=*v* **do** (**tell** change(*x*) \| **next** *x*: *g*(*v*) )<br>**local** *v* **when** *t*=*v* **do** (**tell**(change(*x*) \| (**tell**(change(*y*)<br>\| **next** (*x*: *g*(*v*) \| *y*: (*v*)) |

Table 3: Definition of cells

Finally, a basic recursion can be defined in *ntcc* with the form , where *q* is the process name and is restricted to call *q* at most once and such call must be within the scope of a "next". The reason of using "next" is that we do not want an infinite recursion within a *time-unit*. Recursion is used to model iteration and recursive definitions. For instance, using this basic recursion, it is possible to write a function to compute the factorial function. Further information about recursion in `ntcc` can be found at [24].

## 2.3 Generic Constraint Development Environment (Gecode)

Gecode is a constraint solving library written in C++. Gecode is based on Constraints as Propagation agents (CPA) according to [39]. A CPA system provides multiple propagators to transform a (non-primitive) constraint into primitive constraints supplying the same information. In a finite domain constraint system, primitive constraints have the form $x \in [a..b]$. For instance, in a *store* containing , , a propagator would add constraints  and .

The reader may notice that there is a similarity between CPA and `ntcc`. Both of them are based on concurrent agents working over a constraint *store*. In chapter 6, we explain how we can encode `ntcc` agents as *propagators*.

*Gecode* works on different operating systems and is currently being used as the constraint library for Alice[36] and soon it will be used in Mozart-Oz, therefore it will be maintained for a long time. Furthermore, it provides an extensible API, allowing us to create new *propagators*. Finally, we conjecture that *Gecode*'s performance is better than the constraints solving tool-kits used in Sicstus Prolog and Mozart-Oz based on Gecode's benchmarks[4].

## 2.4 Factor Oracle (*FO*)

The Factor Oracle (*FO*)[1] is a finite automaton that can be built in linear time and space, in an incremental fashion. The *FO* recognizes at least all the sub-sequences (factors) of a given a sequence (it recognizes other sequences that are not factors). All the states of the *FO* are considered as accepting states. A sequence of symbols s =  is learned by such automaton, which states are 0,1,2...n. There is always a transition arrow (called *factor link*) from the state $i-1$ to the state $i$ and there are some transition arrows directed "backwards", going from state $i$ to $j$ (where $i>j$), called *suffix links*. *Suffix links*, opposed to *factor links*, are not labeled. For instance, a *FO* automaton for s = ab is presented in Figure 5, where black headed arrows represent the *factor links* and white headed arrows represent the *suffix links*.

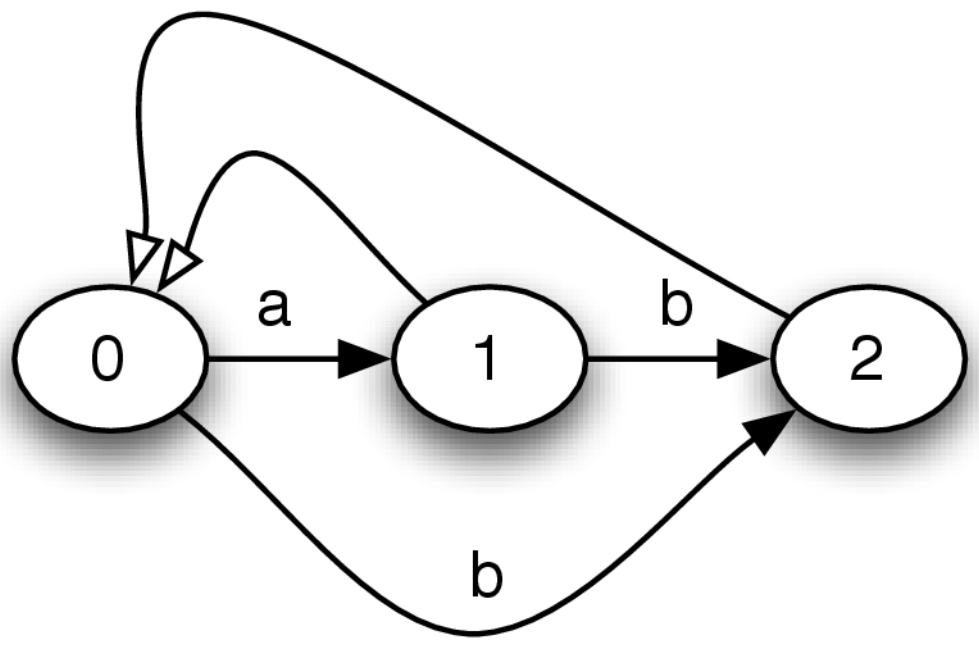

0.6

4 Benchmarks presented in http://www.gecode.org

Figure 5: A *FO* automaton for *s=ab*

The *FO* is built on-line and their authors proved that its algorithm has a linear complexity in time and space[1]. For each new entering symbol σ, a new state *i* is added and an arrow from *i*−1 to *i* is created with label . Starting from *i*−1, the *suffix links* are iteratively followed backward, until a state is reached where a *factor link* with label  going to some state *j*, or until there are no more *suffix links* to follow. For each state met during this iteration, a new *factor link* labeled by  is added from this state to *i*. Finally, a *suffix link* is added from *i* to state *j* or to state 0 depending on which condition terminated the iteration. Further formal definitions and the proof of *FO* complexity can be found in [1]. The on-line construction algorithm is presented with detail in Appendix 9.1

Since the *FO* has a linear complexity in time and space, it was found in [14] that it is appropriate for machine improvisation. In addition, all attribute values for a music event can be kept in an object attached to the corresponding node, since the actual information structure is given by the configuration of arrows (*factor and suffix links*). Therefore a tuple with *pitch* (the frecuency of the note), *duration* (the amount of time that the note is played), and *intensity* (the volume on which is the note is played) can be related to each arrow according to [14].

# 2.5 Concurrent Constraint Factor Oracle Model for Music Improvisation

Concurrent Constraint Factor Oracle Model for Music Improvisation (*Ccfomi*) is defined in [40]. Following, we present a briefly explanation of the model taken from [40]. *Ccfomi* has three kinds of variables to represent the partially built *FO* automaton: Variables  are the set of labels of all currently existing *factor links* going forward from *k*. Variables  are the *suffix links* from each state *i*, and variable  give the state reached from *k* by following a *factor link* labeled . For instance, the *FO* in figure 5 is represented by ,, , , , .

Although it is not stated explicitly in *Ccfomi*, the variables  and  are modeled as infinite rational trees [34] with unary branching, allowing us to add elements to them, each *time-unit*. Infinite rational trees have infinite size. However, they only contain a finite number of distinct sub-trees. For that reason, they have been subjects of multiple axiomatizations to construct a constraint system based on them. For instance, posting the constraints *cons*(*c*,*nil*,*B*), *cons*(*b*,*B*,*C*), *cons*(*a*,*C*,*D*) we can model a list of three elements [*a*,*b*,*c*].

*Ccfomi* is divided in three subsystems: learning (ADD), improvisation (CHOICE) and playing (PLAYER) running concurrently. In addition, there is a synchronization process (SYNC) that takes care of synchronization.

The *ADD* process is in charge of building the *FO* (this process models the learning phase) by creating the *factor links* and *suffix links*. Note that the process *ADD* calls the *LOOP* process.

 !**tell**() | ()

"Process  adds (if needed) *factor links* labeled  to state *i* from all states *k* reached from *i*−1 by *suffix links*, then computes , the *suffix link* from *i*" [40].

**when** *k*≥0 **do**
**unless**
**next**(!**tell** () | !**tell**() | )
| **when** *k*=−1 **do** !**tell**()

| **when**  **do** !**tell**()
"A musician is modeled as a *PLAYER* process playing some note $p$ every once in a while. The *PLAYER* process non-deterministically chooses between playing a note now or postponing the decision to the next *time-unit*" [40].

**when** true **do** (!**tell**() | **tell**($go=j$) | **next** )
+ (**tell** ($go=j-1$) | **next** )

The learning and the simulation phase must work concurrently. In order to achieve that, it is required that the simulation phase only takes place once the sub-graph is completely built. The  process is in charge of doing the synchronization between the simulation and the learning phase to preserve that property.

Synchronizing both phases is greatly simplified by the used of constraints. When a variable has no value, the *when* processes depending on it are blocked. Therefore, the  process is "waiting" until *go* is greater or equal than one. It means that the process has played the note $i$ and the  process can add a new symbol to the *FO*. The other condition  is because the first *suffix link* of the FO is equal to -1 and it cannot be followed in the simulation phase.

**when**  **do** ( | **next** )
| **unless**  **next**
"The improvisation process  uses the distribution function Φ : . The process starts from state $k$ and stochastically, chooses according to probability $q$, whether to output the symbol  or to follow a backward link "[40].

**when**  **do next**( **tell** () | )
| **tell** ()
| **when**  **do next** (**tell** () | )
| **unless**
**next**  **when**  **do** ( **tell** ($out$=σ) |

The whole system is represented by a process doing all the initializations and launching the processes when corresponding. Improvisation starts after $n$ symbols have been created by the *PLAYER* process.

!**tell**($q=p$) | !**tell**() |  |
| !**when** $go=n$ **do** *CHOICE*($n$)

## 2.6 Probabilistic Non-deterministic Timed Concurrent Constraint (pntcc)

"One possible critique to CCP is that it is too generic for representing certain complex systems. Even if counting with partial information is extremely valuable, we find that properly taking into account certain phenomena remains to be difficult, which severely affects both modeling and verification. Particularly challenging is the case of uncertain behavior. Indeed, the uncertainty underlying concurrent interactions in areas such as computer music goes way beyond of what can be modeled using partial information only." [29].

The first attempt to extend ntcc to work with probabilities was the *Stochastic Non-deterministic Timed Concurrent Constraint (sntcc [25])* calculus. Sntcc provides an operator  to decide whether to execute or not a process with a certain probability ϱ. Using `sntcc`, *Ccfomi* models the action of choosing between a *suffix link* or a *factor link* with a probability ϱ. However, when using `sntcc`, it is not possible to use a probability distribution to choose among all the *factor links* following a state in the *FO*. The probability distribution describes the range of possible values that a random variable can take.

`Pntcc` overcomes that problem, it provides a new agent to the calculus for probabilistic choice $\bigoplus$. The probabilistic choice $\bigoplus$ operator has the following syntax:

**when**  **do** (),

where *I* is a finite set of indexes, and for every  we have .

"The intuition of this operator is as follows. Each  associated to  represents its probability of being selected for execution. Hence, the collection of all  represents a probability distribution. The guards that can be entailed from the current *store* determine a subset of enabled processes, which are used to determine an eventual normalization of the 's. In the current time interval, the summation probabilistically chooses one of the enabled process according to the distribution defined by the (possibly normalized) 's. The chosen alternative, if any, precludes the others. If no choice is possible then the summation is precluded." [29].

Using the probabilistic choice we can model a process choosing a *factor link* from the *FO* with a probability distribution ϱ.

**when**  **do** (**tell**(*output*=σ),)

The operational semantic of the $\bigoplus$ agent and other formal definitions about `pntcc` can be found in Appendix 9.5.

# 3 Probabilistic Choice of Musical Sequences

When modeling machine improvisation, we want to choose a certain music sequence, based on the history of user and computer interaction. For instance, when traversing the Factor Oracle (*FO*) in the simulation phase, we want some information to choose among the *factor links* and the *suffix link* following a certain state. To achieve that, we propose to assign integers to the links in the *FO*. Using those integers, we can calculate probabilities to choose a link based on a probability distribution. We recall from the introduction that our main objective is extending *Ccfomi* to model probabilistic choice of musical sequences.

In the beginning of this thesis work, we developed a probabilistic model which changes the complexity in time for building the *FO* to quadratic (see Appendix 9.2.1). The idea behind it was changing the probabilities of all the *factor links* coming from state *i* when modifying a *factor link* leaving from that state. This idea was discarded for not being compatible with soft real-time (consider soft real-time as defined in the introduction).

The probabilistic model we chose is based on a simple, yet powerful concept. Using the system parameters, the probability of choosing a *factor link* in the simulation phase will decrease each time a *factor link* is chosen. Additionally, we calculate the length of the common suffix (*context*) associated to each *suffix link*. Using the *context*, we reward the *suffix links*. Further information about the *context* can be found at [19].

We represent the system with four kind of variables used to represent: the *FO* states and transitions; the musical information attached to the *FO*; the probabilistic information; and the information to change musical attributes in the notes, based on the user style.

In addition to the variables described before, the system has some information parametrized by the user: α,β,γ,τ and *n*. The constant α is the recombination factor, representing the proportion of new sequences desired. β represents the factor for decreasing the importance of a *factor link* when it is chosen in the simulation phase. γ represents the importance of a new *factor link* in relation with the other *factor links* coming from the same state. τ (described in Chapter 4) is a parameter for changing musical attributes in the notes. Finally, *n* is a parameter representing the number of notes that must be learned before starting the simulation phase. In Chapter 5 we describe how can we use *n* to synchronize the improvisation phases.

We label each *factor link* by the *pitch*. Moreover, outside the *FO* definition, we create a tuple of three integers for each *factor link*: *pitch*, *duration*, and *intensity*. These three characteristics are represented by integers. The *pitch* and the *intensity* are represented by integers from 0 to 127 and the *duration* is represented by milliseconds[5]. That way we can build a *pitch FO* (i.e., a *FO* where the symbols are pitches) associating to it other musical information.

At the same time we build a *FO*, we also create three integer arrays: ϱ, *C* and *sum*. There is an integer  for every *factor link*,  for every *suffix link*, and  for every state *i*. Note that  would represent the probability of choosing a *factor link* if *suffix links* were not considered, and  is the *context*.

Next, we show the learning and simulation phases for the probabilistic extension. We present some simple examples explaining how the probabilities are calculated in the learning phase and how they are used in the simulation phase. Finally, we present some concluding remarks and other improvisation models related to our model.

## 3.1 Stylistic learning phase

During the learning phase we store an integer  for each *factor link* going from *i* labelled by σ. We also store an integer  for each state *i* of the automaton. The initial value for  is  (fig. 6), where γ is a system parameter representing the importance of a new sequence in relation with the sequences already learned. When a *factor link* from *i* labeled by  is the first *factor link* leaving from *i*, we assign to  and  the constant *c*. We want *c* to be a big integer, allowing us to have more precision when reasoning about .

The reader may notice that this approach gives a certain importance to a new *factor link* leaving from *i* labeled by , without changing the value of all the other quantities  leaving from *i*. Furthermore, we preserve the sum of all the values  in the variable , for each state *i*. This system exhibits a very important property: For each state *i*, . The sum of all the probabilities associated to the *factor links* coming from the same state are equal to one. This property is preserved, when changing the values of and  in both improvisation phases.

5 Pitch, duration and intensity are represented according to MIDI 1.0 standard

0.7

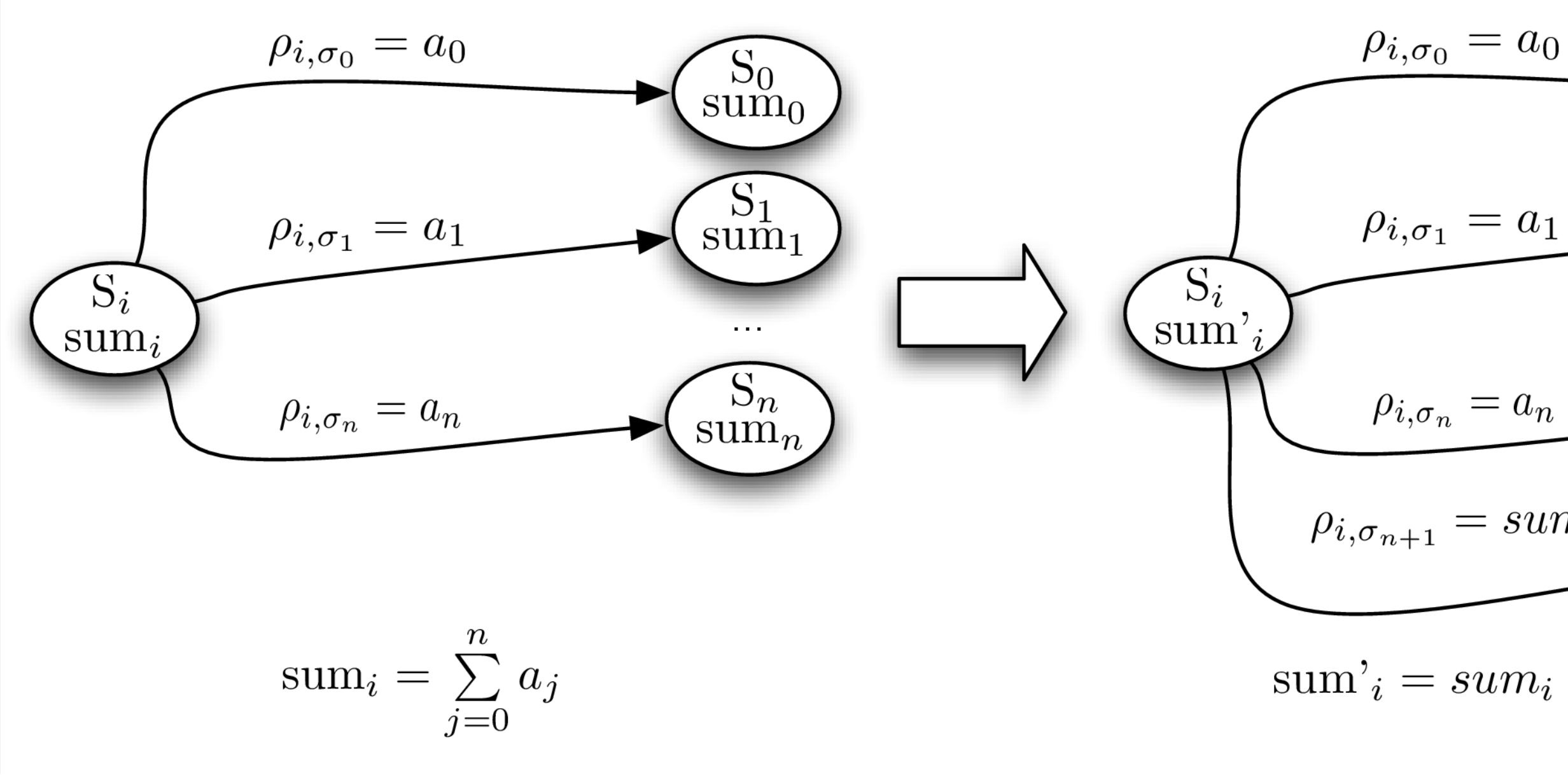

Figure 6: Adding a *factor link* to the *FO*

On the other hand, we give rewards to the *suffix link* using the *context*. To calculate the *context*, Lefebvre and Lecroq modified the *FO* construction algorithm, conserving its linear complexity in time and space [19]. This approach has been successfully used by Cont, Assayag and Dubnov on their anticipatory improvisation model [11].

Figure 7 is a simple example of a *FO* and the integer arrays presented previously. First, we present the score of a fragment of the Happy Birthday song; then we present a sequence of possible tuples <*pitch*, *duration*, *intensity*> for that fragment; and finally the *FO* with the probabilistic information.

0.8

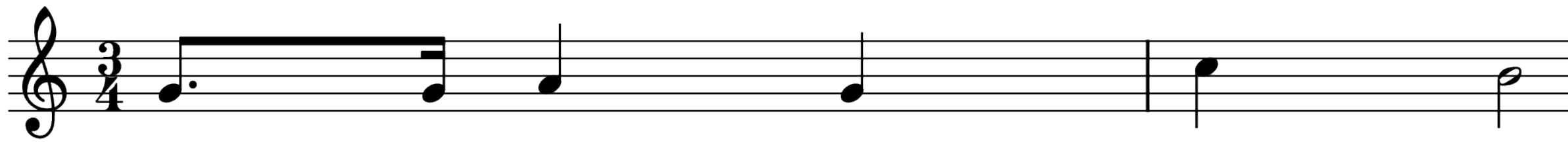

(a) The Score of the fragment

(G, 375,80), (G, 125,60), (A, 500,100), (G, 500,90), (C, 500,100), (B, 1000,60)

(b) Fragment of the Happy Birthday Song represented with tuples

0.7

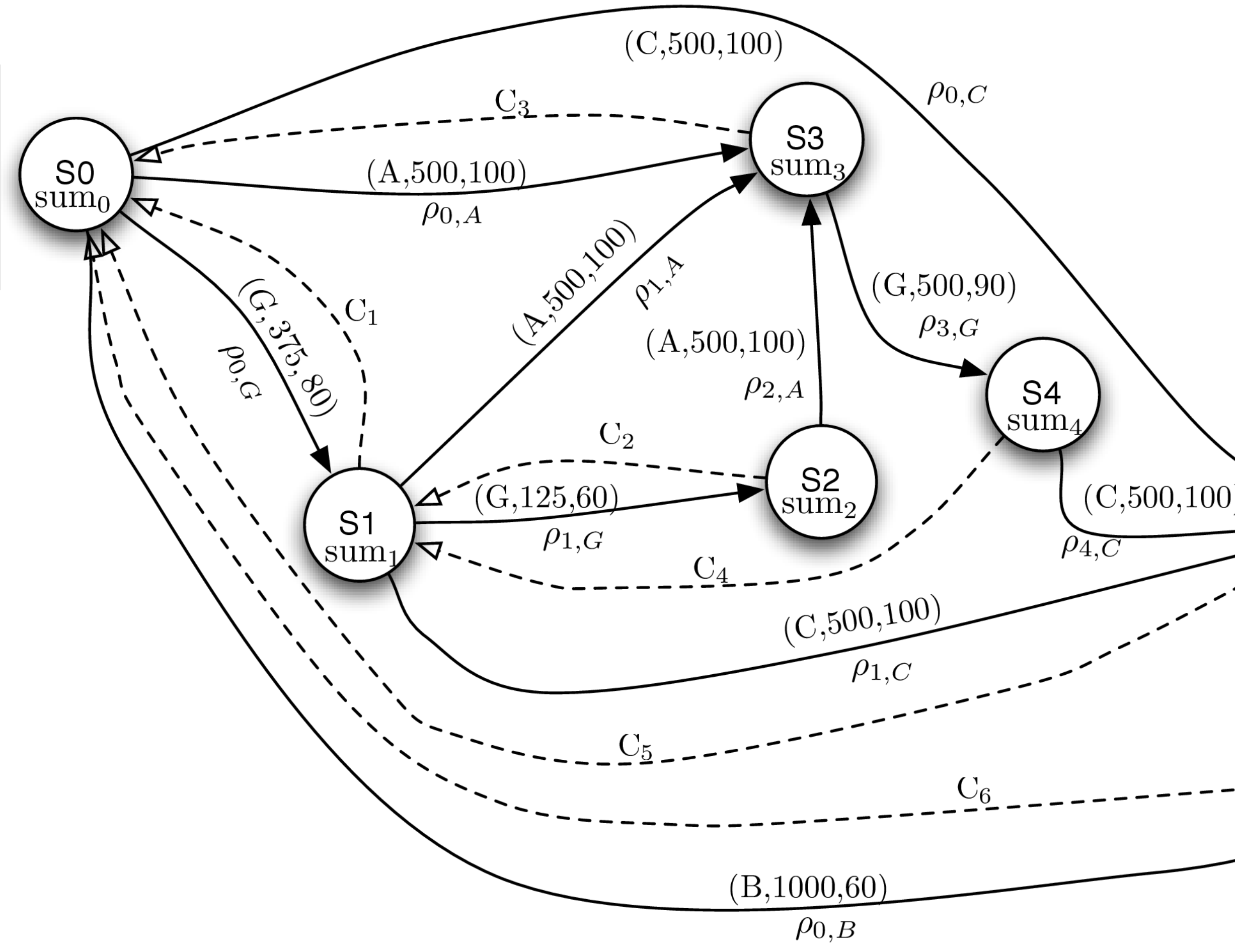

(c) Factor Oracle with the probabilistic information

Figure 7: Factor Oracle used to represent a Happy Birthday fragment

## 3.2 Stylistic simulation phase

In the simulation phase, we use all the information calculated in the learning phase to choose the notes probabilistically. *Factor links* chosen in this phase, will decrease the importance proportionally to β. In addition, the probability of choosing secondary *factor links* is proportional to γ . We consider primary *factor links* those going from the state $i$ to $i$+1, and all the others as secondary . On the other hand, the *suffix links* are rewarded by the *context*, calculated on-line in the learning phase.

If there were not *suffix links*, we would choose a *factor link* leaving from the state $i$ with a probability distribution $\phi(i,\sigma)$ such that . Later on, we will explain how we can extend this concept to work with the *suffix links*, rewarded by their *context*. However, the concept of decreasing the probability of a *factor link* when it is chosen remains invariant.

When the system chooses a certain *factor link* leaving from $i$ and labeled by , the value of  is decremented, multiplying it by β. Subsequently, we update the new value of  by subtracting  (fig. 8). That way, we preserve the property  for each state $i$. Note that we are only adding constant time operations, making our model compatible with soft real-time.

0.7

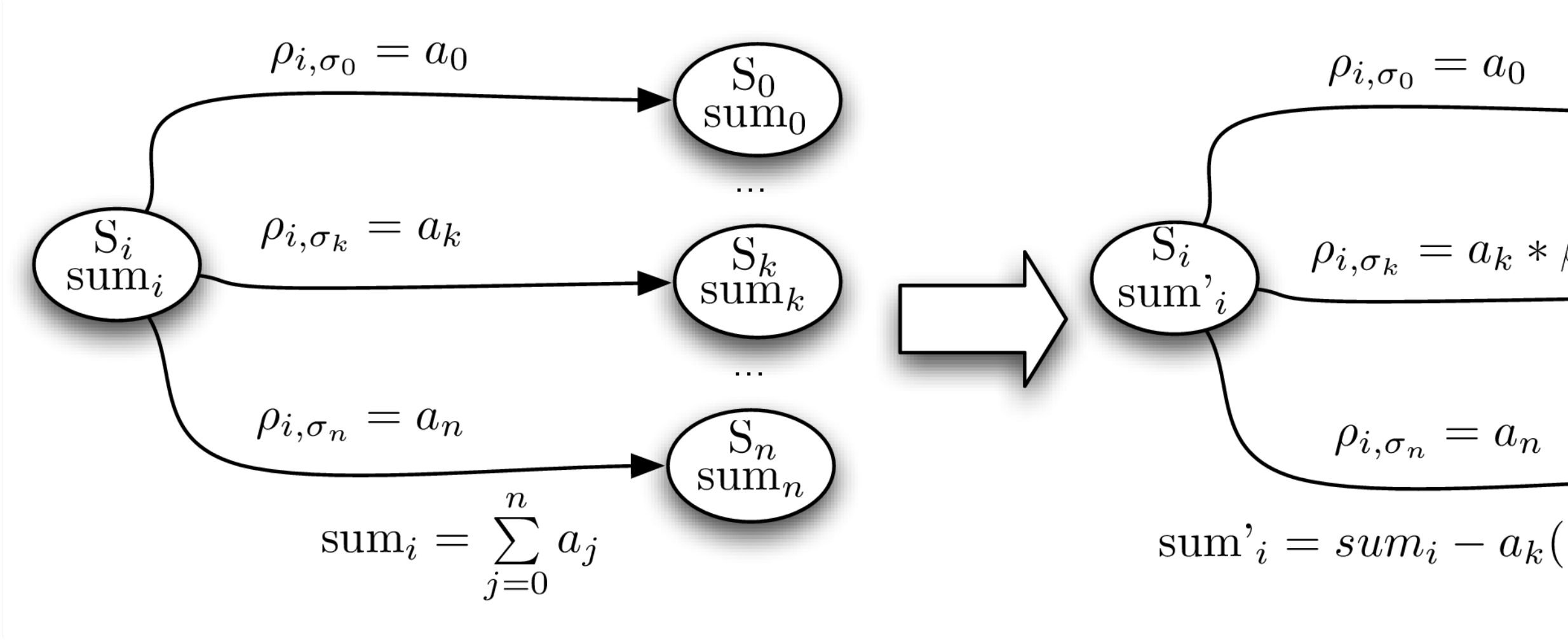

Figure 8: Choosing a *factor link* from $k$ labelled by

Following only the *factor links* we obtain all the factor (subsequences) of the original sequence. This causes two problems: first, if we always follow the *factor links*, soon we will get to the last state of the automaton; second, we only improvise over the subsequences of the information learned from the user, without sequence variation. This would make the improvisation repetitive. Following the *suffix link* we achieve sequence variation because we can combine different suffixes and prefixes of the sequences learned. For instance, in *Omax* [6] –a model for music improvisation processing in real-time audio and video– this is called recombination and it is parametrized by a recombination factor.

Rueda et al approaches this problem in *Ccfomi* by creating a probability distribution parameterized by a value α. The probability of choosing a *factor link* is given by α and the probability of choosing a *suffix link* is given by 1−α. There is a drawback in this approach. Since it does not reward the *suffix links* with the *context* (the length of the common suffix), this system may choose multiple times in a row *suffix links* going back one or two states, creating repetitive sequences.

Our approach is based on rewarding the *suffix links* by their *context*. The intuition is choosing between the *factor links* leaving a state $i$ and the *factor links* leaving the state reached by following the current state's *suffix link*. Rewarding the last ones by the product of the recombination factor α and the *context* . Consider $S(i)$ a function returning the state where a *suffix link* leads from a state *i*. If we only consider the *factor links*, we would have two probability distributions and and no way to relate them. Using the *context* , we create a probability distribution Φ(*i*,σ) ranking the *factor links* leaving from the state *S*(*i*) with the product *C*(*i*).

[Sorry. Ignored \begin{cases} ... \end{cases}]

Using Φ(*i*,σ), the system is able to choose a symbol at any state of the *FO*. The advantage of this probability distribution over the one presented in *Ccfomi*, is that it takes into account the *context*, as well as the recombination factor α.

To exemplify how to build this probability distribution, consider the *FO* with the probabilistic information in figure 9. That example correspond to the *FO* for *s*=*ab* and random values for the integer arrays described in this chapter. Table 4 shows how to build a probability distribution Φ(*i*,σ) for the *FO* in figure 9.

Note that for the states zero and two in the table, the probabilities calculated are the same. This happens because the first state does not have a *suffix link* to go backwards and the last state does not have *factor links* to go forward. On the other hand, the probabilities calculated for the state one combine the probability of choosing a *factor link* following state 1 or choosing the *suffix link* and then choosing a *factor link* from state zero.

0.7

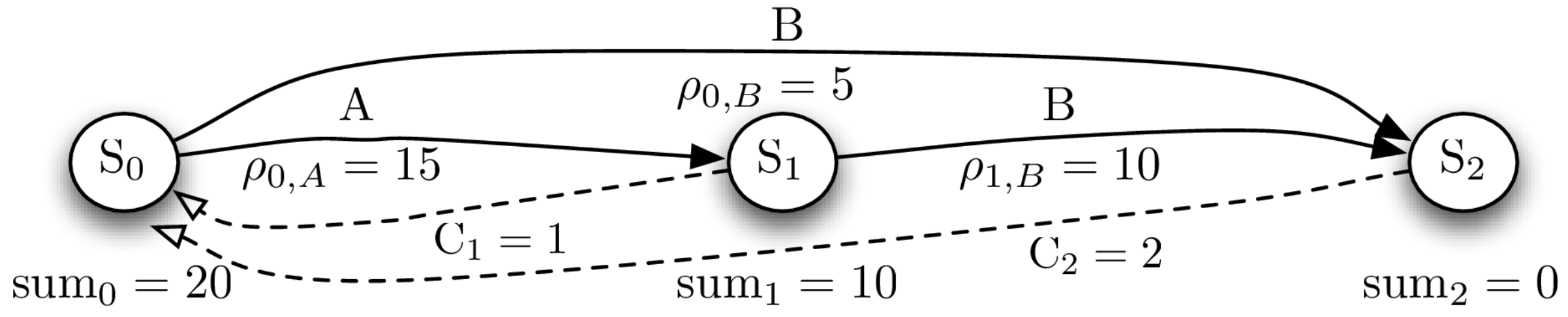

Figure 9: A Factor Oracle including probabilities, for the sequence *s*=*ab*

| i | σ | Φ(*i*,σ) | i | σ | Φ(*i*,σ) | i | σ | Φ(*i*,σ) |
|---|---|---|---|---|---|---|---|---|
| 0 | a | 3/4 | 1 | a | | 2 | a | 3/4 |
| 0 | b | 1/4 | 1 | b | | 2 | b | 1/4 |

Table 4: Probability distribution Φ(*i*,σ) for figure 9 9

## 3.3 Summary

In this chapter we explained how we can model music improvisation using probabilities, extending the notion of non-deterministic choice described in *Ccfomi*. The intuition is decreasing the probability of choosing a *factor link*, each time it is chosen and rewarding a *suffix link* based on the *context*. Furthermore, we explained how the parameters α, β, and γ allow us to parameterize the computation of the probabilities.

This procedure is simple enough so that the probabilities can be computed in constant time when the *FO* is built, preserving the linear complexity in time and space of the *FO* on-line construction algorithm. Additionally, using probabilities allows us to generate different sequences, without repeating the same sequence multiple times in a row like *Ccfomi*.

## 3.4 Related work

For *Omax*, Assayag and Blonch recently proposed a new way to traverse the oracle based on heuristics [5]. They argue that traversing the oracle using only the *suffix links* and not using the *factor links*, produces more "interesting" sequences.

There is an extension of *Ccfomi* using `pntcc`. The use of `pntcc` makes possible to choose the sequences in the simulation phase, based on a probability distribution. Although Perez and Rueda modeled the probabilistic choice of sequences using the *FO*, they do not provide a description of how those probabilities can be calculated during the learning phase.

# 4 Changing Musical Attributes of the Notes

According to Conklin [8], music-generation systems aim to create music based on some predefined rules and a *corpus* (i.e., a collection of musical pieces in a certain music style) learned previously. Those systems can create new musical material based on the style of the *corpus* learned. Unfortunately, they use algorithms with high complexity in time and space, making them inappropriate for music interaction according to [14]. On the other hand, interactive systems for music improvisation (e.g., *Ccfomi*) are usually based on the recombination of sequences learned from the user.

Although recombination creates new sequences based on the user style, it does not create new notes. In fact, it does not even change a single characteristic of a note. To solve that problem, one of the objectives of this thesis work is changing at least one musical attribute of the notes generated during the style simulation.

In the beginning of this work, we tried to develop an algorithm for creating new notes, based on the learned style. The idea was calculating the probability of being on a certain music scale. Based on that probability, we choose a random pitch from that scale. A music scale is an ascending or descending series of notes or pitches. We also developed an algorithm to calculate the duration of those new notes (see Appendix 9.2.5).

We did not include those ideas in this thesis work. First, because choosing a *pitch* based on a supposition of the scale cannot be generalized to music which is not based on scales. In addition, because the procedure for calculating the probability of being on a certain scale was not very accurate, as we found out during some tests. Finally, because the algorithm to generate new durations is not compatible with soft real-time.

The approach we chose to change a musical attribute is again based on simple, but powerful concept. We store the average *intensity* (the other musical attributes are not changed in our model for the reasons mentioned above) of the notes currently being played (*current dynamics*) by the computer. We also store the *current dynamics* of the user. Then, we compare them and change the *current dynamics* of the computer (if necessary), making it similar to the user *current dynamics*. The idea behind this *intensity* variation was originally proposed by musicians Riascos [35] and Collazos [7]. It is based on a concept that they usually apply when improvising with other musicians.

In order to formalize that concept, we calculate, in the learning phase, the *current dynamics* of the last τ (a system parameter) notes played by both, the user and the computer, separately. Concurrently, in the simulation phase, we compare the two *current dynamics*. If they are not equal, we multiply the intensity of the current note being played by the computer by a factor proportional to relation of the user and computer *current dynamics*. As follows, we explain in detail how we can calculate the *current dynamics* in the learning phase and how to change the *intensity* of notes generated in simulation phase.

## 4.1 Stylistic learning phase

The *intensity* in music represents two different things at the same time. When analyzing the *intensity* of a single note in a sequence, we reason about that *intensity* as a musical accent meaning the importance of certain notes or defining rhythms. On the other hand, we reason about the *average intensity* of a sequence of notes as the dynamics of that sequence of notes. The accents may be written explicitly in the score with a symbol  bellow the note and the dynamics for relative loudness may be written explicitly in the score as piano (*p*), forte (*f*), fortissimo (*ff*), etc.

To capture these two concepts, in the learning phase we store the *intensity* in a tuple <*pitch*,*duration*,*intensity*>. In addition, we store the *current dynamics* for the last τ notes played by the user  and the computer .

To calculate the *current dynamic* we propose the Calculate-Current-Dynamics algorithm. The idea of this algorithm is storing the last τ intensities in a queue . This algorithm receives a sequence of intensities *I*, the value for τ, a reference to the queue , and the *current dynamic* . The invariant of the algorithm is always having the average of the queue data in the variable  and the sum in the variable *IntensitySum*. Append 9.1.3 gives an example of the operation of this algorithm.

**CALCULATE-CURRENT-DYNAMICS**(, τ, , )

```
 new Queue(τ)
IntensitySum  0
QueueSize  0
for i  0 to m do
if QueueSize<τ then
IntensitySum
else IntensitySum  IntensitySum +  -
.push()
 IntensitySum/QueueSize
```

## 4.2 Stylistic simulation phase

In this phase, we traverse the *FO* using the probabilistic distribution Φ(*i*,σ) proposed in chapter 3. Remember that there is an *intensity* and a *duration* associated to each *pitch* in the *FO*. If we play the *intensities* with the same value as they were learned, we could have a problem of coherence between the *current dynamics* of the user and the *current dynamics* of the sequences we are producing.

To give an example of this problem, consider the Happy Birthday fragment presented in figure 7. The *current dynamics* for that fragment is 98. Now, suppose the computer *current dynamics* is 30. This poses a problem, because the user is expecting the computer to improvise in the same *dynamics* that he is, according to the interviews with Riascos and Collazos.

The solution we propose is multiplying by a factor  the intensity of every note generated by the computer. In the previous example, the next note generated by the computer would be multiplied by a factor of 30/98.

## 4.3 Summary

We explained how we can change the *intensity* of the notes generated during the improvisation. The idea is to maintain the *current dynamics* of the notes generated by the computer similar the *current dynamics* of the notes generated by the user. This corresponds to formalizing an improvisation technique used by two musicians interviewed for this thesis work.

This kind of variation in the *intensity* is something new for machine improvisation systems as far as we know. We believe that this kind of approach, where simple variations can be made preserving the style learned from the user and being compatible with real-time, should be a topic of investigation in future works.

## 4.4 Related work

To solve this problem of creating new notes and changing the attributes of the notes during the improvisation, the *Omax* model has a parameter called *innovation rate*, indicating the amount of new material desired [6]. Furthermore, Omax calculates a *rhythmic quality function* to compare the density (the number of events for overall duration) between the current state and the place where a link is leading.

Using that *rhythmic quality function*, the improvisation does not "jump" abruptly between different rhythmic patterns. Therefore, *Omax* improvisation is rhythmically coherent within itself. However, generating new rhythms coherent with the user style on machine improvisation is still an open problem.

The anticipatory model developed by Cont et al [10] presents some results where the sequences produced in the improvisation have different pitches, compared to the original sequence. To achieve this, they improvise on a *pitch intervals FO* (a *FO* learning the intervals of the pitches played by the user), allowing them to calculate new pitches, when using the *pitch intervals* attribute to improvise.

Neither *Ccfomi* nor its probabilistic extension provides a way to change musical attributes of the notes nor creating new material based on the user style.

# 5 Modeling the system in `pntcc`

`Ntcc` has been used in a large variety of situations for synchronizing musical processes. From the introduction chapter, we recall the models for interactive scores, audio processing, formalizing musical processes, and music improvisation. In those models, the synchronization is made declaratively. It means that `ntcc` hides the details on how the processes are synchronized and how the shared resources (in the *store*) are accessed. One objective of this work is modeling our improvisation system with `ntcc`. So far, we presented the modifications for the improvisation phases allowing probabilistic choice of musical sequences and changing the musical attributes in the simulation phase. Since we are choosing the sequences probabilistically, we use `pntcc` (the probabilistic extension of `ntcc`) for modeling our improvisation system.

In order to synchronize the improvisation phases, the learning phase must take place from the beginning. However, the simulation phase is launched once the learning phase has learned $n$ notes. After that, both phases run concurrently. Synchronization must be provided because the improvisation phase must not work in partially built graphs, it can only improvise in the fragment of the graph that represents a *FO*. Additionally, the simulation phase can only work in state $k$ once the value for the *current dynamics*, the *context*, and the *probabilistic distribution* has been calculated up to state $k$.

Our approach to synchronize the improvisation phases is similar to the one used in *Ccfomi*. Remember that *Ccfomi* synchronizes the improvisation phases using a variable *go* and the variables . The *PLAYER* process can post constraints over those variables and the processes for building the *FO* (*ADD* and *LOOP*) are activated when they can deduce certain information from those variables. We extend that concept using some of the new variables introduced in this model.

In addition to the variables , , and used in *Ccfomi*, our model has a few more variables: , , , and represent the probabilistic choice of musical sequences; and represent the musical attributes associated to each pitch σ; and , and represent the intensity variation. The variables , , , , and are represented with rational trees of *FD* variables because they do not change their value from a *time-unit* to another. The other variables are represented with cells (cells are defined in chapter 2).

In this chapter, we explain how we can write a sequential algorithm for the learning phase combining the algorithm for building on-line the *FO*, calculating the *context*, calculating the probabilistic distribution and the *current dynamics*. After that, we show how both phases can be modeled in `pntcc`. Finally, we give some concluding remarks and we present related work.

# 5.1 Modeling the stylistic learning phase

The learning phase can be easily integrated to the on-line algorithm that builds a *FO* and calculates the *context* (the original algorithms are presented in Appendix 9.1). The learning phase is represented by the functions *Ext_Oracle_On-line* and *Ext_Add_Letter*. To calculate the *context* we use the *Length_Repeated_Suffix* function proposed by Lefevre et al. The *Length_Repeated_Suffix* calculates the *context*. It finds the length of a repeated suffix of *P*[1..*i*+1] in linear time and space complexity.

The *Ext_Add_Letter* function is in charge of adding new pitches to the *FO*. It also creates a tuple <*pitch*,*duration*,*intensity*>; updates values of and ; and calculates the *current dynamics* of the user , and the *context* for state *i*+1. This function receives a *FO* with *i* states, a pitch σ, the duration, the intensity, the system parameters γ and τ, and the Intensity Queue . During its execution, it uses the constant *c*, the function *S*(*i*), and the temporal variable π. *C* is a big integer constant, *S*(*i*) is a function returning the *suffix link* for state *i*, and π is a temporal variable used to calculate the *context*.

```
EXT_ADD_LETTER(Oracle(P[1..i]),σ,duration,intensity,γ,τ,)
Create a new state i+1
if QueueSize<τ then
intensitySum  IntensitySum+intensity
else IntensitySum  intensity -
.push()
 IntensitySum/QueueSize
 duration
 intensity
While k>−1 and δ(k,σ) is undefined do
if k=−1 then
else
```

**Return** Oracle($P$[1..$i$]σ)

The *Ext_Oracle_On-line* function is a sequential algorithm representing the learning phase. It receives three vectors: the pitches, the durations, and the intensities. In addition, it takes γ, the system parameter for ranking the importance of a new note added to the *FO*, and the system parameter τ, representing the number of notes taken into account to calculate the *current dynamics* . Figure 10 presents the execution of this function for the three first symbols of the Happy Birthday Fragment presented in figure 7.

**EXT_ORACLE_ON_LINE**($P$[1..$m$],$D$[1..$m$],$I$[1..$m$],γ,τ)

Create Oracle(ε) with one single state 0 and $S(0)=-1$

*new Queue*(τ)

*IntensitySum* 0

**for** $i$ 1 **to** $m$ **do** *Oracle*([1..$i$])

**EXT_ADD_LETTER**(Oracle($P$[1..$i$−1]),,,,γ,τ,)

0.8

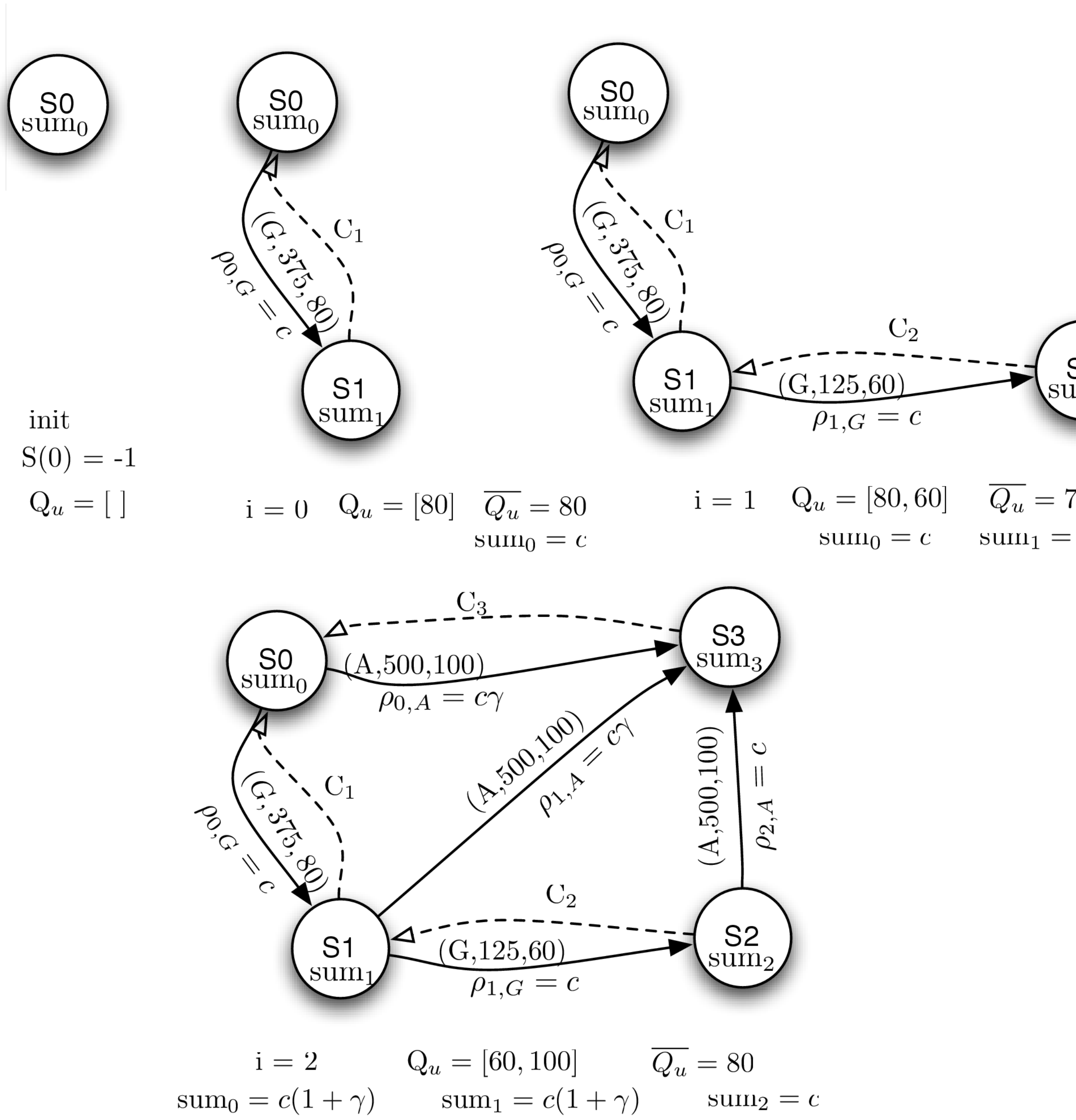

Figure 10: Executing the *Ext Oracle On-line* algorithm with τ=2

The learning phase is modeled in `pntcc` by the processes *PHI*, *ADD*, *LOOP*, , *PLAYER*, *CONTEXT*, and . Process *PHI* calculates the values for the probability distribution , used to choose A *factor link* leaving from state *k* labeled by a symbol σ. Where the recombination factor is parameterized by α. The process represents the act of adding a "fresh" variable to the infinite rational tree (as described in chapter 2). We use infinite rational trees to represent the variable such as *from* and δ that represents the transitions of the *FO*.

*PHI*(*k*,σ,α)

**when** **do** **!tell** ()
|**when** **do** **!tell** () |
The Process *ADD* is the one in charge of adding new pitches to the *FO*. In addition, this process updates the values of the cells ϱ and the variable φ calling the function *PHI*.

| **!tell** () | (*c*) | (*c*)
| | | | | | **next** ()
The *LOOP* process represents the “while” loop in the *Ext_Add_Letter* function. This process adds a new *factor link* in the *FO* that points to the new state *i*, while *k* is greater than -1 and there is not a transition from *k* labeled by σ. The values for *k* depends on the *suffix links*. In addition, it calculates the values for the *context* and the probabilistic information.

**when** $k \geq 0$ **do**(
**when** **do**
(**!tell** () | | )
|**unless** **next** (
| || | |
|| **next** ( **!tell** () || **!tell** ()
|| | | )))
|**when** $k = -1$ **do** ( **!tell** () | | )
In the *CONTEXT* process the reader may notice how we can use **when** $a \neq b$ **do** *P* instead of **unless** $a \neq b$ **next** *P* because we know that *a*,*b* always have a value. The values π, *s*, and are used to calculate efficiently the *context* according to Lefevre et al.’s algorithm.

*CONTEXT*(*i*,π,*s*)
**when** $s=0$ **do** **!tell** ()
|**when** $s \neq 0$ **do** (
**when** **do** **!tell** ()
|**when** **do** )

**when** **do** **!tell** ()
|**when** **do** **next**
The process calculates the value for the *current dynamics*. In addition, it updates *sum* based on the parameter τ.

**when** $index \geq \tau$ **do**
|**when** $index < \tau$ **do** |
Finally, the *PLAYER* stores the values of *pitch*, *duration*, and *intensity* received from the environment when a note is played by the user. Furthermore, it updates the *current dynamics* .

**when** $P>0 \wedge D>0 \wedge I>0$ **do** (

| **next** ( **!tell** () | | **!tell**
| **!tell** () | **!tell** () | **tell** ($go=i$) | ))
|**unless** $P>0 \wedge D>0 \wedge I>0$ **next** (**tell** ($go=j-1$) | )

## 5.2 Modeling the style simulation phase

In this phase, we use the ⨁ agent, defined in `pntcc` to model probabilistic choice. This model is an extension of the model presented in [29]. In our model, the *IMPROV* process chooses a link according to the probability distribution . Furthermore, it updates the values for *sum* and ϱ, sets-up the outputs, and updates the computer *current dynamics* .

In order to ask if a constraint $A \wedge B$ or $A \vee B$ can be deduced from the *store*, we use reification. For instance, the process **when** $a=b \wedge c=d$ **do** P, can be codified as the process **when** *g* **do** P and the constraints , , .

*IMPROV*($k$,τ,β)
|| **when** **do** (
**when** **do** (
**next**( **tell** () || **tell** ()
|| **tell** () || **tell** ()
|| )
|| **when** **do next** (*IMPROV*($k$+1,τ,β) + )
|| **unless** **next** (
|| || )
|| **unless** **next** ( *IMPROV*($k$+1,τ,β)
|| || ),)
|| **unless** **next** *IMPROV*($k$,τ,β)

## 5.3 Synchronizing the improvisation phases

Synchronizing both phases is greatly simplified by the used of constraints. When a variable has no value, *when* processes depending on it are blocked. Therefore, the process is "waiting" until *go* is greater or equal than one. That means that the process has played the note *i* and the process can add a new symbol to the FO. The other condition is because the first *suffix link* of the FO is equal -1 and that *suffix link* cannot be followed in the simulation phase. In addition, the *SYNC* process is also "waiting" for the *current dynamics* to take a value greater of equal than 0.

**when** **do** ( | **next** )
||**unless** **next**
A process is necessary to wait until *n* symbols have been learned to launch the *IMPROV* process.
*WAIT*($n$,τ,β)
**when** $go=n$ **do next** *IMPROV*($n$,τ,β) || **unless** $go=n$ **next** *WAIT*($n$,τ,β)

The system is modeled as the *PLAYER* and the *SYNC* process running in parallel with a process waiting until *n* symbols have been played to start the *IMPROV* process. The reader should remeber that α is the recombination factor, representing the proportion of new sequences desired. β represents the factor for decreasing the importance of a *factor link* when it is chosen in the simulation phase. γ represents the importance of a new *factor link* in relation with the other *factor links* coming from the same state. τ is a parameter for changing musical attributes in the notes. Finally, *n* is a parameter representing the number of notes that must be learned before starting the simulation phase.
*SYSTEM*($n$,α,β,γ,τ)
!**tell** () || || || *WAIT*($n$,τ,β)

## 5.4 Summary

We modeled all the concepts described in previous chapters using `pntcc`. Although synchronization and probabilistic choice are modeled declaratively, matching the *time-units* is not an easy task because the value of a cell only can be changed in the following *time-unit*. If we change the value of a cell in the scope of an *unless* process, we need to be aware that the value will only be changed two *time-units* after.

## 5.5 Related work

The *Omax* model uses *FO*, but instead of using `ntcc`, it uses shared state concurrency (for synchronizing the improvisation phases) and message passing concurrency (for synchronizing OpenMusic and Max/Msp). Although this a remarkable model, we believe that `ntcc` can provide an easier way to synchronize processes and to reason about the correctness of the implementation because it is obviously easier to synchronize declaratively. Constraints provide a much more powerful way to express declaratively complex synchronizing patterns. Since the `ntcc` model has a logical counterpart [24], it is possible to prove properties of the model. For instance, the fact that it always (or never or sometimes) chooses the longest context, or that repetitions of some given subsequence are avoided.

*Probabilistic Ccofmi* [29] fixes the problems with synchronization and extends the notion of probabilistic choice in the improvisation phase, giving it a clear and concise semantic. However, it does not model how can probabilistic distributions may change from a *time-unit* to another based on user and computer interaction.

# 6 Implementation

A `ntcc` interpreter is a program that takes `ntcc` models and creates a program that interacts with an environment, simulating the behavior of the `ntcc` models. `Ntcc` interpreters (including our interpreter) are designed to simulate a finite `ntcc` model. It means that they only simulate a finite number of *time-units*.

During the last decade, three interpreters for `ntcc` have been developed. *Lman* [22] by Hurtado and Muñoz in 2003, *NtccSim* (http://avispa.puj.edu.co) by the Avispa research group in 2006, and *Rueda's sim* in 2006. They were intended to simulate `ntcc` models, but they were not made for real-time interaction. Recall from the introduction that soft real-time interaction means that the user does not experience noticeable delays in the interaction.

When designing a `ntcc` interpreter, we need a constraint solving library or programming language allowing us to check stability (i.e., know when a *time-unit* is over), check entailment (i.e., know if a constraint can be deduced from the *store*), post constraints, and synchronize the concurrent access to the *store*. These tasks must be performed efficiently to achieve a good performance.

The authors of the `ntcc` model for interactive scores proposed to use Gecode as a constraint solving library for future `ntcc` interpreters, and create an interface for Gecode to OpenMusic to specify multimedia interaction applications. Furthermore, they proposed to extend *Lman* to work under Mac OS X using Gecode.

One objective of this thesis is to develop a prototype for a `ntcc` interpreter real-time capable. We followed the advise from the authors of the interactive scores model and we tried out several alternatives to develop an interpreter using Gecode.

Our first attempt was using a thread to represent each `ntcc` process in the simulation. However, we found out that using threads adds an overhead in the performance of the interpreter because of the context-switch among threads, even when using lightweight (lw) threads. Then, we tried using event-driven programming. Performance was better compared with threaded implementations. However, each time a **when** process asks if a condition can be entailed, we need to check for stability, thus adding an unnecessary overhead. The reader may find more information about our previous attempts in Appendix 9.3 and performance results in chapter 7.

Our implementation, *Ntccrt*, is once again based on a simple but powerful concept. The **when** and processes are encoded as propagators in Gecode. That way Gecode manages all the concurrency required for the interpreter. Gecode calls the continuation of a process when a process condition is assigned to true.

On the other hand, **tell** processes are trivially codified to existing Gecode propagators and timed agents (i.e. *, !, **unless**, and **next**) are managed providing different process queues for each *time-unit* in the simulation.

Our interpreter works in two modes, the developing mode and the interaction mode. In the developing mode, the users may specify the `ntcc` system that they want to simulate in the interpreter. In the interaction mode, the users execute the models and interact with them.

This chapter is about the design and implementation of *Ntccrt*. We explain how to encode all the `ntcc` processes. We also explain the execution model of the interpreter. After that, we show how to run *Ccfomi* in the interpreter.

In addition, we describe how we made an interface to OpenMusic and how we can generate binary plugins for data-flow programming languages: Pure Data (Pd) [32] or Max/Msp [33] where MIDI, audio, or video inputs/outputs can interact with a *Ntccrt* binary. Finally, we give some conclusions, future work, and a short description of the other existing interpreters. A detailed description of *Ntccrt*, the generation of binary plugins, Pure Data, Max/Msp, and the previous *Ntccrt* prototypes can be found in a previous publication [48].

## 6.1 Design of *Ntccrt*

Our first version of *Ntccrt* allowed us to specify `ntcc` models in C++ and execute them as stand-alone programs. Current version offers the possibility to specify a `ntcc` model on either Lisp, Openmusic or C++. In addition, currently, it is possible to execute `ntcc` models as a stand-alone program or as an *external* object (i.e., a binary plugin) for Pd or Max.

### 6.1.1 Developing mode

In order to write a `ntcc` model in *Ntccrt*, the user may write it directly in C++, use a parser that takes Common Lisp macros as input or defining a graphical "patch" in OpenMusic. Using either of these representations, it is possible to generate a stand-alone program or an *external* object (fig 11).

0.6

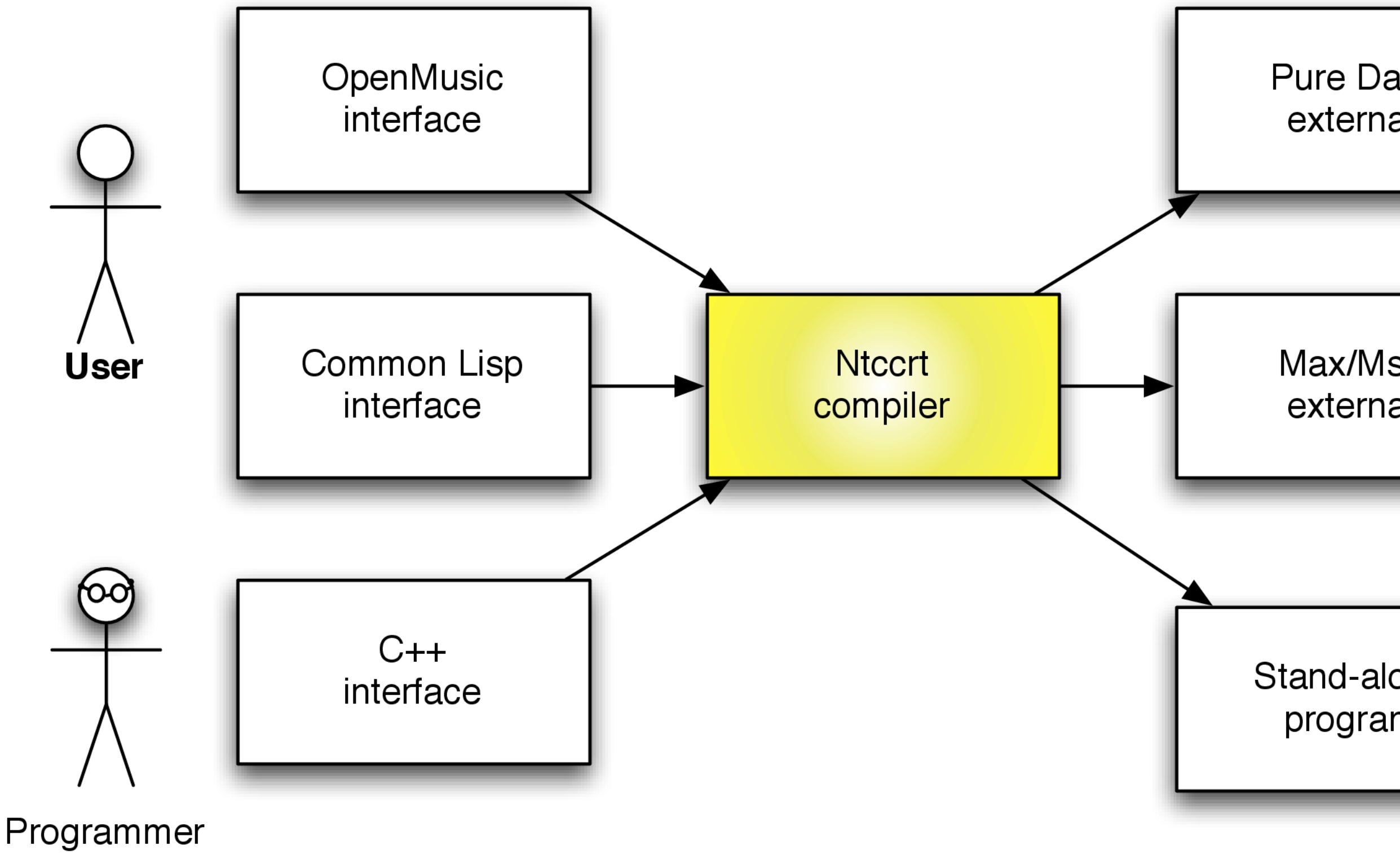

Figure 11: Ntccrt: Developing mode

To make an interface for OpenMusic, first, we developed a Lisp parser using Common Lisp macros to write an `ntcc` model in Lisp syntax and translate it to C++ code. Lisp macros extend Lisp syntax to give special meaning to characters reserved for users for this purpose. Executing those macros automatically compile a `ntcc` program.

After the success with Lisp macros, we created OpenMusic methods to represent `ntcc` processes. Openmusic methods are a graphical representation using the Common Lisp Object System (CLOS). Those graphical objects are placed on a graphical "patch". Executing the "patch" generates a *Ntccrt* C++ program.

### 6.1.2 Execution mode

To execute a *Ntccrt* program we can proceed in two different ways. We can create a stand-alone program that can interact with the Midishare library [12], or we can create an *external* object for either Pd or Max. An advantage of compiling a `ntcc` model as an *external* object lies in using control signals and the message passing API provided by Pd and Max to synchronize any graphical object with the *Ntccrt* external.

0.6

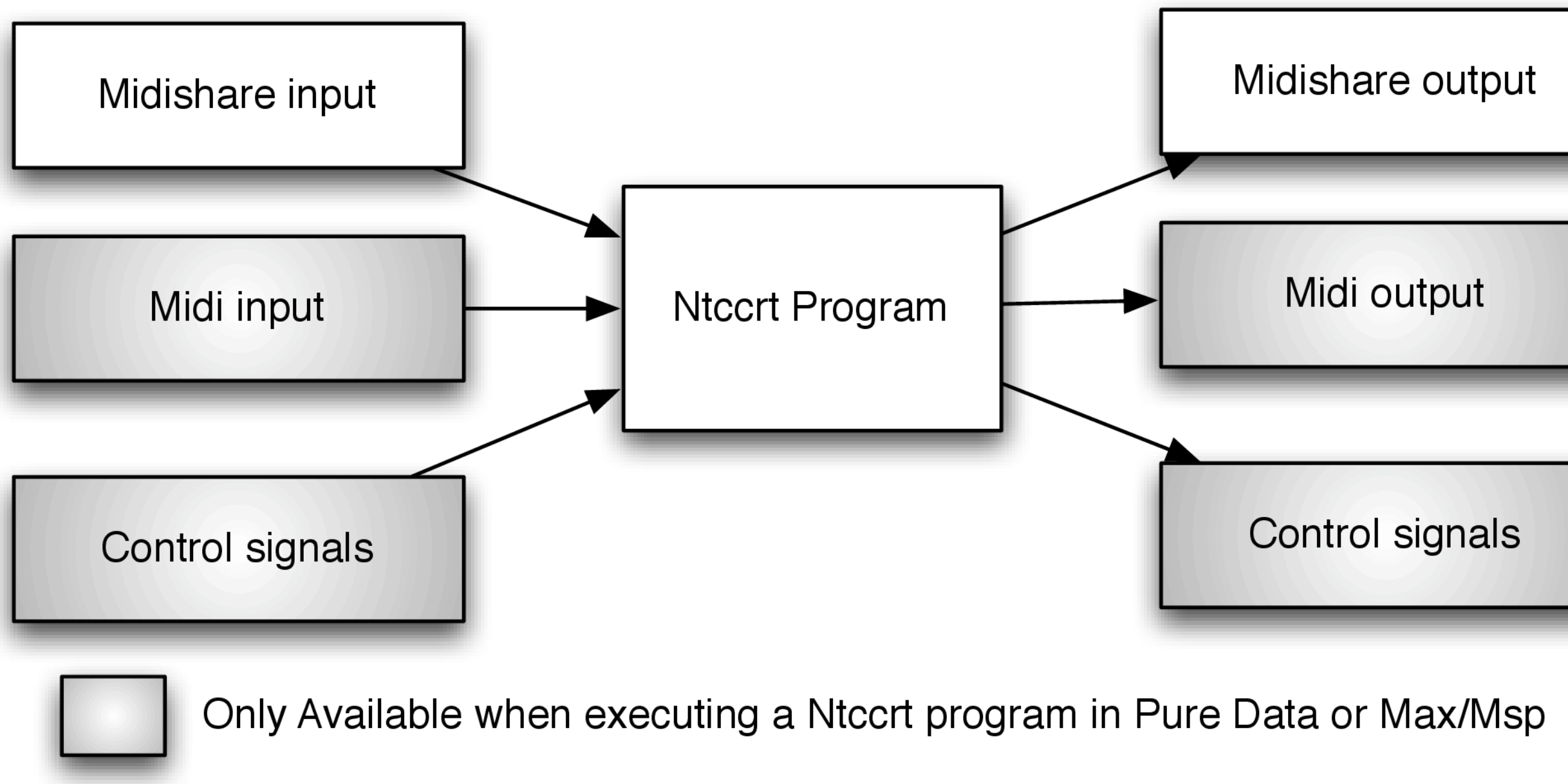

Figure 12: Ntccrt: Interaction mode

To handle MIDI streams (e.g., MIDI files, MIDI instruments, or MIDI streams from other programs) we use the predefined functions in Pd or Max to process MIDI. Then, we connect the output of those functions to the *Ntccrt* binary plugin. We also provide an interface for Midishare, useful when running stand-alone programs (fig. 12).

## 6.2 Implementation of *Ntccrt*

*Ntccrt* is the first `ntcc` interpreter written in C++ using Gecode. In this section, we focus on describing the data structures required to represent each `ntcc` agent. Then, we explain how the interpreter makes a simulation of a `ntcc` model. `Ntcc` agents are represented by classes. To avoid confusions, we write the agents with bold font (e.g., when C do P) and the classes with *italic font* (e.g., *When* class).

### 6.2.1 Data structures

To represent the constraint systems we need to provide new data types. Gecode variables work on a particular *store*. Therefore, we need an abstraction to represent `ntcc` variables present on multiple *stores* with the same variable object. Boolean variables are represented by the *BoolV* class, FD variables by the *IntV* class, FS variables by the *SetV* class, and infinite rational trees (with unary branching) by *SetVArray*, *BoolVArray*, and *IntVArray* classes.

After encoding the constraint systems, we defined a way to represent each process. All of them are classes inheriting from *AskBody*. *AskBody* is a class, defining an *Execute* method, which can be called by another object when it is nested on it.

To represent the **tell** agent, we defined a super class *Tell*. For this prototype, we provide three subclasses to represent these processes: **tell** ($a$=$b$), **tell** ($a$∈$B$), and **tell** (*a*>*b*). Other kind of **tell** agents can be easily defined by inheriting from the *Tell* superclass and declaring an *Execute* method.

For the **when** agent, we made a *When propagator* and a *When* class for calling the propagator. A process **when** *C* **do** *P* is represented by two propagators: (a reified propagator for the constraint *C*) and **if** *b* **then** *P* **else** *skip* (the *When* propagator). The *When propagator* checks the value of *b*. If the value of *b* is true, it calls the *Execute* method of *P*. Otherwise, it does not take any action. Figure 13 shows how to encode the process **when** *a*=*c* **do** *P* using the *When propagator*

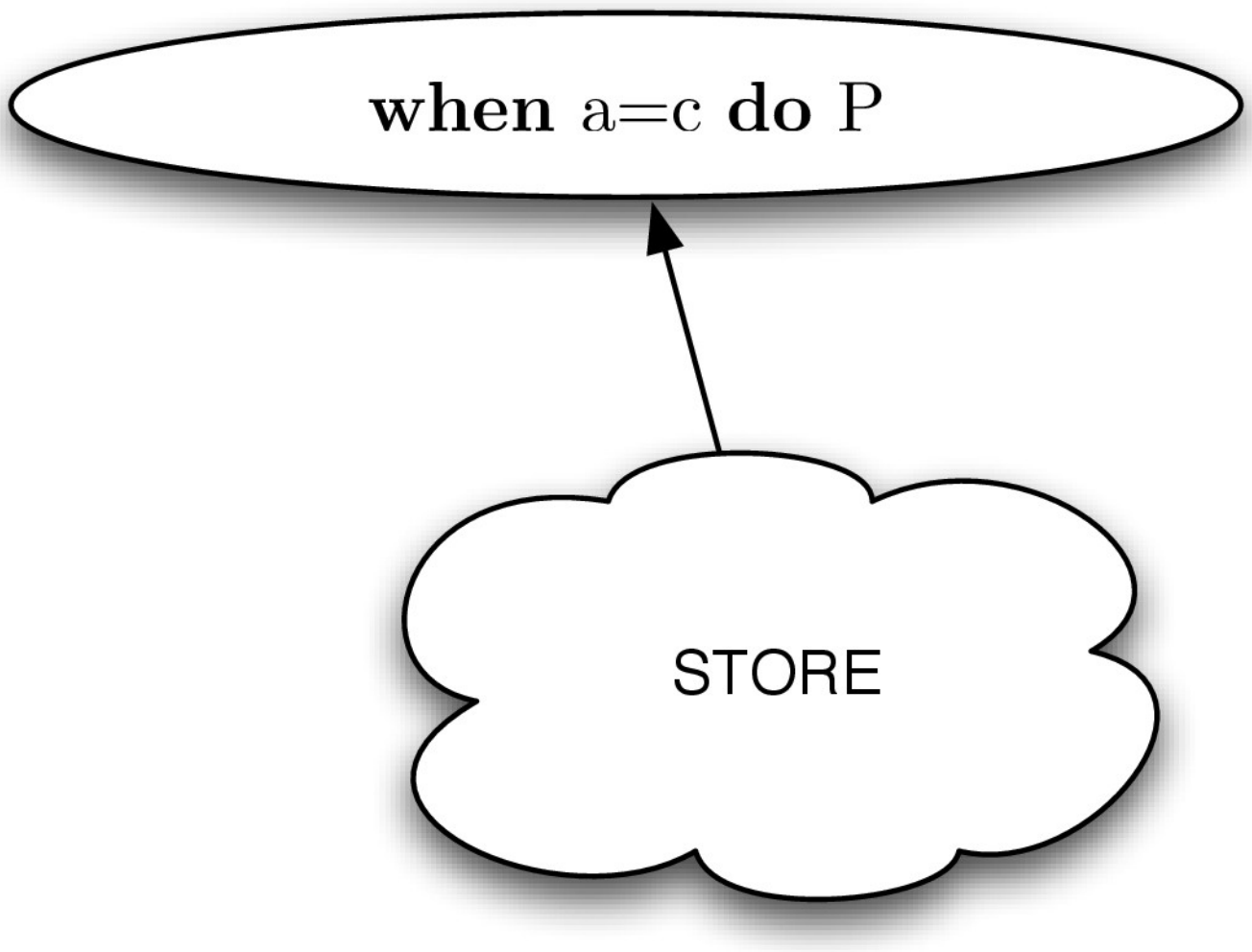

0.7 0.7

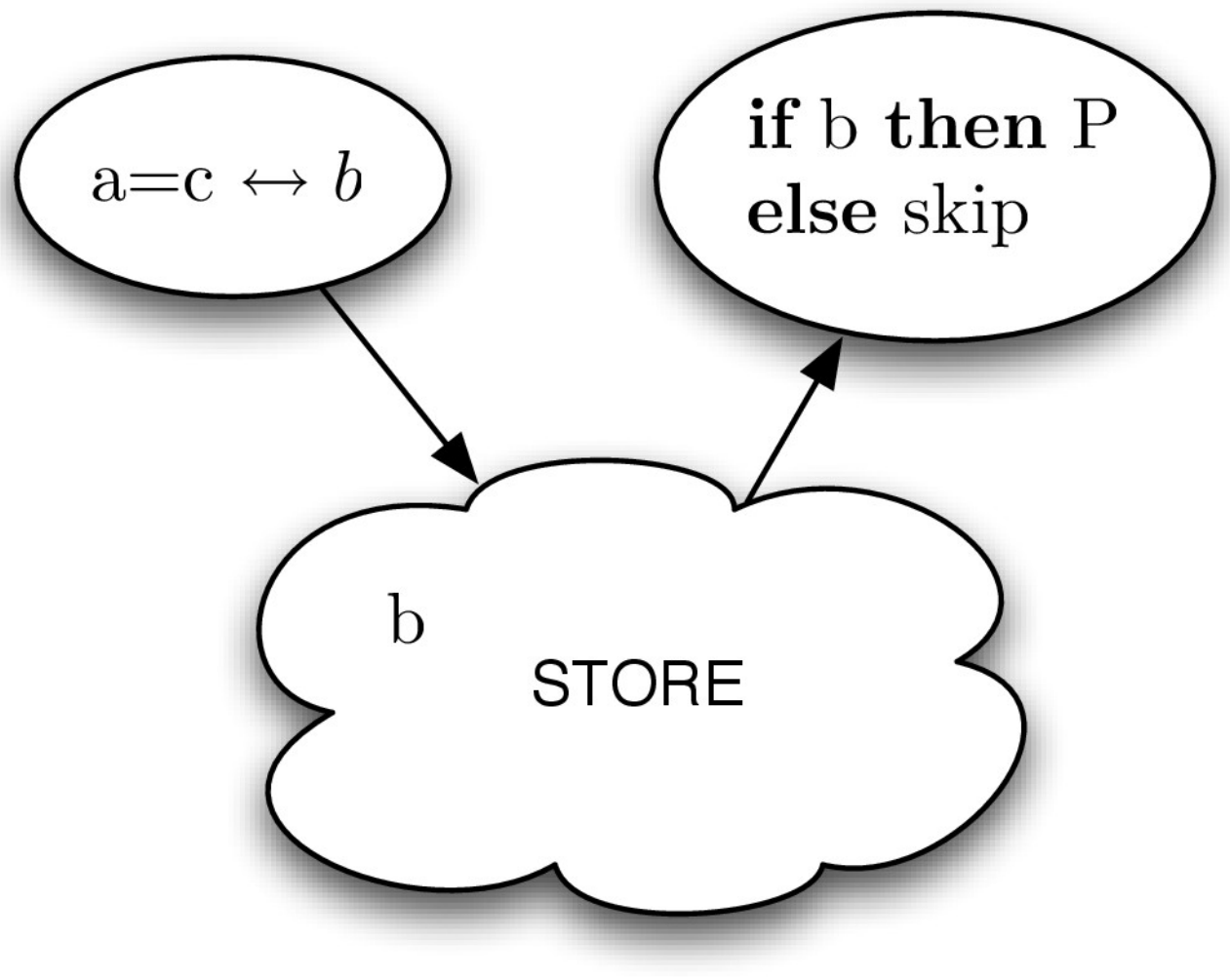

Figure 13: Example of the *When propagator*

To represent the agent (i.e. non-deterministic choice ) we made the *parallel conditional propagator*. This propagator receives a sequence of tuples , where is a *Gecode* boolean variable representing the condition of a *reified propagator* (e.g., ) and  (a pointer to an *AskBody* object) is the process to be executed when  is assigned to *true*.

The *When propagator* executes the process  associated to the first guard that is assigned to true. It means  such that . Then, its work is over. If all the variables are assigned to *false*, its work is over too.

The *When propagator* is based on the idea of the *Parallel conditional combinator* proposed by Schulte [46]. A curious reader might ask how we obtain a non-deterministic behavior. In order to make a non-deterministic choice, we pass the parameters to the propagator in a random order. That way, the propagator always chooses the first process which condition is true, but since the processes (and conditions) are given in a random order, it will simulate a non-deterministic choice. Figure 14 shows how to encode the process  **when**  **do tell** () using the *parallel conditional propagator*. This process is explained in Appendix 9.4.

0.7

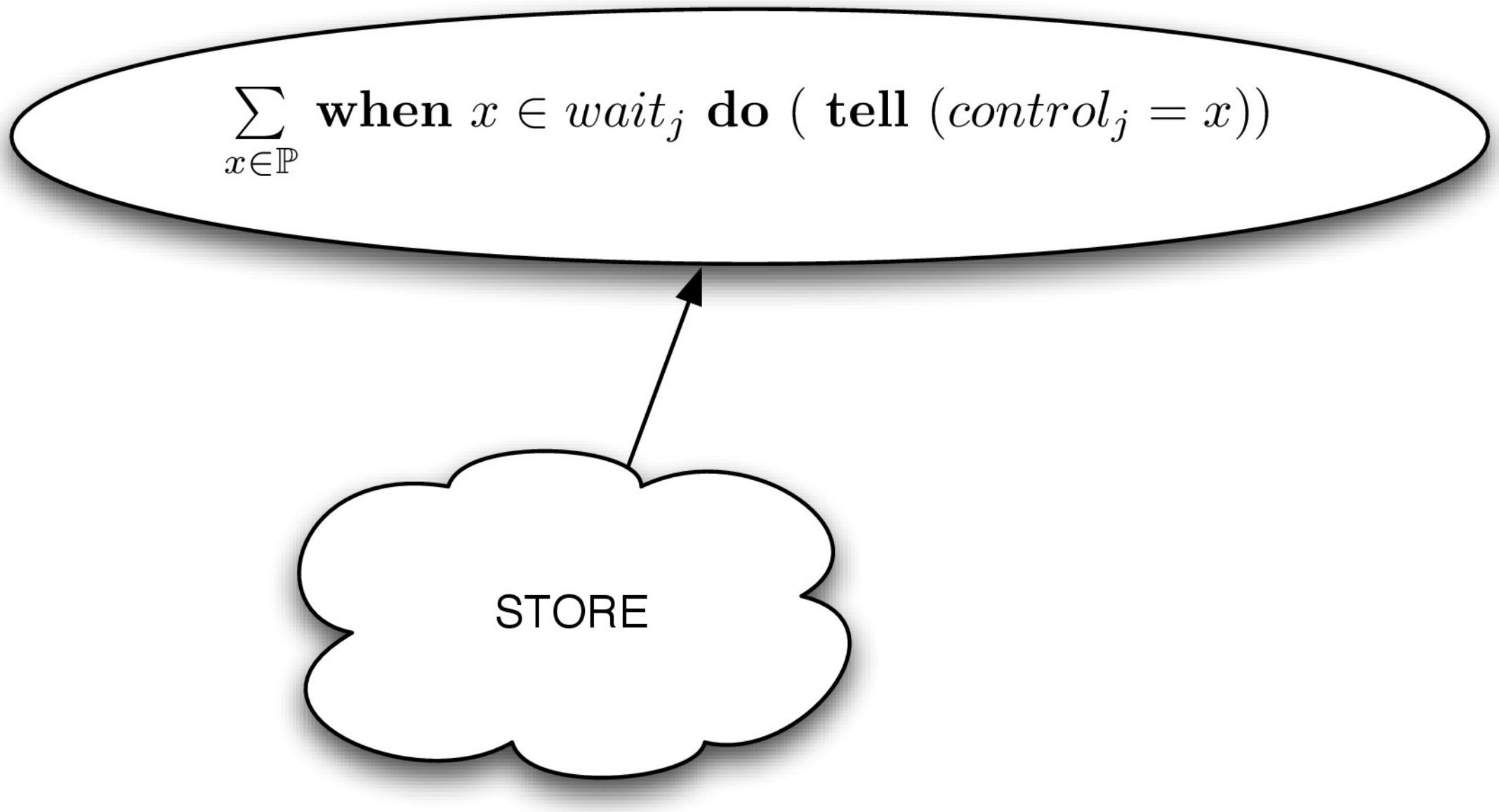

0.7

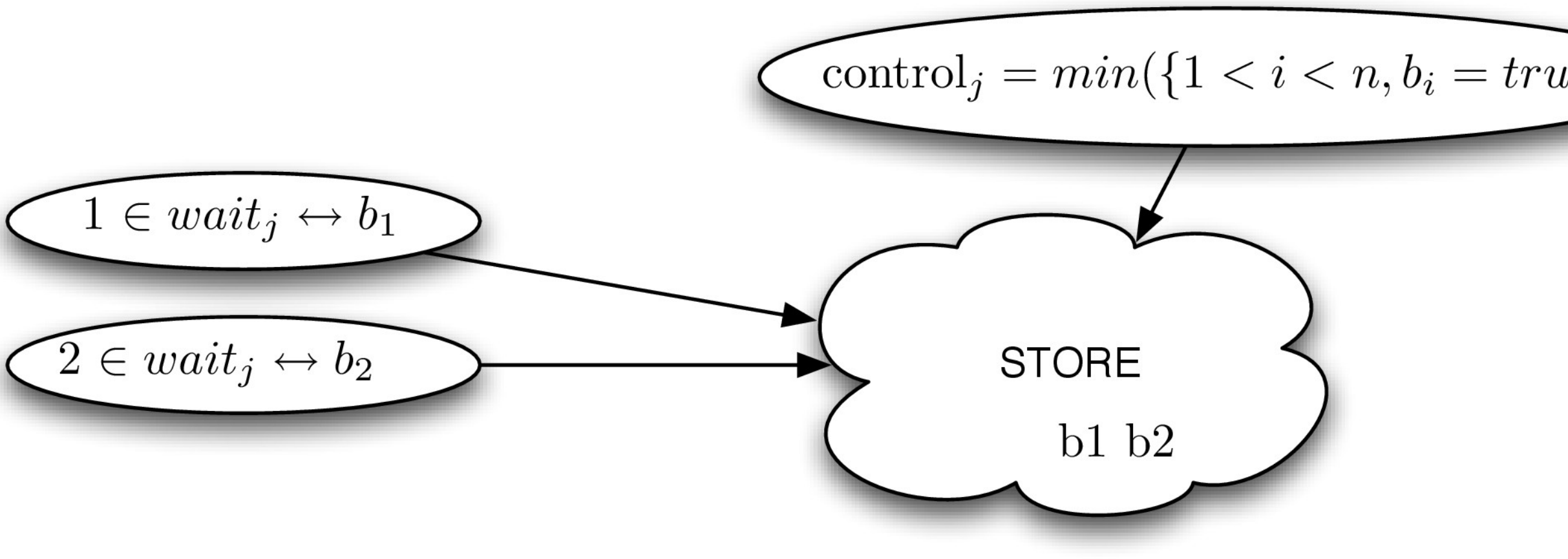

Figure 14: Example of the *Parallel conditional propagator*

Local variables are easily represented by an instruction allowing the user to create a new variable at the beginning of a procedure. Then, that new variable persists during the following *time-units* being simulated. This implementation of local variables is useful when there is a process !*P* and *P* contains local variables. The other variables are declared at the beginning of the simulation.

Timed agents are represented by the *TimedProcess* class. *TimedProcess* is an abstract class providing a pointer for the current *time-unit*, for a queue used for the **unless** processes, for a queue used for the persistent assignation processes, for a queue used for the other processes, and for the continuation process. Each subclass defines a different *Execute* method. For instance, the *Execute* method for the *Star* class randomly chooses the *time-unit* to place the continuation (an *AskBody* object) in its corresponding *process queue*.

The *Unless* class and the *Persistent assignation* class are different. The *Execute* method of the *Unless* objects and the the *Persistent assignation* objects are called after calculating a *fixpoint* common to all the processes in the *process queue*. Formally, a *propagator* can be seen as a function , receiving a *store* and returning a *store*. A *fixpoint* for a *propagator* is a *store* $x$ such that $F(x)=x$. When Gecode calculates a *store*, which is a *fixpoint* for all *propagators*, we said that the *store* is stable.

After calculating a *fixpoint*, if the condition for the *Unless* cannot be deduced from the stable *store*, its continuation is executed in the next *time-unit*. On the other hand, the *Persistent assignation* copies the domain *D* of the variable assigned, when the *store* is stable. Then, it assigns *D* to that variable in following *time-units* (creating a *tell* object for each following *time-unit*).

We also have a *Procedure* class used to model both, `ntcc` simple definitions (e.g., A **tell**(*a*=2)) and `ntcc` recursive definitions (e.g., B(*i*) *B*(*i*+1)), which are invocated using the *Call* class. For `ntcc` recursive definitions, we create local variables simulating call-by-value (as it is specified in the formalism). Recursion in `ntcc` is restricted. Parameters have to be variables in the *store* and we can only make a recursive call in a recursive procedure. However, *Ntccrt* does not check these conditions (they are left to the user) and implements general recursion.

### 6.2.2 Execution model

In order to execute a simulation, the users write a `ntcc` mdel in *Ntccrt*, compile it, and then they call the compiled program with the number of units to be simulated and the parameters (if any) of the main `ntcc` definition. For each *time-unit i*, the interpreter executes the following steps: First, it creates a new *store* and new variables in the *store*. Then, it processes the input (e.g., MIDI data coming from *PD* or *Max*). If it is simulating the first *time-unit*, it calls the main `ntcc` definition with the arguments given by the user.

After that, it moves the *unless* processes to the *unless queue*, moves the *persistent assignation* processes to the *persistent assignation queue*, and executes all the remaining processes in the *process queue*. Then, it calculates a *fixpoint*. Note how we only calculate one *fixpoint* each *time-unit*, opposed to the previous prototypes.

After calculating a *fixpoint*, it executes the *unless* processes in the *unless queue* and executes the *persistent assignations* in the *persistent assignation queue*. Then, it calls the *output processing* method (e.g., sending some variable values to the standard output or through a MIDI port). Finally, it deletes the current *store*. Figure 15 illustrates the execution model.

0.5

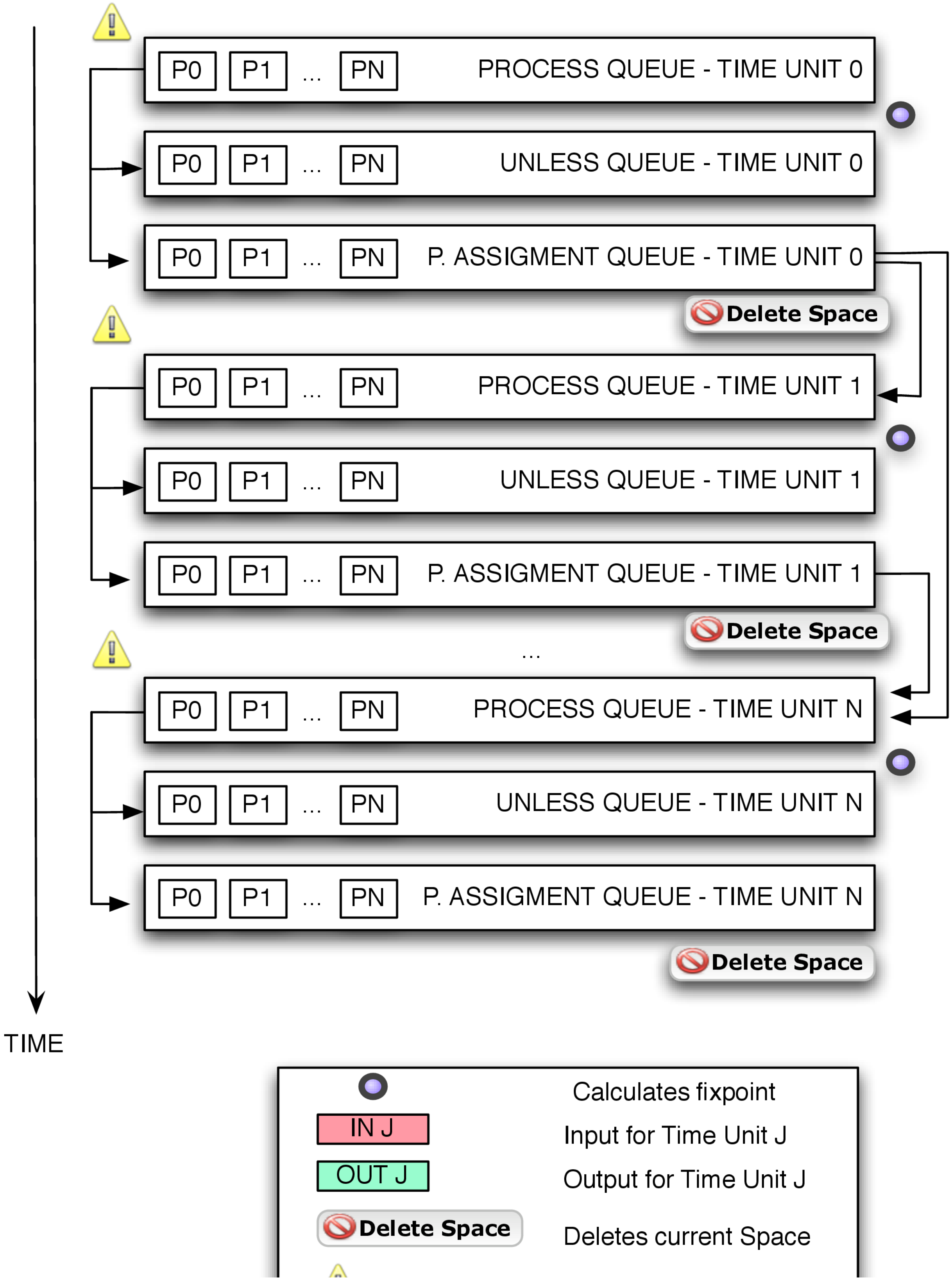

P0 P1 ... PN PROCESS QUEUE - TIME UNIT 0
P0 P1 ... PN UNLESS QUEUE - TIME UNIT 0
P0 P1 ... PN P. ASSIGMENT QUEUE - TIME UNIT 0
Delete Space
P0 P1 ... PN PROCESS QUEUE - TIME UNIT 1
P0 P1 ... PN UNLESS QUEUE - TIME UNIT 1
P0 P1 ... PN P. ASSIGMENT QUEUE - TIME UNIT 1
Delete Space
...
P0 P1 ... PN PROCESS QUEUE - TIME UNIT N
P0 P1 ... PN UNLESS QUEUE - TIME UNIT N
P0 P1 ... PN P. ASSIGMENT QUEUE - TIME UNIT N
Delete Space
TIME
Calculates fixpoint
IN J
Input for Time Unit J
OUT J
Output for Time Unit J
Delete Space
Deletes current Space

Figure 15: Execution model of the `ntcc` interpreter

## 6.3 Implementation of *Ccfomi*

Rueda et al ran *Ccfomi* on their interpreter. They wrote Lisp macros to extend Lisp syntax for the definition of `ntcc` processes. We provide a similar interface to write `ntcc` processes in Lisp. Furthermore, we can write *Ccfomi* definitions in *Ntccrt* in an intuitive way using OpenMusic. For instance, the  process (presented in chapter 2), in charge of the synchronization between the  and the  processes, is represented with a few boxes: one for **parallel** processes, one for the ≤ condition, one for the = condition, and one for **when** and **unless** processes (fig. 16)

0.4

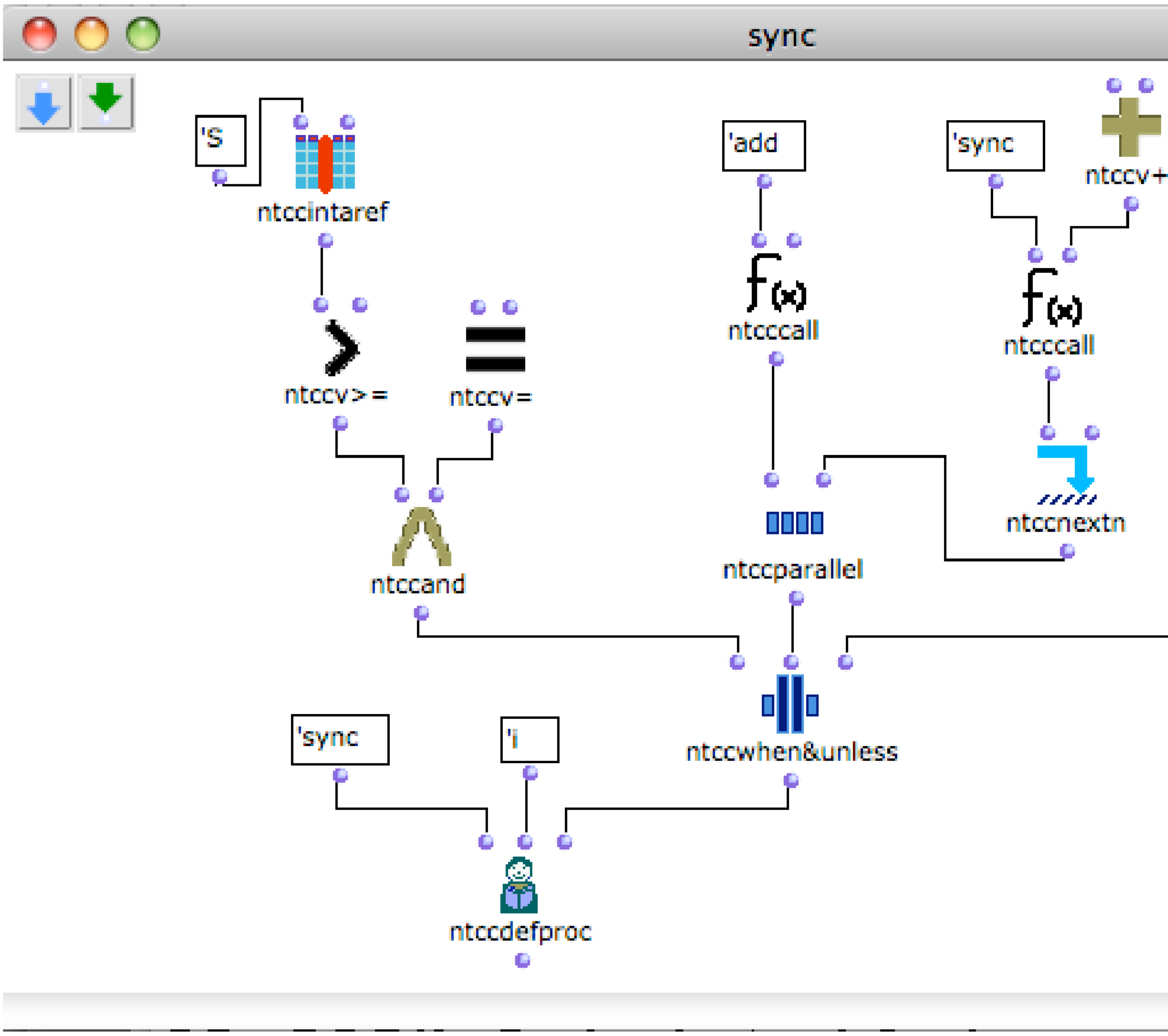

Figure 16: Writing the  process in OpenMusic

We successfully ran *Ccfomi* as an stand-alone program using Midishare. We present the results of our tests with the stand-alone program in Chapter Error: Reference source not found. We also ran it as a Pd plugin generated by *Ntccrt*. The plugin is connected to the midi-input, midi-output, and a clock (used for changing from a *time-unit* to the other). For simplicity, we generate a clock pulse for each note played by the user (fig. 17). In the same way, we could connect a *Metronome* object. *Metronome* is an object that creates a clock pulse with a fixed interval of time.

0.5

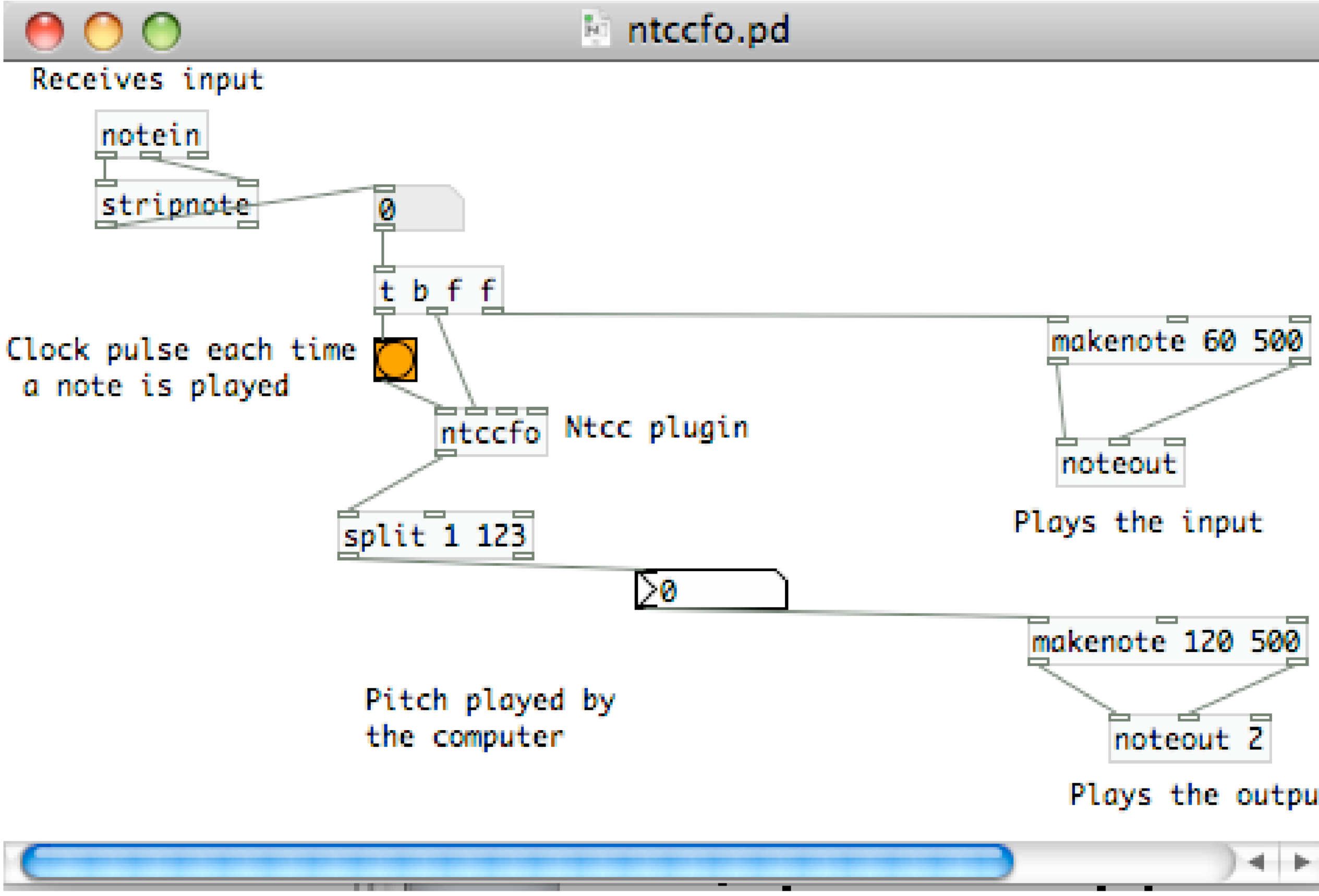

Figure 17: Running *Ccfomi* in Pure Data (Pd)

## 6.4 Summary

Further than just developing an interpreter, we developed an interface for OpenMusic to write `ntcc` models for *Ntccrt*. Although the OpenMusic interface generates code for *Ntccrt*, it is not able to embed Lisp code in the interpreter. In addition, the current version of the interpreter does not support probabilistic choice nor cells, required to run our model. This is acceptable because our objective was just to develop a `ntcc` interpreter prototype. For that reason, we still do not support `pntcc` nor cells (which are not basic operators on `ntcc`). In the following, we will describe the possibilities and limitations of the interpreter and possible solutions for future work.

Additionally, since we encoded the **When** processes as a *Gecode* propagators, we are able to use search in `ntcc` models without using the  agent. This is not possible when encoding the **when** processes as lightweight threads or OS threads because threads cannot be managed inside Gecode search engines. Models using non-deterministic choices are incompatible with the recomputation used in the search engines.

*Ntccrt* cannot simulate processes leading the *Store* to false. For instance,

**when** *false* **do next tell** (*fail*=*true*)
|**tell** (*a*=2)|**tell** (*a*=3)
Since the **when** agent is represented as a propagator, once the propagation achieves a fail state no more propagators will be called in that *time-unit*, causing inconsistencies in the rest of the simulation. Fortunately, processes reasoning about a false *Store* can be rewritten in a different way, avoiding this kind of situations. For instance, the process above can be rewritten as:

**when** *state*=*false* **do next tell** (*fail*=*true*)
|**tell** (*a*=2)|**tell** (*a*=3)|**tell** (*state*=*false*)
Although in many applications we do not want to continue after the *store* fails in a *time-unit* because a failed *store* is like an exception in a programming language (e.g., division by zero).

In addition, *Ntccrt* restricts the domains for the different constraint systems. The domain for FD variables is  and each set or tree in the FS and the rational trees variables cannot have more than  elements[6]. This limitation is due to Gecode, which uses the C++ integer data type for representing its variables.

Another problem arises when we want to call Lisp functions in the interpreter. This will be usefull to make computer music programs (written in Lisp) to interact with *Ntccrt*. Currently, we are only using Lisp to generate C++ code. However, it is not possible to embed Lisp code in the interpreter (e.g., calling a Lisp function as the continuation of a **when** process). To fix that inconvenient, we propose using *Gelisp* for writing a new interpreter, taking advantage of the call-back functions provided by the Foreign Function Interface (FFI) to call Lisp functions from C++. That way a propagator will be able to call a Lisp function. Although, this could have a negative impact on performance and in the correctness of the system (e.g., when the Lisp function does not end).

The implementation of cells is still experimental and it is not yet usable. The idea for a real-time capable implementation of cells is extending the implementation of persistent assignation. Cells, in the same way than persistent assignation, require to pass the domain of a variable from the current *time-unit* to a future *time-unit*. However, persistent assignation usually involves simple equality relations. On the other hand, the cells assignation may involve any mathematical function $g(x)$ (e.g ).

Probabilistic choice is not yet possible neither. For achieving it, we propose extending the idea used for non-deterministic choice agent . To model , it was enough by determining the first condition than can be deduced and then activate the process associated to it. For probabilistic choice, we need to check the conditions after calculating a *fixpoint*, because we need to know all the conditions that can be entailed before calculating the probabilistic distribution. When multiple probabilistic choice $\bigoplus$ operators are nested, we need to calculate a *fixpoint* for each nested level.

By implementing cells and probabilistic choice it would be easy to implement the model proposed for this work. Valencia proposed in [61] to develop model checking tools for `ntcc`. In the future, we propose using model checking tools for verifying properties of complex systems, such as ours.

6 It is not  because one bit is used for the sign.

In addition, Pérez and Rueda proposed in [29] exploring the automatic generation of models for probabilistic model checker such as *Prism* [18]. The reader should be aware that *Prism* has been used successfully to check properties of real-time systems. We believe that this approach can be used to verify properties in our system.

Finally, we found out that the *time-units* in *Ntccrt* do not represent uniform *time-units*, because in the stand-alone simulation they have different durations. This is a problem when synchronizing a `ntcc` program with other programs. To fix it, we made the duration of each *time-unit* take a fixed time. We did that easily by using the clock provided by Pd or Max and providing a clock input in *Ntccrt* plugins. That way we only start simulating a new *time-unit* once we receive a clock pulse.

On the other hand, fixing the duration of a *time-unit* has two problems. First, if the fixed time is less than the time required to compute all the processes in a *time-unit*, this makes the simulation incoherent. Second, it makes the simulation last longer because the fixed time has to be an upper limit for the *time-unit* duration.

## 6.5 Related work

*Lman* was developed as a framework to program RCX Lego Robots. It is composed of three parts: an abstract machine, a compiler and a visual language. We borrowed from this interpreter the idea of having several queues for storing `ntcc`'s processes, instead of using threads. Regrettably, since *Lman* only supports finite domain constraints.

*NtccSim* was used to simulate biological models [16]. It was developed in Mozart-Oz [37]. It is able to work with finite domains (FD) and a constraint system to reason about real numbers. We conjecture (it has not been proved) that using Mozart-Oz for writing a `ntcc` interpreter it is not as efficient as using *Gecode*, based on the results obtained in the benchmarks of Gecode, where Gecode performs better than Mozart-Oz in constraint solving.

*Rueda's sim* was developed as a framework to simulate multimedia semantic interaction applications. This interpreter was the first one representing rational trees, finite domain , and finite domain sets constraint systems. One drawback of this interpreter is the use of *Screamer* [47] to represent the constraint systems. *Screamer* is a framework for constraint logic programming written in Common Lisp. Unfortunately, *Screamer* is not designed for high performance. This makes the execution of the `ntcc` models in *Rueda's sim* not suitable for real-time interaction.

# 7 Tests and Results

Since the creation of *Lman*, performance and correctness have been the main issues to evaluate a `ntcc` interpreter. *Lman* was a great success in the history of `ntcc` interpreters because by using *Lman* it was possible to program Lego Robots, and formally predict the behavior of the robots. A few years later, *Rueda's sim* was capable to model multimedia interaction systems.

Although, it is beyond the scope of this research to evaluate whether those interpreters are faster than *Ntccrt* or whether they are able to interact in real-time with a human player, we conjecture that they are not appropriate for real-time interaction for simulating hundreds of *time-units* in complex models such as *Ccfomi*, based on the results presented by their authors.

In this chapter we want evaluate the performance of our `ntcc` interpreter prototypes and also to evaluate the behavior of *Ntccrt*. In order to achieve these goals, we performed different tests to *Ntccrt* and to our previous implementations of `ntcc`.

First, we tried to develop a generic implementation of lightweight threads that could be used in Lispworks. The purpose was to use threads to manage concurrency in `ntcc` interpreters. We compared Lisp processes (medium-weight threads), our implementation of threads based on continuations, and our implementation of threads based on event-driven programming.

We found out that continuations are not efficient in Lispworks. We also found out that the event-driven implementation of threads is faster than using Lisp processes or continuations. However, it is very difficult to express instructions such as go-to jumps, exceptions and local variable definition on event-driven programming.

Then, we tried using both Lisp processes and the event-driven threads to implement `ntcc` interpreters (explained in Appendix 9.3). We found out that context-switch of threads and the fact that it checks for stability constantly adds an overhead in the performance on the `ntcc` interpreter. For those reasons, we discarded using threads for the `ntcc` interpreter. We also found out that encoding `ntcc` processes as Gecode propagators outperforms the threaded implementations of the interpreter.

Each test presented in this chapter was taken with a sample of 100 essays. Time was measured using the *time* command provided by Mac OS X and the *time* macro provided by Common Lisp. All tests were performed under Mac OS X 10.5.2 using an Imac Intel Core 2 duo 2.8 Ghz and Lispworks Professional 5.02.

In the graph bars, we present the average of those samples. The vertical axe is measure in seconds in all graphs. We do not present standard deviation nor other statistical information because the differences of performances between one implementation and another were considerable high to reason about the performance of the implementations. Sometimes, we do not present all the bars in a graph because they do not fit the scale of the graph.

## 7.1 Testing *Ntccrt* performance

In order to test *Ntccrt* performance, we made two tests. First, we compared a `ntcc` specification to find paths in a graph with other three implementations. Second, we tested *Ccfomi* using *Ntccrt*. Recall from the beginning of this chapter that each test was taken with a sample of 100 essays. Time was measured using the *time* command provided by Mac OS X and the *time* macro provided by Common Lisp. All tests were performed under Mac OS X 10.5.2 using an Imac Intel Core 2 duo 2.8 Ghz and Lispworks Professional 5.02.

## 7.2 Test: Comparing implementations of `ntcc` interpreters

We compared the execution times of simulating the specification presented to find paths in graph concurrently (explained in detail in Appendix 9.4) running on the event-driven Lisp interpreter and *Ntccrt*. We also compared them with a concurrent constraint implementation on Mozart/OZ and a recursive implementation in Lisp (fig. 18).

0.8

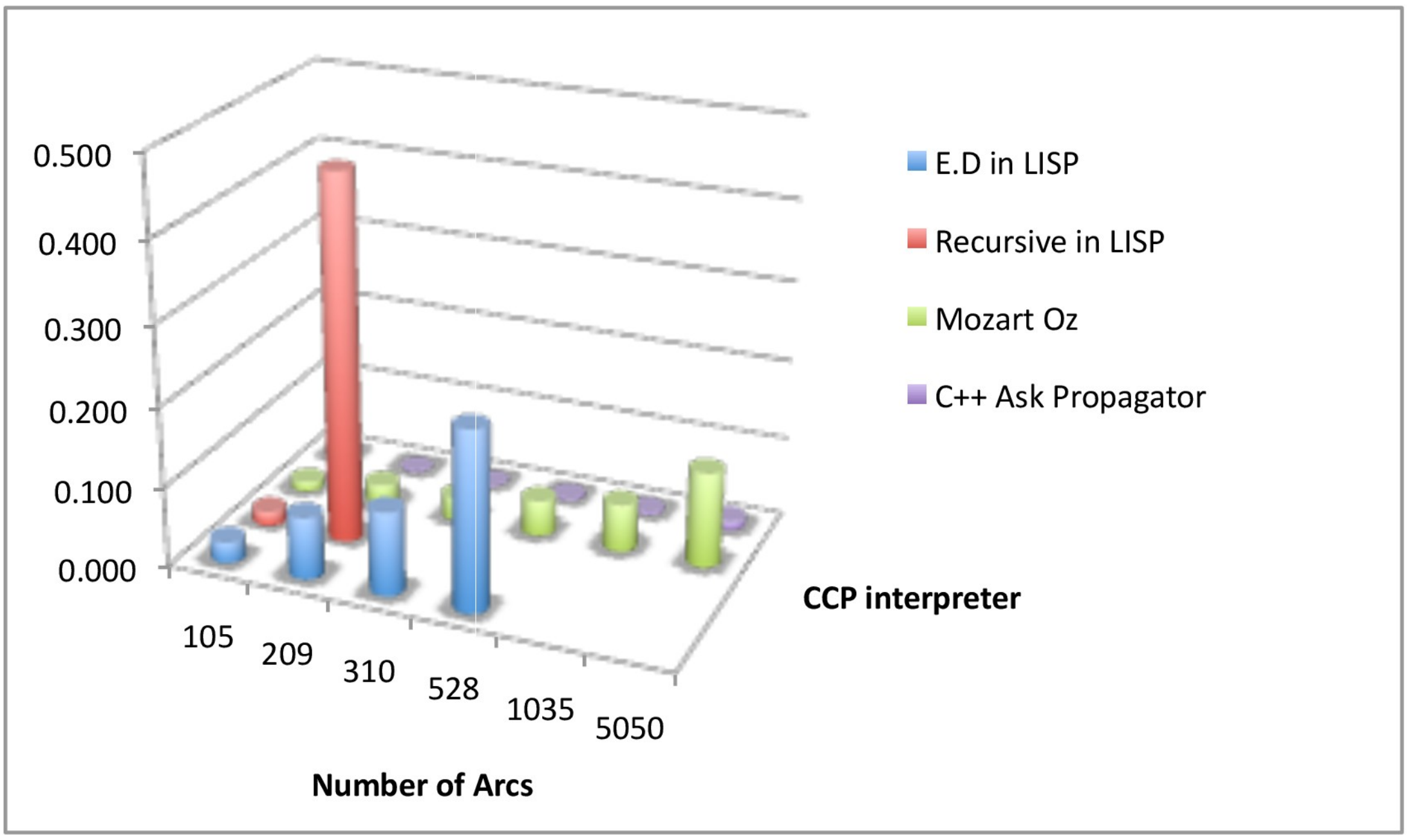

Figure 18: Comparing implementation to find paths in a graph

## 7.3 Test: Executing *Ccfomi*

*Ccfomi* is able to receive up to one note each *time-unit*. A reasonable measure of performance is the average duration of a `ntcc` *time-unit* during the simulation. We ran *Ccfomi* in *Ntccrt* with a player interpretating at most the first 300 notes of J.S Bach's two-part Invention No. 5, as studied in [9]. The player chooses (non-deterministically) to play a note or postpone the decision for the next *time-unit*. It took an average of 20 milliseconds per *time-unit*, scheduling around 880 processes per time-unit, and simulating 300 *time-units*. We simulated these experiment 100 times. Detailed results can be found at Appendix 9.6. We do not present musical results, since it is out of the scope of this work to conclude whether *Ccfomi* produces or not an improvisation "appealing to the ear". We are only interested on performance tests.

Pachet argues in [28] that an improvisation system able to learn and produce sequences in less than 30ms is appropriate for real-time interaction. Since *Ccfomi* has a response time of 20ms in average for a 300 *time-units* simulation, we conclude that it is capable of real-time interaction according to Pachet's research.

## 7.4 Summary

*Ntccrt*, our `ntcc` interpreter based on encoding `ntcc` processes as Gecode propagators outperforms our threaded and our event-driven implementations of `ntcc`.

Since we are learning and producing sequences with a response time lower than 30 milliseconds then, according to the authors of the *Continuator*, we have a system fast enough to interact with a musician.

## 7.5 Related work

*Lman*'s developers ran a specification to play a MIDI pitch with a fixed duration each *time-unit* [21]. The tests were made using a Pentium III 930 MHz, 256 MB Ram, Linux Debian Woody (3.0), and the RCX 2.0 Lego robot with running BrickOS 2.6.1. They made a simulation with 100 *time-units*.

This simple process takes an average of 281.25 ms to run each *time-unit* using *Lman*, unfortunately it is not suitable for real-time interaction in music, even if we would run it on modern computers.

On the other hand, Rueda's interpreter ran *Ccfomi* on a 1.67 GHz Apple PowerBook G4 using Digitool's MCL version of Common Lisp, taking an average of 25 milliseconds per *time-unit*, scheduling around 20 concurrent processes. They also made a simulation with 100 *time-units*.

Unfortunately, Rueda's implementation uses some MCL's functions (not defined in the Common Lisp standard) and we were not able to run his interpreter in Mac OS X Intel to compare it with *Ntccrt*. On the other hand, *Lman* is designed for Linux and it is no longer maintained for current versions of Linux and Tcl/tk.

# 8 Conclusions

In this chapter, we present a summary of the thesis, some concluding remarks, and we propose some future work thoughts.

## 8.1 Summary

- We explained how we can model music improvisation using probabilities, extending the notion of non-deterministic choice described in Ccfomi. Although this idea is very simple, the probabilities are computed in constant time and space when the FO is built. We managed to preserve the linear complexity in time and space of the FO on-line construction algorithm.
- Calculating the probability of being on a certain scale makes the model more appropriate for certain music genres, but it requires to calculate multiple constants, which vary according to the genre of tonal music where the user is improvising. For that reason, it is discarded.
- We explained how we can change the intensity of the notes generated in the improvisation. This kind of variation in the intensity is something new for machine improvisation systems as far as we know. We believe that this kind of approach, where simple variations can be made while preserving the style learned from the user and being compatible with real-time implementations, should be a topic of investigation in future work.
- We used cells to represent the variables changing from a *time-unit* to another. Using cells we modeled a probabilistic distribution that changes according to the user and computer interaction. As far as we know, this is the first `pntcc` model where probabilistic distributions change between *time-units*. Unfortunately, current version of *Ntccrt* does not support cells nor probabilistic choice.
- We ran *Ccfomi* in *Ntccrt* taking an average of 20 milliseconds per time-unit (see Chapter 7). Since we are learning and producing sequences with a response time less than 30 milliseconds then, according to the authors of the Continuator, we have a system fast enough to interact with a musician.

- Although Gecode was designed for solving combinatory problems using constraints, we found out that using Gecode for Ntccrt give us outstanding results for writing a `ntcc` interpreter.
- Unfortunately, the interpreter is not able to execute processes leading the *store* to false. However, processes reasoning about a false *store* can be rewritten in a different way, avoiding this kind of situations.

# 8.2 Concluding remarks

We show how we can make a probabilistic extension of *Ccfomi* using the Factor Oracle. This extension has three main features. First, it preserves the linear time and space complexity of the on-line Factor Oracle algorithm. The Factor Oracle was chosen as the data structure to capture the user style in *Ccfomi* because of its linear complexity. Our extension would not be worth if we had changed the complexity fo the Factor Oracle on-line construction algorithm in order to add probabilistic information to the model, making it incompatible with real-time.

Second, we are using `pntcc` (a probabilistic extension of `ntcc`) for our model. The advantage of `pntcc` is that we do not need to to model all the processes in a new calculus to extend *Ccfomi*, instead we use `pntcc` where we have all the agents defined in `ntcc` (except the * agent, which is not used in this work) and a new agent for probabilistic choice. Adding probabilistic choice to *Ccofmi*, we avoid loops without control during the improvisation that may happen without control in *Ccfomi* due to its non-deterministic nature. In addition, changing the probability distribution, we could favor repetitions in the improvisation, if desired.

Third, the variation in the intensity during the improvisation. This is, as far as we know, the first model considering this kind of variation. Generating variations in the intensity during improvisation, we avoid sharp differences between the user and computer intensity, making the improvisation appealing to the ear (according the musicians we interviewed). Variations in the musical attributes are well-known for decades in Computer Assisted Composition, but in interactive systems (such as machine improvisation) variations are still an open subject, in part, due to the real-time requirements of the interactive systems.

If the reader does not consider relevant using process calculi (such as `pntcc`) to model, verify and execute a real-time music improvisation system, we pose the reader the following questions. Has the reader developed a real-time improvisation system on a programming language mixing non-deterministic and probabilistic choices? Try verifying the system formally! Is it an easy task? Would the reader be able to write such system in 50 lines of code? Using `pntcc`, we did it.

If we can model such systems using `pntcc` and process calculi have been well-known in theory of concurrency for the past two decades, why they have not been used in real-life applications? Garavel argues that models based on process calculi are not widespread because there are many calculi and many variants for each calculus, being difficult to choose the most appropriate. In addition, it is difficult to express an explicit notion of time and real-time requirements in process calculi. Finally, he argues that existing tools for process calculi are not user-friendly.

We want to make process calculi widespread for real-life applications. We left the task of representing real-time in process calculi and choosing the appropriate variant of each calculus for each application to senior researchers. This work focuses on developing a real-life application with `pntcc` and showing that our interpreter *Ntccrt* is a user-friendly tool, providing a graphical interface to describe `ntcc` processes easily and compile models such as *Ccfomi* to efficient C++ programs capable of real-time user interaction. We also showed that our approach to design *Ntccrt* offers better performance than using threads or event-driven programming to represent the processes.

The reader may argue that although we can synchronize *Ntccrt* with an external clock provided by Max or Pd, this does not solve the problem of simulating models when the clock step is smaller than the time necessary to compute a *time-unit*. In addition, the reader may argue that we encourage formal verification of `ntcc` and `pntcc` models, but there is not an existing tool to verify these models automatically, not even semi-automatically.

The reader is right! For that reason, currently the Avispa research group (sponsored by Pontificia Universidad Javeriana de Cali) is developing an interpreter for an extension of `ntcc` capable of modeling *time-units* with fixed duration. In addition, Avispa is proposing to Colciencias a project called Robust theories for Emerging Applications in Concurrency Theory: Processes and Logic Used in Emergent Systems (REACT-PLUS). REACT-PLUS will focus on developing verification tools for `ntcc`, `pntcc` and other process calculi. In addition, the project will continue developing faster and easier to use interpreters for them.

We invite the reader to read the following section to know about the future work thoughts that we propose. In addition, the reader may find more information about the REACT-PLUS proposal at http://www.lix.polytechnique.fr/comete/pp.html.

## 8.3 Future work

In the future, we propose extending our research in the following directions.

## 8.4 Extending our model

We propose capturing new elements in the music sequences. For instance, considering the music timbre, music pitch/amplitude variation (e.g., vibrato, bending and acciaccatura), and resonance effects (e.g., delay, reverb and chorus).

## 8.5 Improvisation set-ups

Several concurrent improvisation situation set-ups have been proposed [6], [10], but none of them have been implemented for real-time music improvisation. Rueda et al. in [40] propose the following set-ups: *n* performers and *n* oracles learning and performing; one performer, one oracle learning, and several improvisation processes running concurrently in the same oracle; one performer and several oracles learning from different viewpoints of the same performance.

## 8.6 Using Gelisp for *Ntccrt*

Currently, we are only using Lisp to generate C++ code. However, it is not possible to embed Lisp code in the interpreter. To work around that, we propose using *Gelisp* for writing a new interpreter, taking advantage of the call-back functions provided by the Foreign Function Interface (FFI) to call Lisp functions from C++. That way a process can trigger the execution of a Lisp function.

## 8.7 Adding support for cells for *Ntccrt*

The idea for a real-time capable implementation of cells is to extend the implementation of persistent assignation. Cells, in the same way than persistent assignation, require to pass the domain of a variable from the current *time-unit* to a future *time-unit*.

## 8.8 Developing an interpreter for `pntcc`

Pérez and Rueda already propose an interpreter for pntcc. To achieve probabilistic choice in Ntccrt, we propose extending the idea used for non-deterministic choice agent . To model , it was enough by determining the first condition that can be deduced and then activate the process associated to it. For probabilistic choice, we need to check the conditions after calculating a *fixpoint*, because we need to know all the conditions that can be entailed before calculating the probabilistic distribution. When multiple probabilistic choice $\bigoplus$ operators are nested, we need to calculate a *fixpoint* for each nested level.

## 8.9 Developing an interpreter for `rtcc`

There is not a way to describe the behavior of a ntcc *time-unit* if the fixed time is less than the time required to execute all the processes scheduled. For that reason, we propose developing an interpreter for the *Real Time Concurrent Constraint* (rtcc) [45] calculus.

Rtcc is an extension of ntcc capable of dealing with strong time-outs. Strong time-outs allow the execution of a process to be interrupted in the exact instant in which internal transitions cause a constraint to be inferred from the *store*. Rtcc is also capable of delays inside a single time unit. Delays inside a single time unit allows to express things like "this process must start 3 seconds after another starts". Sarria proposed in [45] developing an interpreter for rtcc. We believe that we can extend *Ntccrt* to simulate rtcc models.

## 8.10 Adding other graphical interfaces for *Ntccrt*

For this work, we conducted all the tests under Mac OS X using Pd and stand-alone programs. Since we are using Gecode and Flext to generate the externals, they could be easily compiled to other platforms and for Max. We used Openmusic to define an iconic representation of ntcc models. In the future, we also propose exploring a way of making graphical specifications for ntcc similar to the graphical representation of data structures in Pd.

## 8.11 Developing model checking tools for *Ntccrt*

Valencia proposed using model checking tools for verifying properties in complex ntcc models. In addition, Pérez and Rueda proposed developing model checking tools for pntcc. For instance, they propose exploring the automatic generation of models for *Prism* based on a pntcc model. We propose generating models to existing model checkers automatically to prove properties of the systems before simulating them on *Ntccrt*.

## Acknowledgements

Removed for double-blind reviewing.

# 9 Appendix

## 9.1 Algorithms

Following, we present four algorithms. The Factor Oracle (*FO*) on-line algorithm, the *FO* algorithm that calculates the *context*, our first approach to extend the *FO* algorithm with probabilistic choice, and an example of the *current dynamics* algorithm.

### 9.1.1 Factor Oracle on-line algorithm

This is the on-line algorithm to build a *FO* presented in [1].

**ADD-LETTER**(,σ)
Create a new state *m*+1
Create a new transition from *m* to *m*+1 labeled by σ
k
**while** *k*>−1 and there is no transition from *k* by σ **do**
Create new transition from *k* to *m*+1 by σ
k
**if** *K*==−1 **then**
*s* 0
**else** *s* where leads the transition from k by σ
(*m*+1) *s*
**return** Oracle() *s*

**ORACLE-ON-LINE**()
Create Oracle(ε) with one single state
−1
**for** *i* 0 **to** *m* **do**
Oracle() ADD-LETTER(Oracle(),)

**Theorem 1** *The complexity of Oracle-On-line is O(m) in time and space [1].*

### 9.1.2 Factor Oracle on-line algorithm that calculates the context

Following the present the FO algorithm that calculates the context, preserving linear time and space complexity. It was taken from [19].
The algorithm to add a new symbol to the *FO*

```
NewAddLetter(Oracle(p[1..i],σ)
Create a new state i+1
δ(i,σ)  i+1
j
  i
while j>−1 and δ(j,σ) is undefined do
δj,σ  i+1
  j
j
if j=−1 then
s  0
else s  δ(j,σ)
  s
lrs[i+1]  LengthReppeatedSuffix(
return Oracle(p[1..i],σ)
```

Finding the length of the repeated suffix of *p*[*i*..*i*+1]

```
LengthReppeatedSuffix()
if s=0 then
return 0
return LengthCommonSuffix()+1
```

Finding the common suffix of *p*[1..*i*] and *p*[1..*S*[*i*+1]−1]

```
LengthCommonSuffix()
if  then
return
else while  do

return
```

### 9.1.3 Example of the current dynamics algorithm

This is an example of executing the *current dynamics* algorithm (fig. 19) presented in chapter 4 for the sequence *D*=[28,28,38,25,40,30].

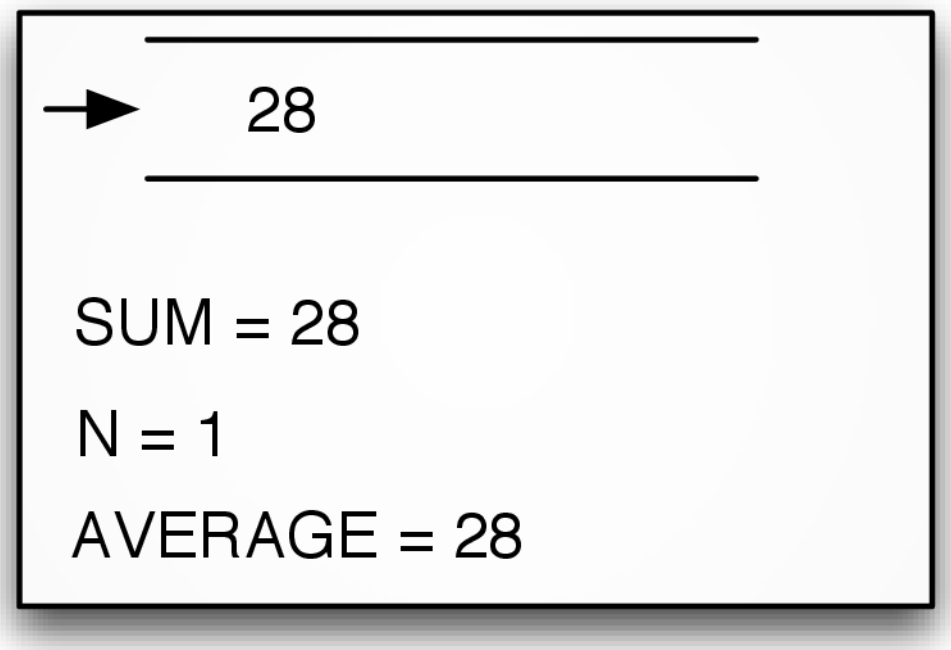

0.9

(a)

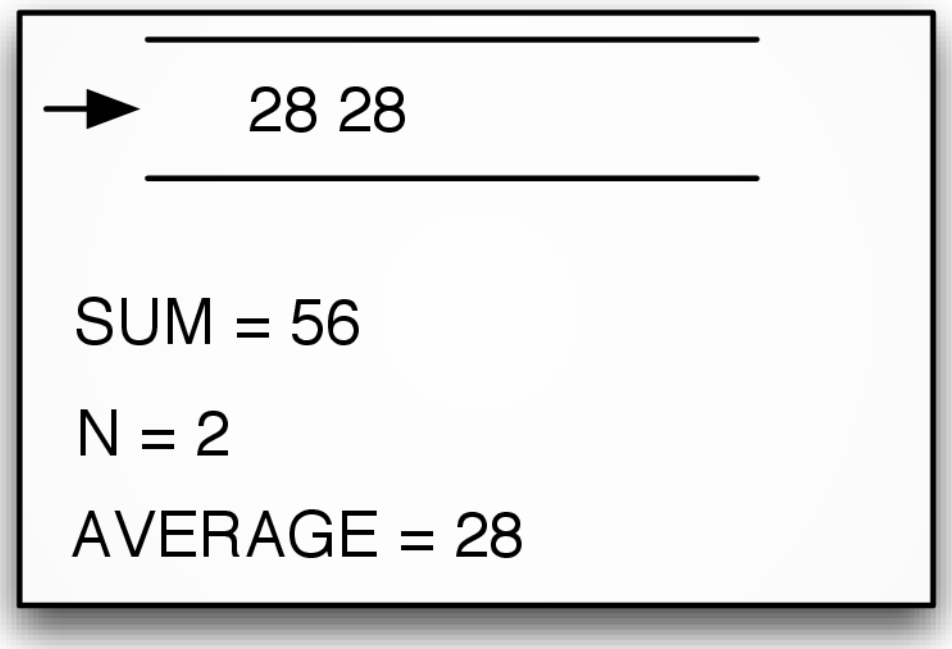
28 28
SUM = 56
N = 2
AVERAGE = 28

0.9

(b)

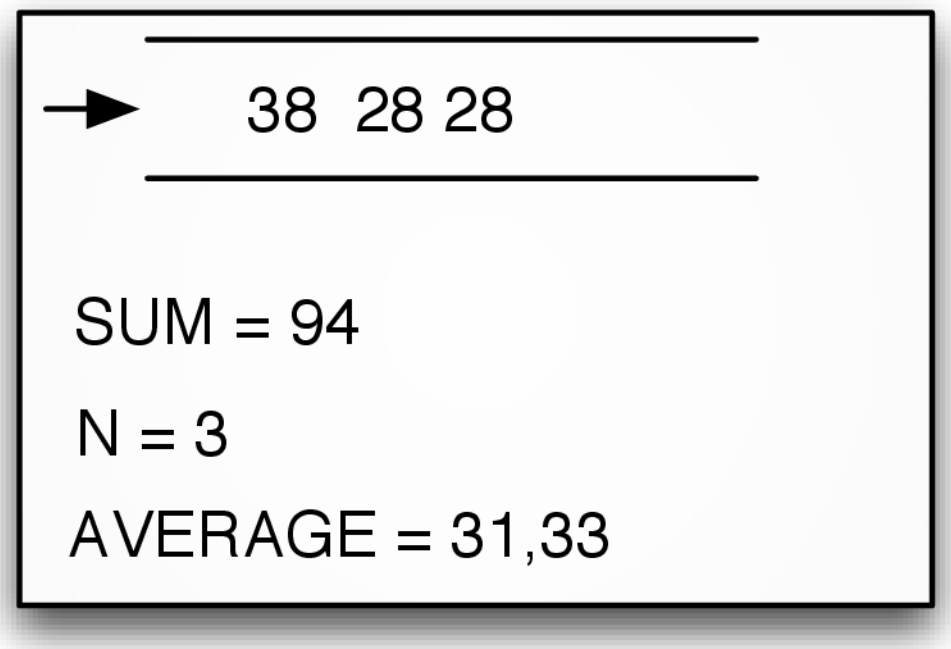
38 28 28
SUM = 94
N = 3
AVERAGE = 31,33

0.9

(c)

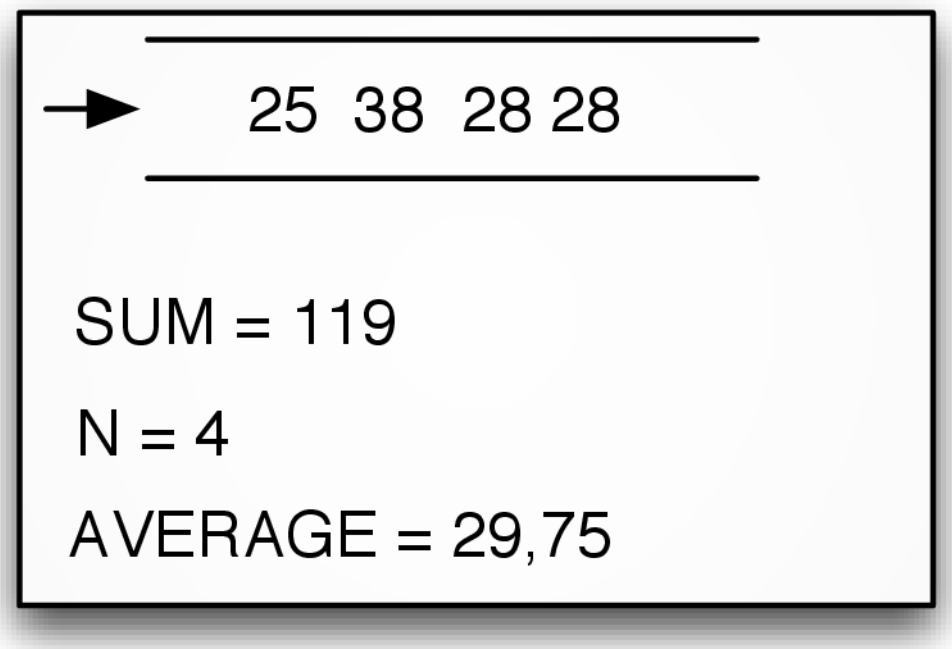
25 38 28 28
SUM = 119
N = 4
AVERAGE = 29,75

0.9

(d)

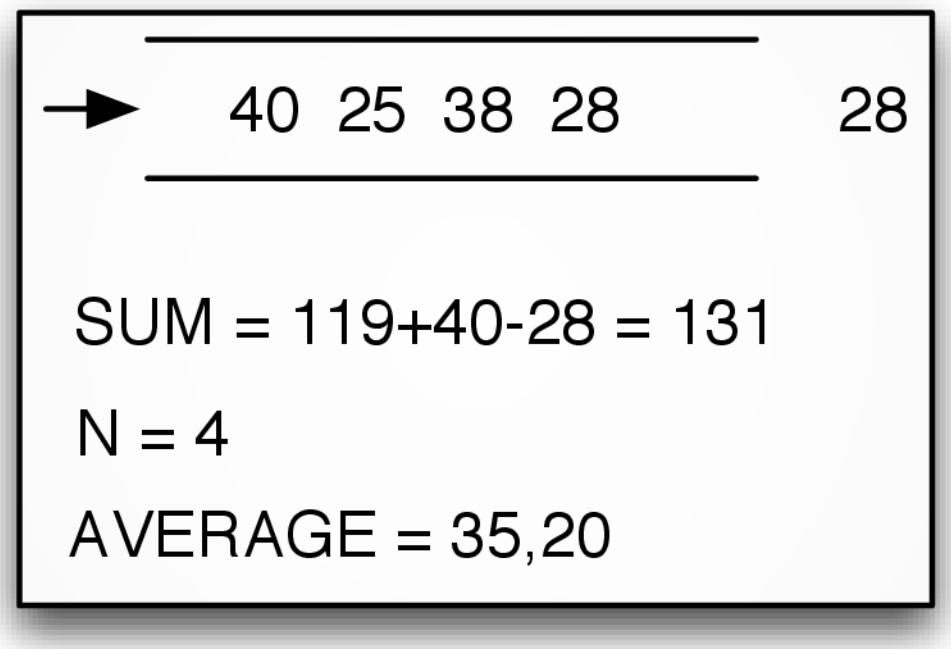

0.9

(e)

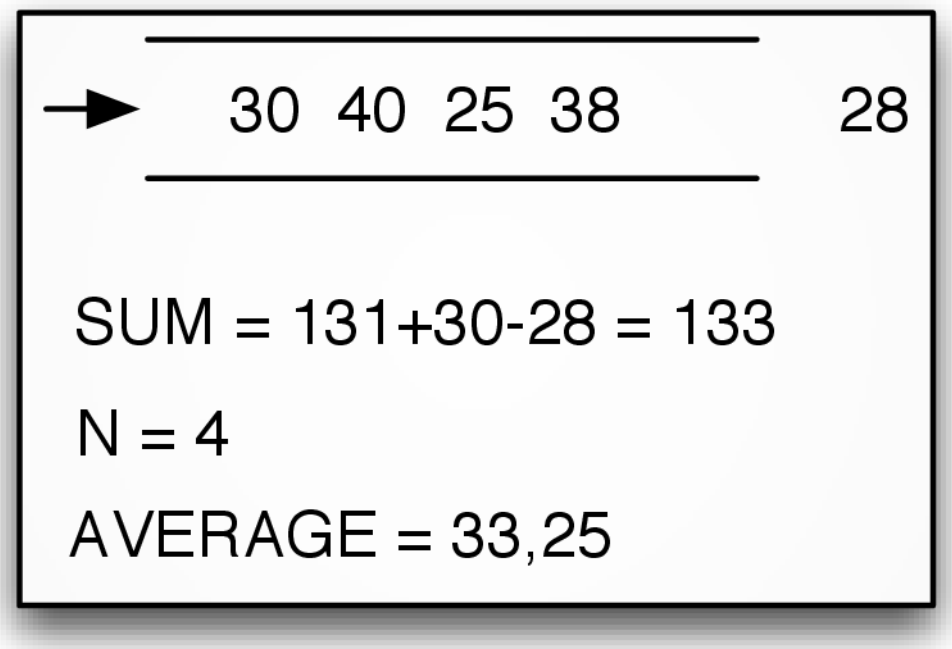

0.9

(f)

Figure 19: *Current dynamics* for *D*=[28,28,38,25,40,30] and τ=4.

## 9.2 Our previous approaches for the model

Following, we present our previous approaches to model probabilistic choice and changing the attributes of the notes during the improvisation. Probabilistic choice was discarded because it is not compatible with real-time. On the other hand, changing the pitch and the duration during the improvisation was discarded because it is not suitable for all music genres (e.g., music genres that are not based on music scales) and it requires elaborate training.

### 9.2.1 Extension for probabilistic choice

The idea behind this extension is to change all the values for ϱ leaving from state *i* when adding a new transition leaving from state *i*. In addition, when choosing a transition during the improvisation phase, it is necessary to change the value ϱ for all the transitions leaving from state *i*. This posses a big problem, to change the value of ϱ for all the transitions leaving from state *i*, changes the complexity of the *FO* on-line algorithm from linear to quadratic in time. For that reason, this extension was discarded.

### 9.2.2 Adding a new transition to the *FO*

Let γ take values in the range [0..1]. γ is a constant that regulates the priority for the new transitions added to the *FO*. Figure 20 represents the process of adding a new transition to the *FO*.

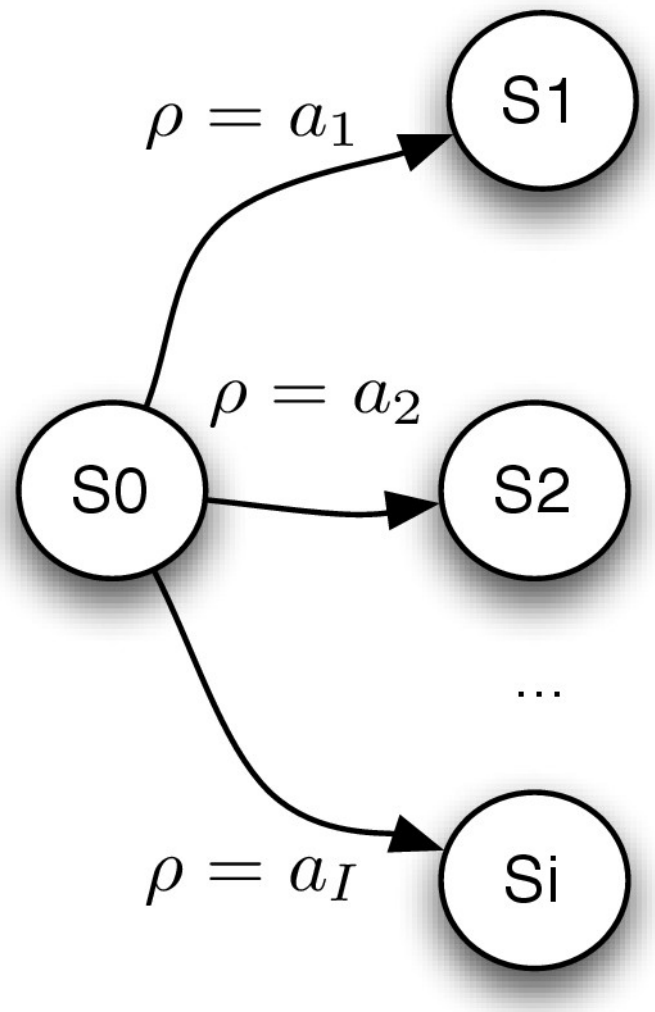

0.6

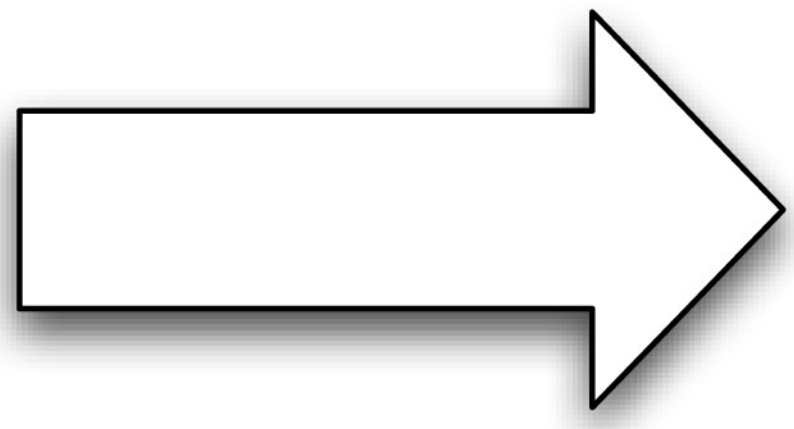

0.2

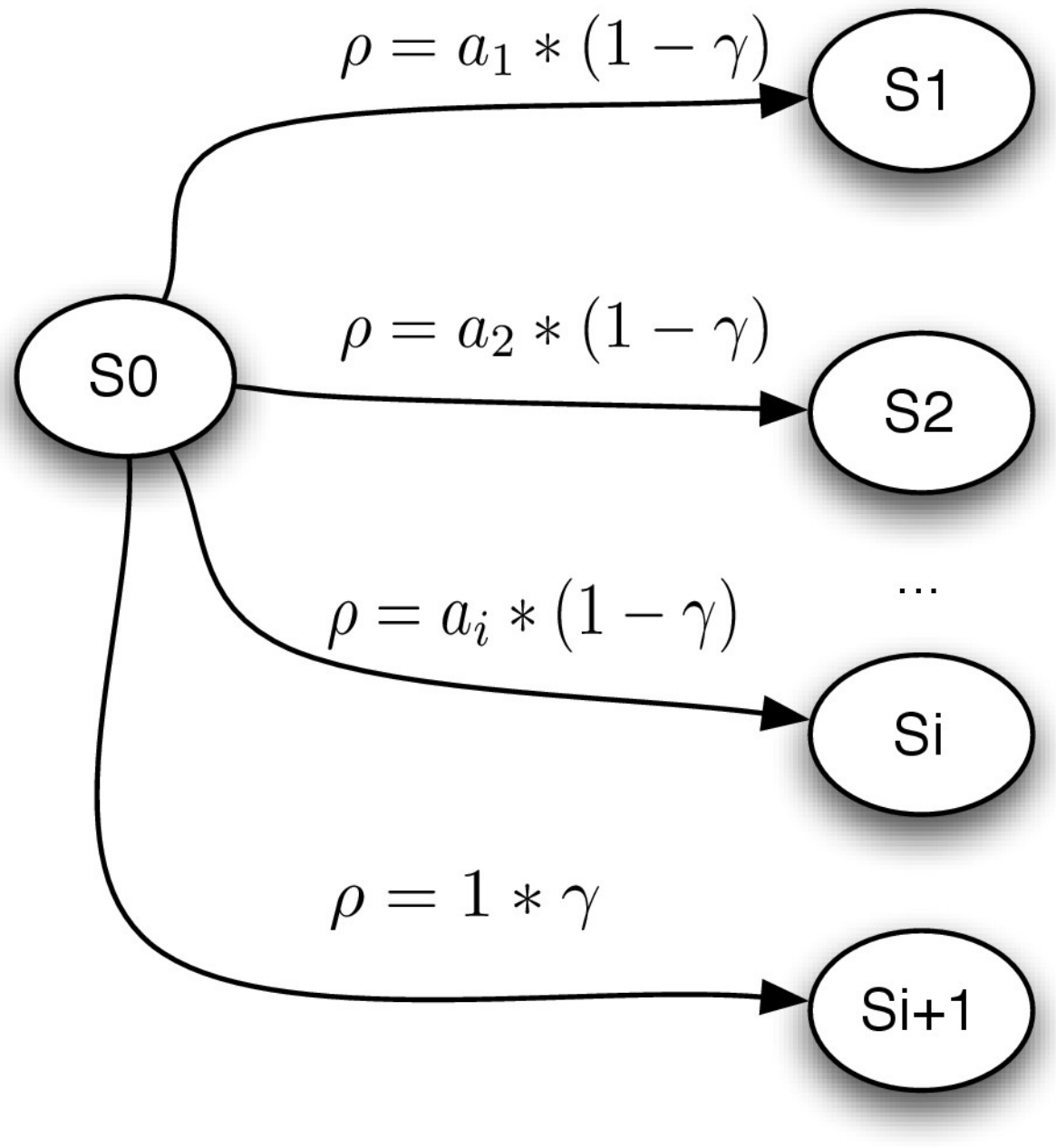

0.6

Figure 20: Adding a new transition to the FO

### 9.2.3 Choosing a transition during improvisation

Let β take values in the range [0..1], and be a constant regulating the change of probabilities when choosing a transition. The process of changing the probabilities when choosing a transition is represented in figure 21.

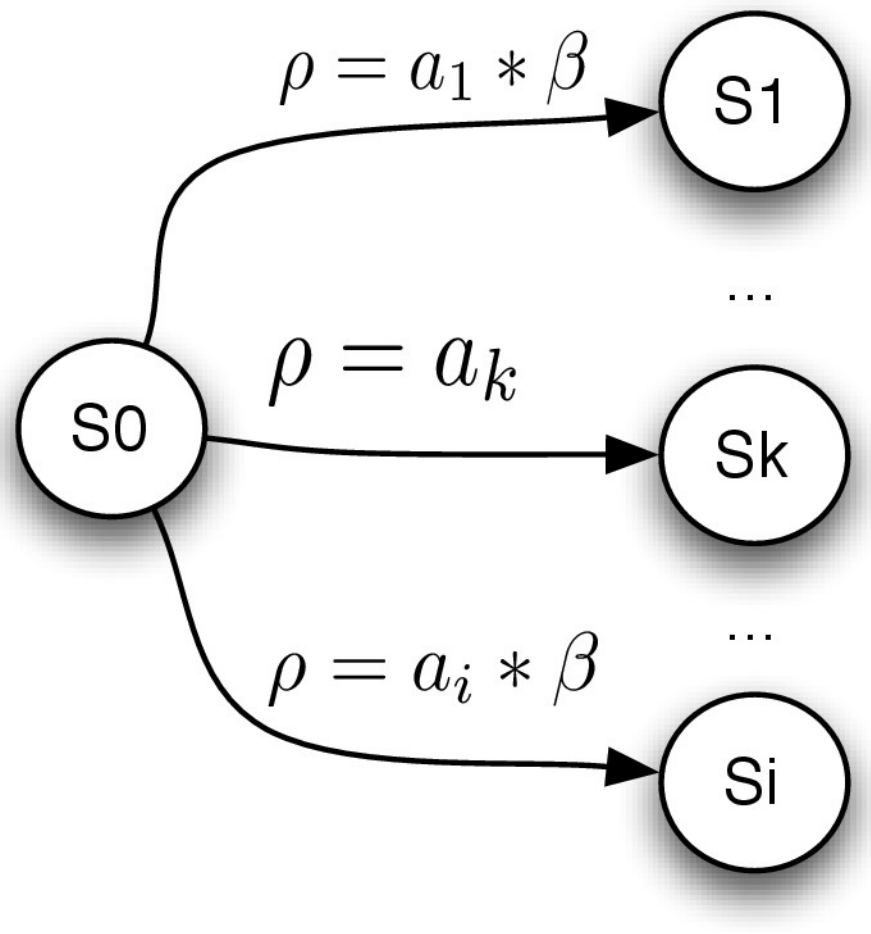

0.6

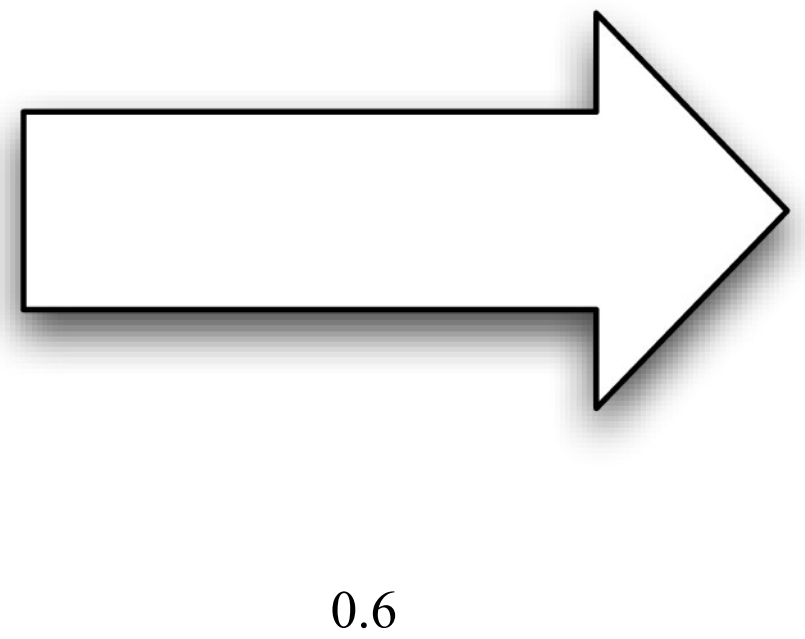

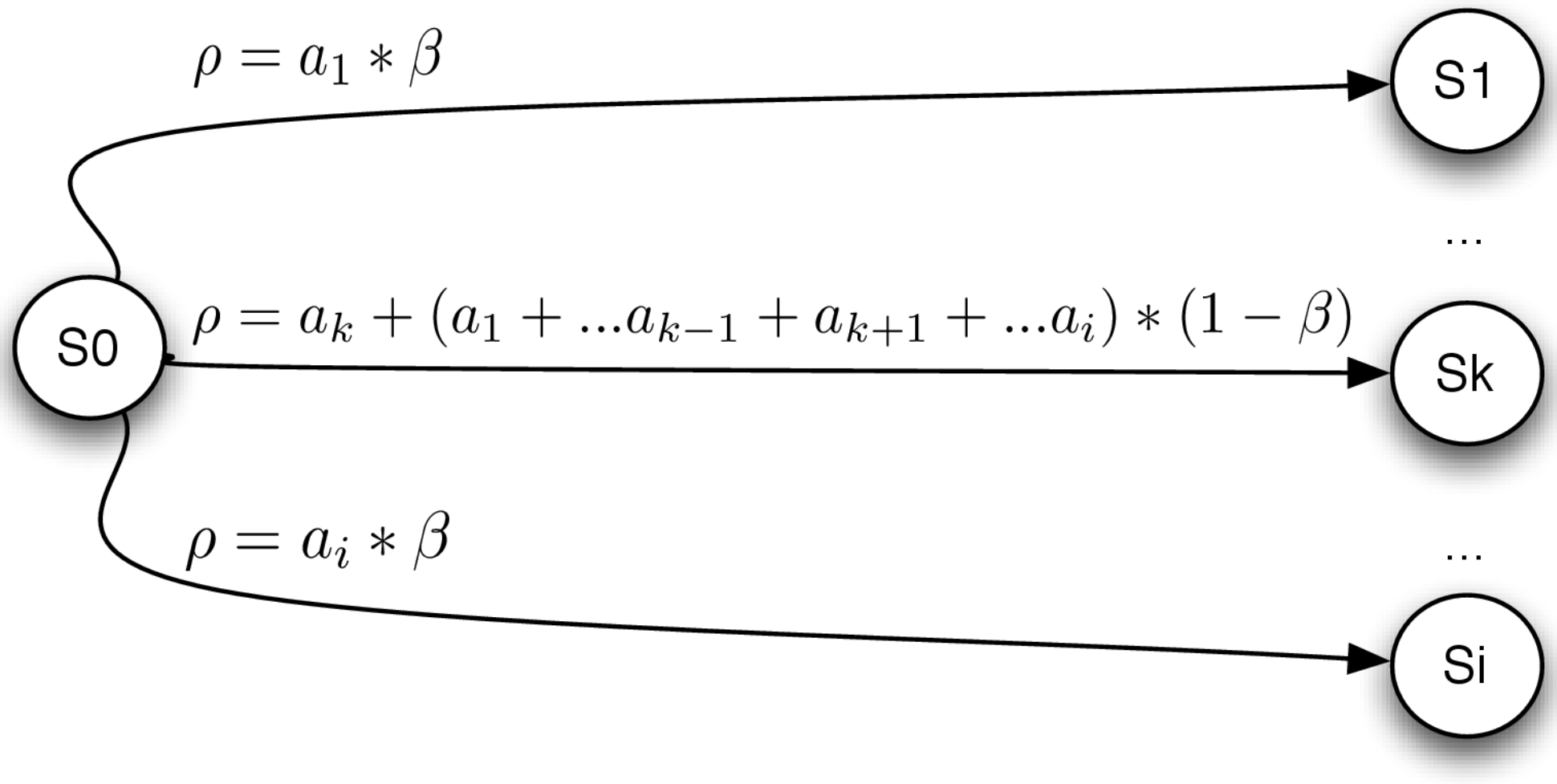

Figure 21: Choosing a transition *k* during improvisation

### 9.2.4 Pitch variation

The idea is finding on which scale the user is playing in. Based on that supposition, we generate new pitches that belong to that scale. This has two problems. First, it is necessary to calculate the scale on which the user is playing. Second, it is necessary to rank the notes of the scale to give a higher priority to some notes of the scale over others.

### 9.2.5 Probability of being on a certain music scale

In order to know on which music scale is the improvisation learned from the user, we count the notes played by the user that are contained in the scale. We are considering five types of scales used in western music [17]: Major, Minor, Pentatonic, Major Harmonic and Minor Harmonic. For each of those, there are 12 scales corresponding to *C*,*C#*,*D*,*D#*,*E*,*F*,*F#*,*G*,*G#*,*A*,*A#*,*B*. Therefore, we are considering 60 different scales. The goal is to find out on which of those 60 scales the user is playing.

For instance, the fragment of the Happy Birthday (fig. 7) was analyzed in the 60 possible scales. We found out that multiple scales have the same result as it is show in table 5 . How can we differentiate, between C Major, C Major Harmonic and A Minor? We tried ranking each degree of the scale, multiplying each degree of the scale by a factor. This partially solves the problem to differentiate among scales, but how can the value of such factors calculated? This would required additional training and it will be specific for some music genres. For that reason, we discarded the development of this extension.

| Scale | Formula | Result |
|---|---|---|
| C Major | | 6 |
| A Minor | | 6 |
| C Major Harmonic | | 3 |
| A minor Pentatonic | | 5 |
| G major | | 6 |

Table 5: Automatically finding the scale for the Happy Birthday fragment

In addition, this idea is not compatible music genres that are not based on the music scales we proposed.

### 9.2.6 Duration variation

In order to preserve the style learned, we are going to replace a note with a duration $\Delta$ by a sequence of notes (already played by the user) whose total duration is equal to the duration of $\Delta$. For instance, in the Happy Birthday fragment, we can replace ($B$,1000,60) by a sequence already played such as

[($G$,500,90),($C$,500,100)]
[($G$,375,80),($G$,125,60),($A$,500,100)]
[($A$,500,100),($G$,500,90)]
preserving the original duration, in this case .

## 9.3 Our previous prototypes for the `ntcc` interpreter

Before developing *Ntccrt*, we explored some combinations of programming languages (C++ and *Common Lisp*) and concurrency models, threads and event-driven programming.

The first problem we faced when designing the interpreter was interfacing *Gecode* to *Common Lisp*. Since *OpenMusic* is written on *Common Lisp*. First, we redesigned the *Gecol* library to work with *Gecode* 2.2.0 (current version of *Gecode*). *Gecol* is an Opensource interface for *Gecode* 1.3.2 originally developed by Killian Sprote. Unfortunately, *Gecol 2* is still a low-level API as *Gecol*. For that reason, using it requires deep knowledge of *Gecode* and it has a difficult syntax.

To fix that inconvenient, we decided to upgrade the *Gelisp* [39] library (originally developed by Rueda for *Gecode* 1.3.2) to *Gecode* 2.2.2. We successfully used this library to solve Constraint Satisfaction Problems (CSP) in the computer music domain in [48]. This library is easy to use and could be the foundation of a new version of *Ntccrt*.

### 9.3.1 Threaded interpreters in Lisp and C++

Using *Gecol 2*, we developed a prototype for the `ntcc` interpreter in Lispworks 5.0.1 professional using Lispworks processes (based on *pthreads*) under Mac OS X. In a similar way, we made another interpreter using C++, *Gecode*, and *Pthreads* (for concurrency control).

In both threaded prototypes, the *tell* agents are modeled as threads adding a constraint to the *store*, which access is controlled by a lock. On the other hand, the *when* processes are threads waiting until the *store* is free and asking if their condition can be deduced from the *store*. If they can deduce its condition they execute their continuation, else they keep asking (fig. 22). The conditions for the *when* processes are represented by boolean variables linked to reified propagators. Fortunately, *Gecode* provides reified propagators for most constraints used in multimedia interaction (e.g. equality and boolean constraints).

0.5

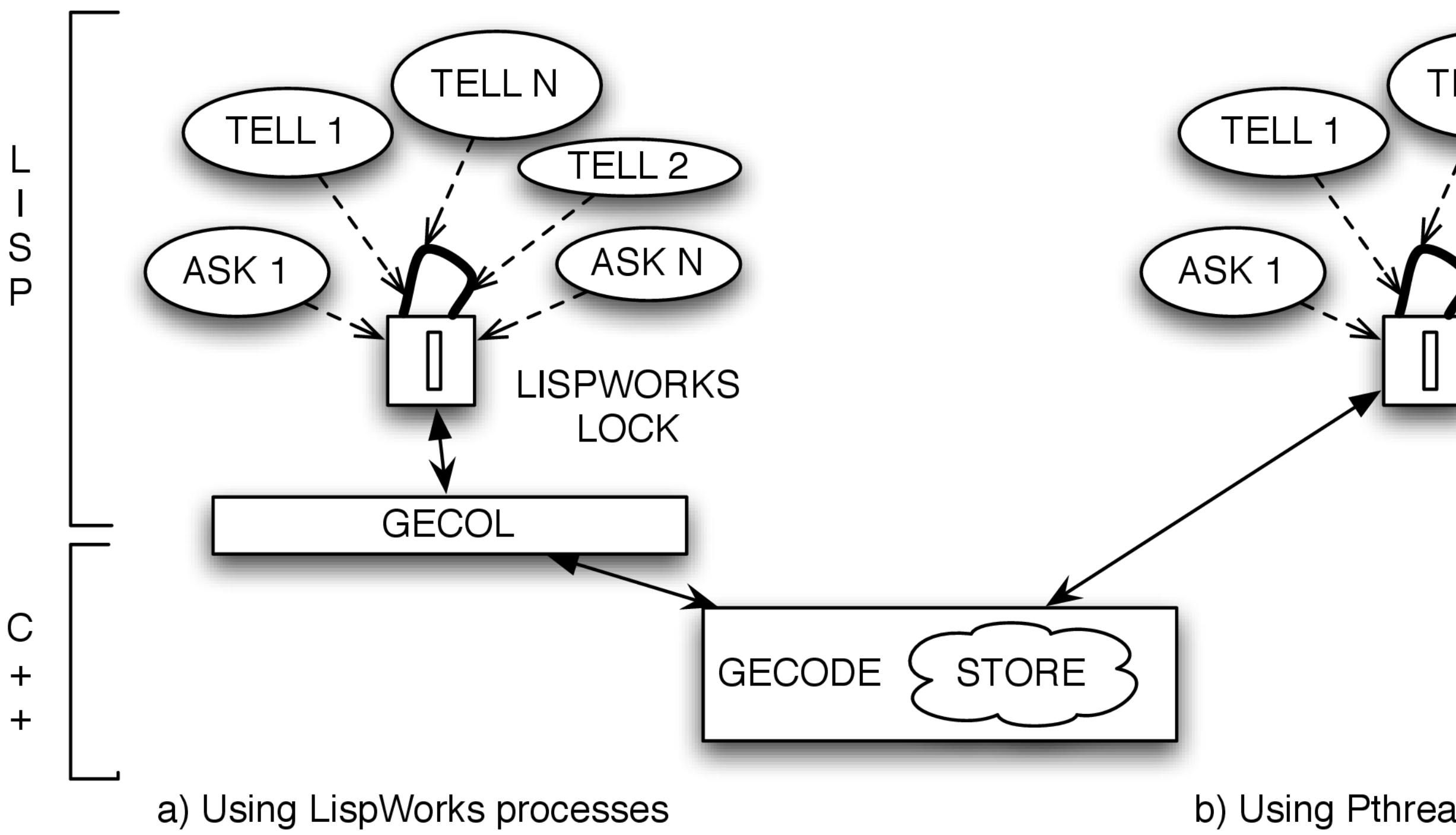

Figure 22: Threaded `ntcc` interpreters using Lispworks and using C++

Since *Gecode* is not thread-safe, we protect the access to the *store* with a lock, synchronizing the access to *Gecode*. A library is thread-safe when it supports the concurrent access to its variables and functions. However, we still have a problem. Each time we want to ask if a condition can be deduced from the *store*, we calculate a *fixpoint*, because propagators in *Gecode* are "lazy"(they only act by demand).

The drawback of both threaded implementations (in C++ and Lispworks) is the overhead of calculating a *fixpoint* each time they want to query if the "when" condition can be deduced. Making extensive use of *fixpoints* would be inefficient even if we use an efficient lightweight threads library such as *Boost* (http://www.boost.org) for C++ .

### 9.3.2 Event-driven interpreter in Lisp

After discarding the threading model, we found a concurrency model giving us better performance. We chose event-driven programming for the implementation of the next prototype. This model is good for a `ntcc` interpreter because we do not use synchronous I/O operations and all the synchronization is made by the *ask* processes (**when**, , and **Unless**) using constraint entailment. The reader may see a comparison between the event-driven prototype and the threaded prototype in chapter 7.

This prototype works on a very simple way. There is an event queue for the `ntcc` processes, the processes are represented by events, and there is a dispatcher handling the events. The handler for the *When* events checks if the boolean variable *b*, representing their waiting condition, is assigned. If it is not assigned, it adds the same *When* event to the queue, else it checks the value of *b*. If *b* is true, it adds the continuation of the *When* events to the event queue, otherwise no actions are taken. On the other hand, the handler for *tell* events add a constraint to the *store*. Finally, the handler for the *Parallel* events adds all its sub-processes to the event queue (fig. 23).

Using *event-driven programming* led us to a faster and easier implementation of `ntcc` than the approaches presented before. However, we realized that instead of creating handlers for *tell*, *ask*, and *parallel*; and a *dispatcher* for processing the events, we could improve the interpreter's performance taking advantage of the *dispatcher* and event queues provided by Gecode for scheduling its propagators, encoding `ntcc` processes as Gecode propagators.

0.5

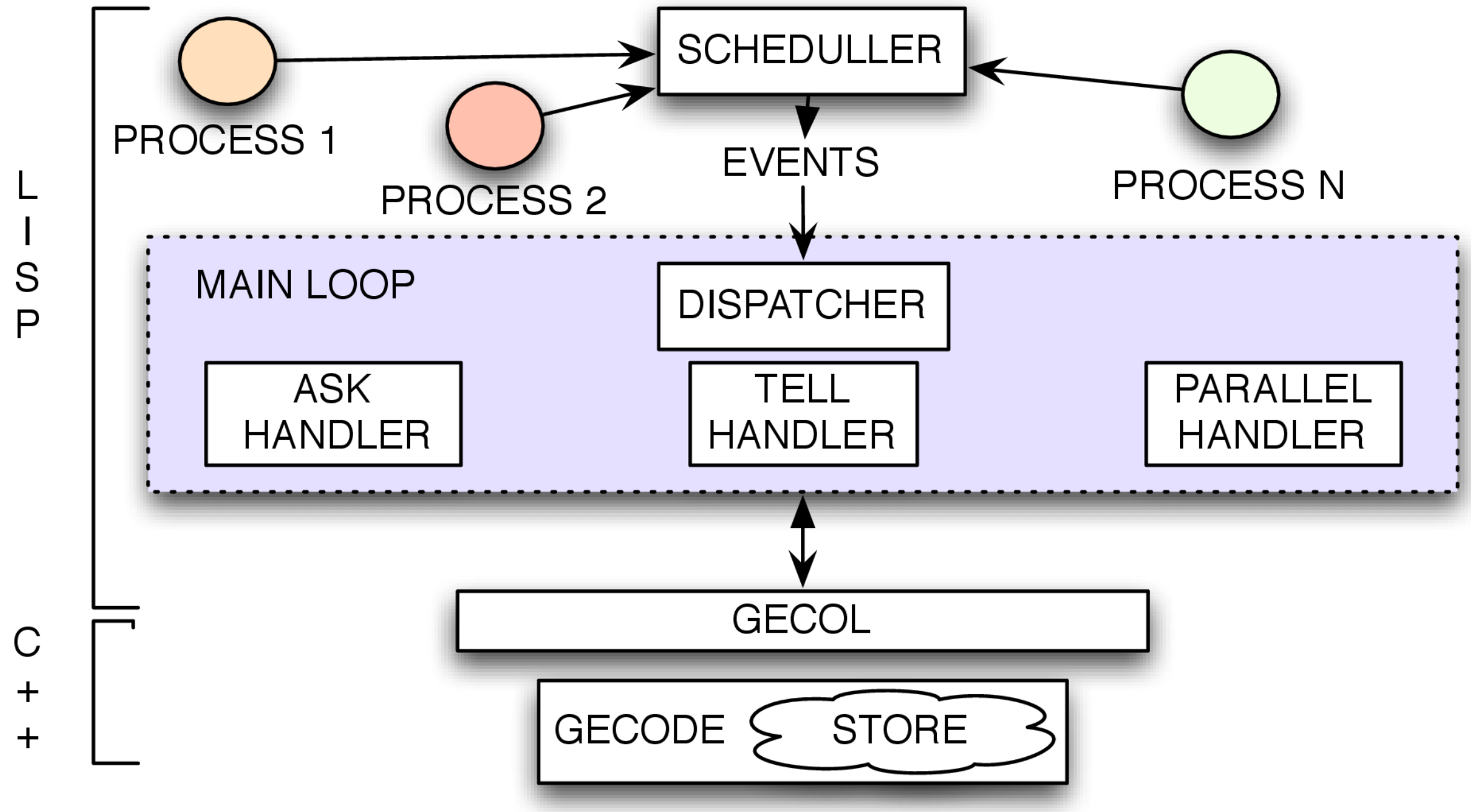

Figure 23: `Ntcc` interpreter using event-driven programming and Gecol 2

## 9.4 Other applications ran in *Ntccrt*

In this section, we present other applications that we ran in *Ntccrt* besides *Ccfomi*. These applications were specified using the graphical interface provided in OpenMusic and they were tested in Pure Data, using an external generated by *Ntccrt*. More details can be obtained at [48].

### 9.4.1 The dining philosophers

Synchronization of multiple operations is not an easy task. For instance, consider the problem of the *dining philosophers* proposed by Edsger Dijkstra. It consists of *n* philosophers sitting on a circular table and *n* chopsticks located between each of them. Each philosopher, is thinking until it gets hungry. Once he gets hungry, he has to take control of the chopsticks to his immediate left and right to eat. When he is done eating, he restarts thinking.

The *dining philosophers* problem mentioned before, has a few constraints. The philosophers cannot talk between them and they require both chopsticks to eat. Furthermore, a solution to this problem must not allow deadlocks, which could happen when all the philosophers take a chopstick and wait forever until the other chopstick is released. Additionally, it must not allow starvation, which could happen if one or more philosophers are never able to eat.

We propose a solution to this problem for *n* philosophers, using the *Ntcc* formalism. All the synchronization is made by reasoning about information that can be entailed (i.e., deduced) from the store or information that cannot be deduced (using the *unless* agent). This way, we can have a very simple model of this problem on which the synchronization is made declarative. The recursive definition *Philosopher*(*i*,*n*) represents a philosopher living forever. The philosopher can be in three different states: thinking, hungry or eating. When the philosopher is on the thinking or eating state, it will choose non- deterministically to change to the next state or remain on the same state in the next time-unit. It means it can choose to go from thinking to hungry or from eating to thinking.

On the other hand, when the philosopher is on the hungry state, it will wait until he can control the first (F) chopstick (left for even numbered and right for odd numbered). As soon as he controls the first chopstick, it will wait until he can control the second (S) chopstick. Once he controls both chopsticks, it will change to the eating state in next time unit.

### 9.4.2 Formal definition

*Philosopher*(*i*,*n*)
**when** **do next**
(**tell** () + **tell** ())
| **when** **do**
**when** **do**
**when** **do next**
(**tell** () |**tell** () |**tell** ())
| **unless** **next**
(**tell** () | **tell** () | **tell** ())
| **unless** **next** (**tell** () | **tell** ())
| **when** **do next**
((**tell** () | **tell** () | **tell** ())+ (**tell** () | **tell** () | **tell** ()))
| **when** $i\%2=0$ **do tell** ($F=(i-1)\%n$) | **tell** ($S=(i+1)\%n$)
| **when** $i\%2=1$ **do tell** ($F=(i+1)\%n$) | **tell** ($S=(i-1)\%n$)
| **next** *Philosopher*(*i*,*n*)

The *Chopstick*(*j*) process chooses non- deterministically one of the philosophers waiting to control it, when the it is not being controlled by a process.

*Chopstick*(*j*)
**unless** **next**
**when** **donext** ( **tell** ())
| **next** *Chosptick*(*j*))

Finally, the system is modelled as *n* philosophers and *n* chopsticks running in parallel. The philosophers start their lives in the thinking state and all the chopsticks are free.

$System(n)$
$(Philosopher(i) \mid Chopstick(i) \mid \mid )$

### 9.4.3 Implementation

Figure 24 shows a *Pd* program where the philosophers are represented as *bangs* (a graphical object design to send a message when the user clicks over it or when it receives a message from another object) and the concurrency control is made by a *Ntccrt external*. When the philosophers start eating, the *Ntccrt external* sends a message to the *bang* changing its color. Chopsticks are represented as commentaries for simplicity.

0.6

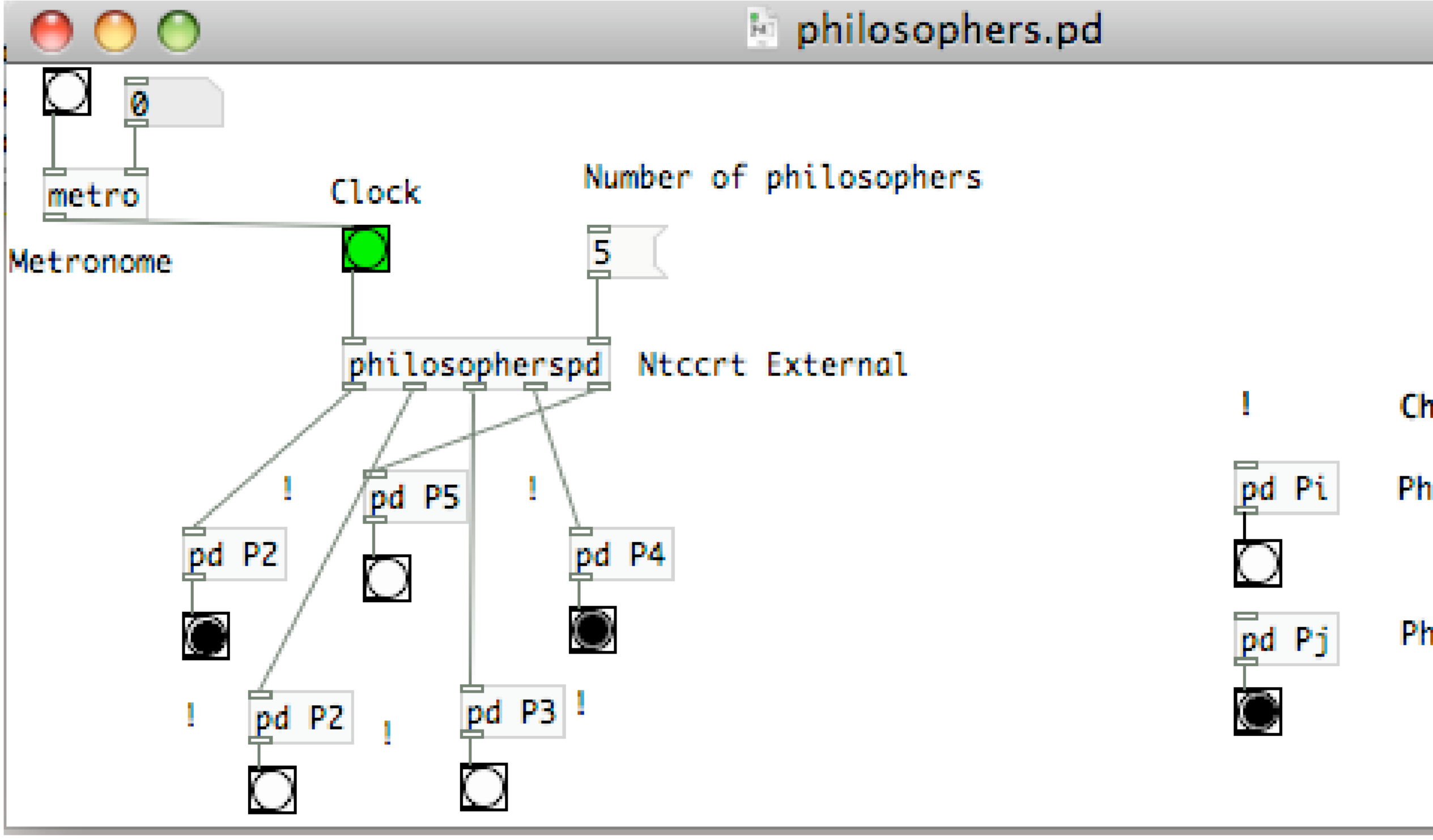

Figure 24: Synchronizing the dining philosophers using a *Ntccrt external* in Pd

### 9.4.4 Signal processing

`Ntcc` was used in the past as an audio processing framework [42]. In that work, Valencia and Rueda showed how this modeling formalism gives a compact and precise definition of audio stream systems. They argued that it is possible to model an audio system and prove temporal properties using the temporal logic associated to `ntcc`. They proposed that each time-unit can be associated to processing the current sample of a sequential stream. Unfortunately in practice this is not possible since it will require to execute 44000 time units per second to process a 44Khz audio stream. Additionally, it poses problems to find a constraint system appropriate for processing signals.

Another approach to give formal semantics to audio processing is the visual audio processing language *Faust* [27]. Faust semantics are based on an algebra of block diagrams. This gives a formal and precise meaning to the operation programmed there. *Faust* has also been interfaced with *Pd* [15].

Our approach is different since we use a `ntcc` program as an *external* for *Pd* or *Max* to synchronize the *graphical objects* in charge of audio, video or MIDI processing in *Pd*. For instance, the `ntcc` *external* decides when triggering a *graphical object* in charge of applying a delay filter to an audio stream and it will not allow other *graphical objects* to apply a filter on that audio stream, until the delay filter finishes its work.

To illustrate this idea, consider a system composed by a collection of *n* processes (graphical objects applying filters) and *m* objects (midi, audio or video streams). When a process is working on an object , another process cannot work on until is done. A process is activated when a condition over its input is true.

The system variables are: represents the identifier of the process working on the object *j*. represents when the object *j* has finished its work. Values for are updated each time unit with information from the environment. represents the conditions necessary to launch process *i*, based on information received from the environment. Finally, represents the set of processes waiting to work on the object *j*.

Objects are represented by the *IdleObject*(*j*) and *BusyObject*(*j*) definitions. An object is *idle* until it non - deterministically chooses a process from the variable. After that, it will remain busy until the constraint can be deduced from the store.

### 9.4.5 Formal definition

*IdleObject*(*j*)
**when do next** *BusyObject*(*j*)
| **unless next** *IdleObject*(*j*)
| **when do tell**
*BusyObject*(*j*)
**when do** *IdleObject*(*j*) | **unless next** *BusyObject*(*j*)

A process *i working* on object *j* is represented by the following definitions. A process is idle until it can deduce (based on information from the environment) that .

*IdleProcess*(*i*,*j*)
**when do** *WaitProcess*(*i*,*j*) | **unless next** *IdleProcess*(*i*,*j*)

A process is *waiting* when the information for launching it can be deduced from the store. When it can control the object, it goes to the *busy* state.

*WaitingProcess*(*i*,*j*)
**when do** *BussyProcess*(*i*,*j*) | **unless next**
*WaitingProcess*(*i*,*j*) | **tell**

A process is *busy* until it can deduce (based on information from the environment) that the process finished working on the object associated to it. *BusyProcess*(*i*,*j*)
**when do** *IdleProcess*(*i*,*j*) | **unless next** *BusyProcess*(*i*,*j*)

This system models a situation with 2 objects and 4 processes. The implementation of this *external* can be adapted to any kind of objects and processes, represented by *graphical objects* in *Pd*. `Ntcc` only triggers the execution of each process , receives an input when the process is done and another input when the conditions to execute the process *i* are satisfied.

*System*()
*IdleObject*(1) | *IdleObject*(2) | *IdleProcess*(1,1) | *IdleProcess*(1,2)
| *IdleProcess*(2,1) | *IdleProcess*(2,2)

### 9.4.6 Implementation

This system is described in OpenMusic using the graphical boxes we provide. We present the graphical description of the processes *IdleProcess*, *BusyProcess* and *WaitingProcess* (see fig. 25).

0.13

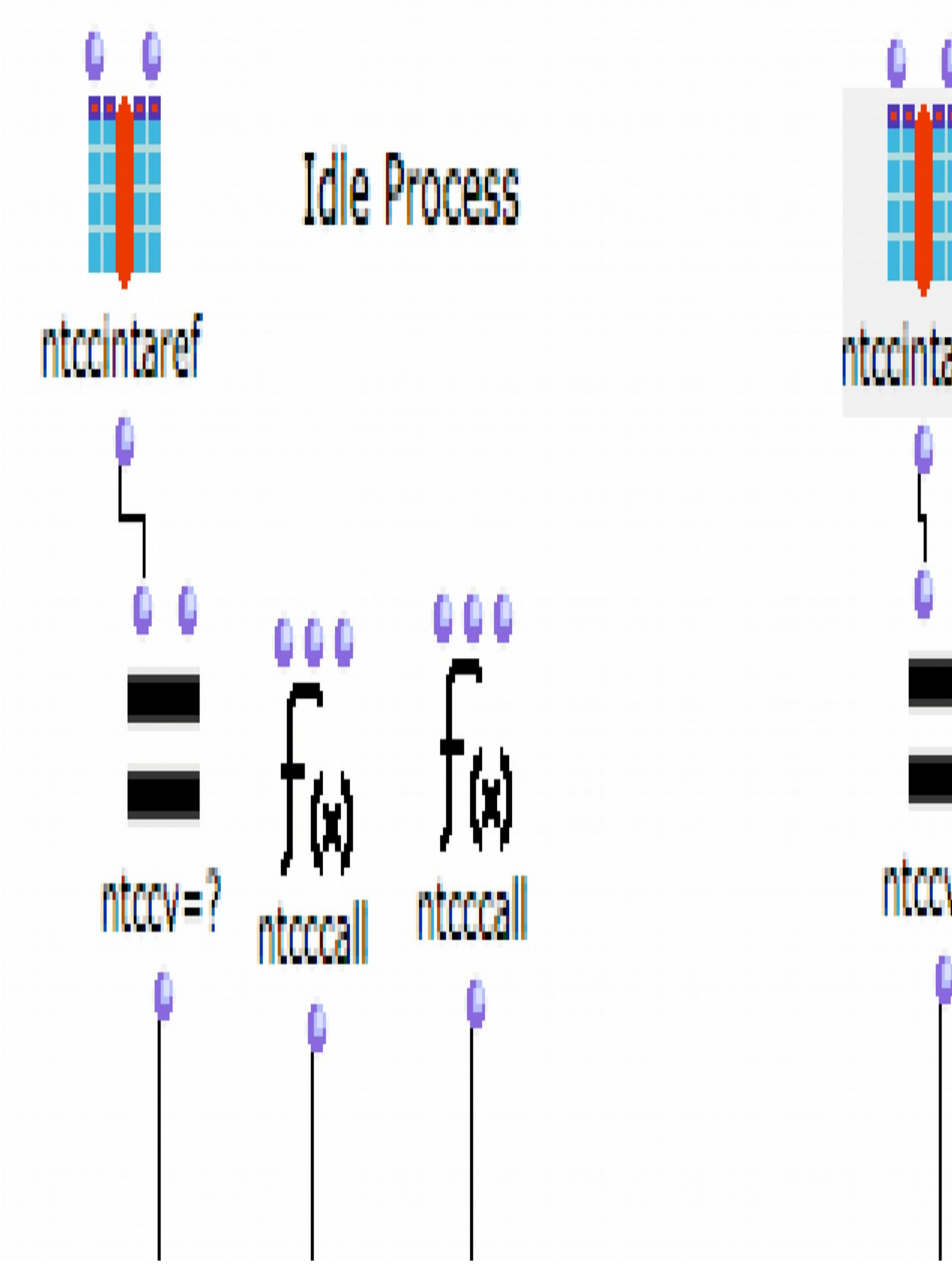
Idle Process
ntccintaref
ntccv=?
ntcccall
ntcccall

Figure 25: Writing a synchronization *Ntccrt external* in OpenMusic

### 9.4.7 Finding paths in a graph concurrently

Following, we describe an application where we use *Ntccrt* to find, concurrently, paths in a graph. The idea is having one *Ntccrt* process for each edge. Each sends *forward* "signals" to its successors and *back* "signals" to its predecessors. When an receives a *back* "signal" and a *forward* "signal", it tells the *store* that there is a path and adds *j* to the set (A finite set variable containing the successors of the vertex *i*). After propagation finishes, we iterate over the resulting sets to find different paths. For instance, we can build a path in the graph getting the lower-bound of each variable .

### 9.4.8 Formal definition

represents an edge in a graph.

**when** **do** (**tell** () | **tell** () )
| **when** **do** **tell** ()
| **when** **do** **tell** ()

The *Main* process finds a path between the vertices *a* and *b* in a graph represented by *edges* (a set of pairs (*i*,*j*) representing the graph edges). The *Main* process calls for each (*i*,*j*)∈*edges* and concurrently, sends *forward* "signals" to processes with the form and *back* "signals" to processes with the form . Notice that sending and receiving those "signals" is greatly simplified by using *tell*, *ask* and the `ntcc` *store*.

*Main*(*edges*,*a*,*b*)
() | **tell** | **tell**

### 9.4.9 Example

Following, we give an intuition about how this system works. To find a path between the vertices 1 and 5 (fig. 26), it starts by sending *forward* "signals" to all the processes with the form and *back* "signals" to all the processes with the form . As soon as an receives a *back* "signal" and a *forward* "signal", it tells the store that there is path (i.e., **tell** () ).

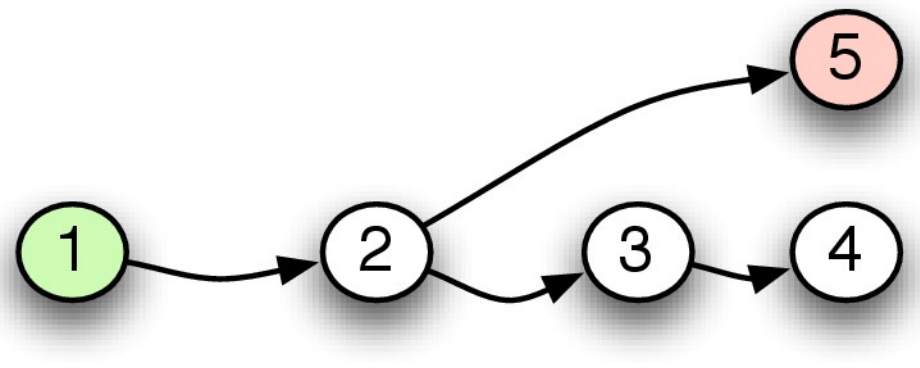

0.8

Figure 26: Example of finding paths in a graph concurrently (1)

Additionally, the reader may notice that there is not a path between vertices 1 and 5 in figure 27. In that example, *back* "signals" sent to processes are not received by any process. Therefore, none of the receives a *back* and a *forward* signal. After calculating a *fixpoint*, we can ask the constraint system for the value of . Since the variable is not bounded, we can infer that there is not a path.

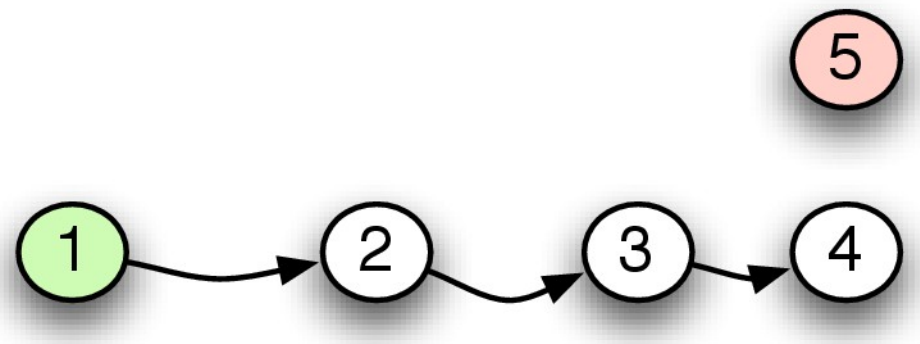

0.8

Figure 27: Example of finding paths in a graph concurrently (2)

## 9.5 Operational semantics from `ntcc` and `pntcc`

"The operational semantics defines the states in which programs can be during execution. This semantics is called this way because is dynamic, that is, it sees a system as a sequence of operations. Each occurrence of an operation is called a transition. A transition system is a structure (Γ,⟶), where Γ is a set of configurations γ , and ⟶⊆Γ×Γ is a transition relation. Notation γ⟶γ' defines the transition from configuration γ to configuration γ'. The transitions are often divided in internal and external, depending on the system's behavior. Normally, external transitions are denoted by ⟹" [45].

### 9.5.1 Operational semantics for `ntcc`

Following, we present the description given in [24]. "Rule OBS says that an observable transition from *P* labeled by (*c*,*d* is obtained by performing a sequence of internal transitions from the initial configuration (*P*,*c*) to a final configuration (*Q*,*d*) in which no further internal evolution is possible. The residual process *R* to be executed in the next time interval is equivalent to *F*(*Q*) (the "future" of *Q*). The process *F*(*Q*), defined below, is obtained by removing from *Q* summations that did not trigger activity within the current time interval and any local information which has been stored in *Q*, and by "unfolding" the sub-terms within "next" and "unless" expressions. This "unfolding" specifies the evolution across time intervals of processes of the form **next** *R* and **unless** *c* **next** *R*."

Following, we present the internal reduction, presented by  and the observable reduction represented by ⇒. "The relations  are the smallest, which obey the form

A rule states that whenever the given conditions have been obtained in the course of some derivation, the specified conclusion may be taken for granted as well. "[45]

TELL

SUM

PAR

UNL

LOC

STAR

REP

STR and

OBS

The future function (F). Let be the partial function defined by

[Sorry. Ignored \begin{cases} ... \end{cases}]

### 9.5.2 Operational semantics for `pntcc`

Following, we present the description given in [29]. “In `pntcc`, an observable transition assumes a particular internal sequence leading to a state where no further computation is possible. F (Q) is obtained by removing from Q summations that did not trigger activity and any local information which has been stored in Q, and by “unfolding” the sub-terms within “next” and “unless” expressions.” Next, we present the internal reduction, presented by and the observable reduction represented by ⇒.

TELL

PSUM

PAR

UNL

LOC

REP

STR and

OBS

The future function (F). Let be the partial function defined by

[Sorry. Ignored \begin{cases} ... \end{cases}]

## 9.6 Tests with Ccfomi: In detail

Detailed information about the tests can be found at http://ntccrt.sourceforge.net/

## 9.7 Tests with the model to find paths in a graph: In detail

Detailed information about the tests can be found at http://ntccrt.sourceforge.net/